%% file: main.tex
\documentclass[10pt,twocolumn,letterpaper]{article}

\usepackage{cvpr}              

\usepackage{graphicx}
\usepackage{amsmath}
\usepackage{amssymb}
\usepackage{booktabs}
\usepackage{multirow}
\usepackage{colortbl}
\usepackage{tabularray}
\usepackage{float}

\usepackage{color}
\usepackage{soul}
\usepackage{enumitem}
\input{utils/color}

\definecolor{gscolor}{rgb}{1.0,0.6,0.0} %

\definecolor{yzybest}{rgb}{0.96, 0.57, 0.58}
\definecolor{yzysecond}{rgb}{0.98, 0.78, 0.57}
\definecolor{yzythird}{rgb}{1.0, 1.0, 0.56}

\usepackage[pagebackref,breaklinks,colorlinks]{hyperref}

\usepackage[capitalize]{cleveref}
\crefname{section}{Sec.}{Secs.}
\Crefname{section}{Section}{Sections}
\Crefname{table}{Table}{Tables}
\crefname{table}{Tab.}{Tabs.}

\def\paperID{3225} 
\def\confName{CVPR}
\def\confYear{2024}

\begin{document}

\title{SIRe-IR: \underline{I}nverse \underline{R}endering for BRDF Reconstruction with \underline{S}hadow and \underline{I}llumination \underline{Re}moval in High-Illuminance Scenes}

\author{
    Ziyi Yang$^{1}$ 
    \quad Yanzhen Chen$^{1}$ 
    \quad Xinyu Gao$^{1}$ 
    \quad Yazhen Yuan$^{2}$ 
    \quad Yu Wu$^{2}$
    \\
    \quad Xiaowei Zhou $^{1}$
    \quad Xiaogang Jin$^{1\dagger}$ 
    \\
    $^1$State Key Laboratory of CAD\&CG, Zhejiang University \quad 
    $^2$Tencent
}

\input{sec/0_abstract}
\input{sec/1_intro}
\input{sec/2_related_work}
\input{sec/3_method}
\input{sec/4_experiment}
\input{sec/5_conclusion}
{\small
\bibliographystyle{ieee_fullname}
\bibliography{ref}
}
\input{sec/X_supple}

\end{document}

%% file: utils/color.tex
\definecolor{yzybest}{rgb}{0.96, 0.57, 0.58}
\definecolor{yzysecond}{rgb}{0.98, 0.78, 0.57}
\definecolor{yzythird}{rgb}{1.0, 1.0, 0.56}

%% file: sec/0_abstract.tex

\twocolumn[{%
\renewcommand\twocolumn[1][]{#1}%
\maketitle
\begin{center}
    \centering
    \captionsetup{type=figure}
    \includegraphics[width=0.95\textwidth]{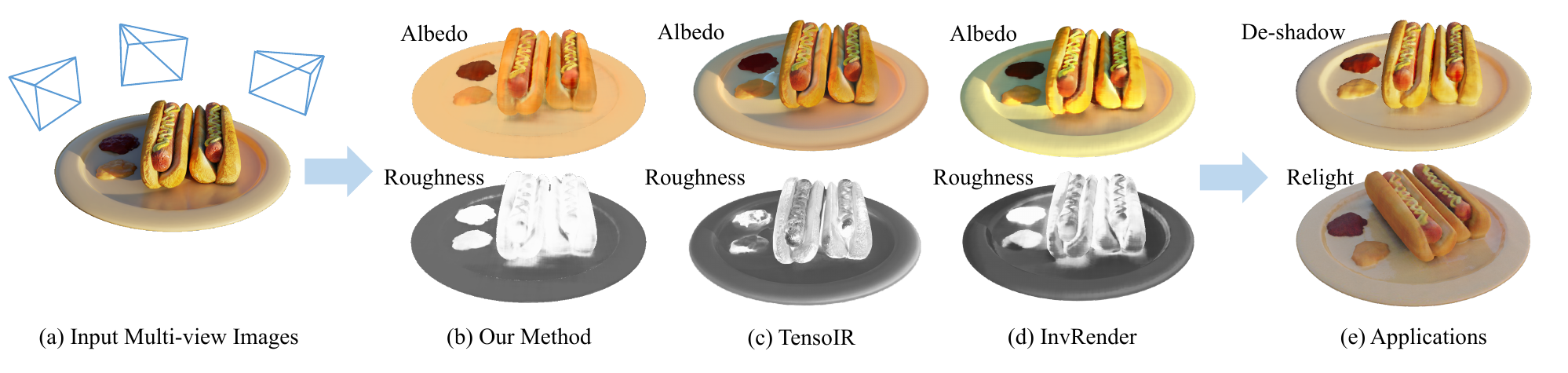}
    \captionof{figure}{Given a set of multi-view images and corresponding camera poses (a), our proposed method, \emph{SIRe-IR}, 
can reconstruct accurate scene geometry and decompose high-quality materials
\textbf{without baking shadow and indirect illumination} into the albedo and roughness (b). In most scenes with intense illumination and shadows, we show that \emph{SIRe-IR} outperforms \emph{TensoIR} \cite{Jin2023TensoIR} (c) and \emph{Invrender} \cite{zhang2022modeling} (d). As a result, our method is suitable for relighting and shadow-free rendering without light and shadow artifacts.}
\label{fig:teaser}
\end{center}%
}]

\begin{abstract}
\vspace{-4mm}
Implicit neural representation has opened up new possibilities for inverse rendering. However, existing implicit neural inverse rendering methods struggle to handle strongly illuminated scenes with significant shadows and indirect illumination. The existence of shadows and reflections can lead to an inaccurate understanding of scene geometry, making precise factorization difficult.
To this end, we present \emph{SIRe-IR}, an implicit inverse rendering approach that uses non-linear mapping and regularized visibility estimation to decompose the scene into environment map, albedo, and roughness.
By accurately modeling the indirect radiance field, normal, visibility, and direct light simultaneously, we are able to remove both shadows and indirect illumination in materials without imposing strict constraints on the scene. Even in the presence of intense illumination, our method recovers high-quality albedo and roughness with no shadow interference. \emph{SIRe-IR} outperforms existing methods in both quantitative and qualitative evaluations. 
We will release our code at \href{https://github.com/ingra14m/SIRe-IR}{https://github.com/ingra14m/SIRe-IR}.
\end{abstract}

%% file: sec/1_intro.tex
\section{Introduction}
\par Inverse rendering, the task of extracting the geometry, materials, and lighting of a 3D scene from 2D images, is a longstanding challenge in computer graphics and computer vision. 
Previous methods, such as providing geometry for the entire scene \cite{nimier2021material, vicini2022differentiable}, modeling shape representation \cite{li2018differentiable, munkberg2022extracting, zhang2021ners, hasselgren2022nvdiffrecmc} or pre-providing multiple known light information \cite{cheng2021multi}, have achieved plausible results using prior information. To achieve clear albedo and roughness decomposition, factors such as light obscuration, reflection, or refraction must be taken into account. Among these, hard and soft shadows are particularly challenging to eliminate, as they play a critical role not only in obtaining cleaner material but also in accurately modeling geometry and light sources. Some data-driven approaches \cite{li2020inverse, sengupta2019neural} have performed plausible shadow removal at the image level. However, these methods are not generally applicable for inverse rendering.

\begin{figure*}[ht] 
  \centering
  \centering
  \includegraphics[width=0.95\textwidth]{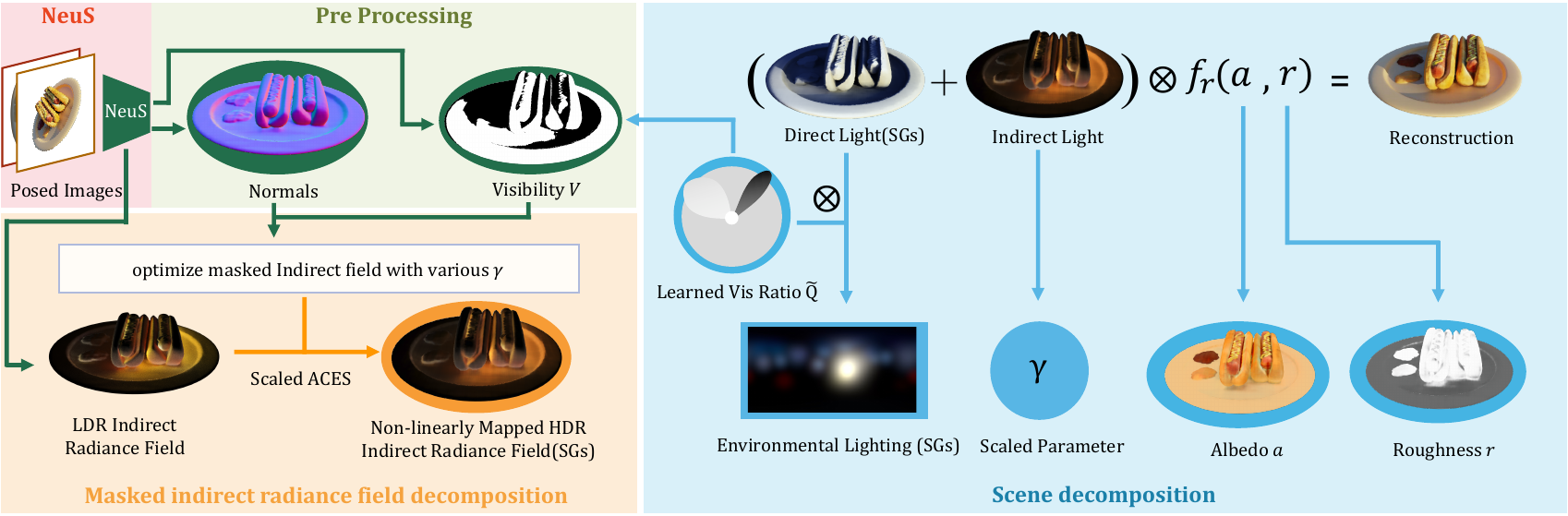}
  \caption{\textbf{The pipeline of our \emph{SIRe-IR}.}
  During the NeuS stage, we reconstruct the scene as an implicit representation by NeuS \cite{wang2021neus}. From the implicit representation, we extract normal (MLPs), indirect radiance field (SGs), and visibility $V$ (MLPs). Next, by taking ACES tone mapping into account, we train a masked non-linearly mapped indirect radiance field with visibility and randomly sampled $\gamma$ values. During scene decomposition, we optimize learnable parameters, including environmental lighting, the scaled parameter $\gamma$, albedo $a$, and roughness $r$, to minimize reconstruction loss under the constraint of the rendering equation. To reduce the stubborn visibility errors caused by reflection, we employ a learnable visibility ratio of direct SGs, $\tilde Q$, and perform regularized visibility estimation on $V$ to obtain more accurate visibility, which is critical for eliminating shadows at the edges and boundaries.
  } \label{fig:pipeline}
\end{figure*}

\par Since the advent of NeRF \cite{mildenhall2020nerf}, implicit neural representation has garnered significant interest in portraying scenes as neural radiance fields. 
Furthermore, the high-quality geometry and radiance fields modeled by NeRF are exceptionally useful for \textit{inverse rendering}. 
By applying implicit neural representation to inverse rendering \cite{boss2021nerd, knodt2021nreural, zhang2021physg}, plausible factorization can be achieved in simple scenes with weak light intensity. Thanks to NeRFactor \cite{zhang2021nerfactor} and its relevant work \cite{chen2022tracing}, which extend previous works by explicitly representing visibility, implicit inverse rendering can be improved with simple shadow removal and clear edge in albedo and roughness.
Recently, InvRender \cite{zhang2022modeling} has taken the scene factorization problem to a new level by modeling indirect illumination, serving as the baseline in our experiment.

\par The current methods for implicit inverse rendering mentioned above have shown limitations when dealing with scenarios with intense illumination and strong shadow, which reflects the inaccuracy modeling of each decomposed part. 
To deal with such scenes, the following challenges arise in order to obtain clear physically based rendering (PBR) materials.

\par First, previous methods for inverse rendering have faced the challenge in separating complex light phenomena such as reflection and obscuration in low dynamic range (LDR) space. While these methods perform well for scenes with weak light intensity, they struggle to accurately decompose albedo and roughness in scenarios with intense lighting. As shown in Fig. \ref{fig:teaser}, both shadow and indirect illumination remain in the albedo.
According to the rendering equation, decreasing the intensity of direct and indirect illumination in the shadow area is required to remove shadows, while increasing the intensity of indirect illumination is necessary to eliminate the remaining indirect light in the albedo.
To address the aforementioned challenge, we propose a novel approach that explicitly applies a non-linear mapping function on the indirect radiance field to nonlinearly and monotonically map the value $\in [0, 1]$ of indirect illumination intensity into a wider range and enhance the contrast between different regions. Specifically, we propose an ACES tone mapping search algorithm that can automatically learn the appropriate non-linear mapping curve for a specific scene, eliminating the need for additional parameters such as camera exposure time.

\par Second, previous approaches have not adequately modeled decomposed components. For example, the discontinuous indirect radiance field has difficulty accurately converging based on smooth spherical Gaussians (SGs), particularly in scenes with intense directional lighting. 
Our approach accurately models the indirect radiance field and visibility simultaneously by introducing a novel masked indirect radiance field. We also apply octree tracing instead of sphere tracing to improve the speed and precision of ray intersection. These enable us to better remove indirect illumination without strong constraints on the scene.

\par Third, existing training strategies encounter difficulties in accurately modeling visibility due to the sheer number of learning parameters. To address this, we introduce a prior assumption that the visibility ratio \cite{zhang2022modeling} at a specific point x from a given direct light SG is smoothly varying. Consequently, we employ a Regularized Visibility Estimation (RVE) to achieve more accurate visibility. This technique significantly contributes to the scene decomposition, enabling the separation of environment maps, albedo, and roughness without the baked shadows.

\par In summary, the major contributions of our work are:
\begin{itemize}[noitemsep,nolistsep,leftmargin=*]
    \item 
    A novel non-linearly mapped indirect radiance field for the inverse rendering task. It enables the production of clean albedo and roughness in scenes with intense lighting and strong shadows, allowing for the creation of high-quality visual content.
    \item A novel masked indirect light representation that uses SGs and a visibility network to accurately model visibility and discontinuous indirect radiance field simultaneously. 
    \item A novel regularized visibility estimation that uses an intermediate layer to fine-tune the visibility field. It reduces shadow residue and improves the convergence stability of the ill-posed inverse rendering task.
\end{itemize}

%% file: sec/2_related_work.tex
\section{Related Work}
\subsection{Inverse Rendering}
\par Inverse rendering is a process in computer graphics that aims to derive an understanding of the physical properties of a scene from a set of images. 
Because the problem is highly ill-posed, most previous works have incorporated priors such as illumination, shape, and shadow, as well as additional observations such as scanned geometry \cite{nimier2021material, schmitt2020joint, lensch2003image} and known light conditions \cite{cheng2021multi}, to ensure proper regularization during the optimization process of the rendering components.
Simplified approaches, such as those assuming outdoor and natural light \cite{song2022novel} or white light \cite{10.1145/3374753}, aim to reduce the number of fitting parameters in an ill-posed problem. Recently, data-driven methods \cite{barron2014shape, li2018learning, sang2020single, sengupta2019neural, yu2019inverserendernet, boss2020two} have focused on decomposing scene information from a single or two-shot image(s), heavily relying on geometric priors and training complexity. In contrast, our research focuses on a more general inverse rendering framework that reduces the model's reliance on specialized equipment and scene complexity while also improving model generalization through more efficient use of geometric prior. 

\subsection{Implicit Neural Representation}
\par Neural rendering has gained popularity due to its ability to produce photorealistic images.
Recently, NeRF \cite{mildenhall2020nerf} enables photo-realistic novel view synthesis using MLPs. It can handle complex light scattering and reconstruct high-quality scenes for downstream tasks. 

\par Subsequent work has enhanced NeRF's efficiency in various ways, elevating it to new heights and enabling its use in other domains. Structure-based techniques \cite{yu2021plenoctrees, garbin2021fastnerf, reiser2021kilonerf, hedman2021baking, chen2022mobilenerf} have explored ways to improve inference or training efficiency by caching or distilling implicit neural representation into the efficient data structure. Hybrid methods \cite{liu2020neural, martel2021acorn, sun2022direct, sun2022improved, Chen2022ECCV} aim to improve the efficiency by incorporating explicit voxel-based data structures. Among them, Instant-NGP \cite{mueller2022instant} achieves minute training by additionally incorporating hash encoding.

\par In addition, some follow-up methods \cite{oechsle2021unisurf, wang2021neus, yariv2020multiview} are dedicated to recovering clear surfaces for scenes with complex solid objects by modeling a learnable SDF network, the value of which indicates the minimum distance between the input coordinate and surfaces in the scene. 
In our work, we show that the simple and continuously differentiable nature of SDF makes it suitable for learning geometry priors in inverse rendering. Furthermore, drawing inspiration from PlenOctree \cite{yu2021plenoctrees}, we construct an octree tracer from the SDF to improve inference efficiency and accuracy compared to sphere tracing.

\subsection{Implicit Neural Inverse Rendering}
\par In recent years, there has been a surge of interest in implicit inverse rendering, building on the success of NeRF and its fully differentiable implicit representation. To model spatially-varying bidirectional reflectance distribution function (SVBRDF) under more casual capture conditions, many recent methods \cite{boss2021nerd, knodt2021nreural, boss2021neural, boss2022samurai, yao2022neilf, zhang2022iron} have relied on implicit representation. Other works \cite{zhang2021nerfactor, srinivasan2021nerv, yang2022ps, Jin2023TensoIR, liu2023nero} have focused on physical-based modeling for complex scenes via visibility prediction. L-Tracing \cite{chen2022tracing} introduced a new algorithm for estimating visibility without training, while NeRFactor \cite{zhang2021nerfactor} proposed a canonical normal and BRDF smoothness to address NeRF's poor geometric quality, which is critical to the decomposition stage.

\par Inverse Rendering with dynamic light \cite{kuang2022neroic} has also shown promise by exploiting illumination differences between input images and decomposing them into multiple low-rank principal components. In contrast, our work focuses on a more arbitrary but static light condition.

\par Our work builds on the implicit representations, significantly enhancing inverse rendering capabilities beyond prior methods. Specifically, we are inspired by PhySG \cite{zhang2021physg}, which augmented the environment light modeling through the Spherical Gaussians (SGs), and InvRender \cite{zhang2022modeling}, which extends previous work by modeling indirect illumination. 

\subsection{Microfacet BRDF}
\par On the point \textbf{x}, the color $c$ is calculated by the rendering equation: 
\begin{equation}
\label{equ:rendering-eqation}
    c\left(\mathbf{x}, \omega_{\mathbf{o}}\right)=\int_{\Omega} f_{r}\left(\mathbf{x}, \omega_{\mathbf{i}}, \omega_{\mathbf{o}}\right) L\left(\mathbf{x}, \omega_{\mathbf{i}}\right) (\omega_i \cdot \textbf{n}) d \omega_{\mathbf{i}},
\end{equation}
where $c(\mathbf{x},\mathbf{\omega_o})$ is the output color leaving point $\mathbf{x}$ in the view direction $\mathbf{\omega_o}$, $f_r(\mathbf{x}, \mathbf{\omega_i}, \mathbf{\omega_o})$ is the BRDF function, $L(\mathbf{x},\mathbf{\omega_i})$ is the incoming radiance at point $\mathbf{x}$ from direction $\mathbf{\omega_i}$, and \textbf{n} is the surface normal. In SIRe-IR, the normal \textbf{n} is computed from NeuS. We use the simplified Disney BRDF \cite{burley2012physically}, which comprises both diffuse and specular (denoted as $f_s$ for further discussion) components:
\begin{equation}
    \label{equ: disney-brdf}
    f_{r}\left(\mathbf{x}, \omega_{\mathbf{i}}, \omega_{\mathbf{o}}\right) = \frac{a}{\pi} + \frac{\mathcal{DFG}}{4(\omega_i \cdot \textbf{n})(\omega_o \cdot \textbf{n})},
\end{equation}
where $a$ is the albedo color of the point, $\mathcal D$ is the normal distribution function, $\mathcal F$ is the Fresnel term and $\mathcal G$ is the geometry term. $\mathcal D$, $\mathcal F$ and $\mathcal G$ are all determined by the metallic $m$, the roughness $r$ and the albedo $a$. 
By substituting the color function in the NeuS \cite{wang2021neus} framework with the shading function based on Equ. \eqref{equ: disney-brdf}, we can achieve robust PBR material decomposition through image loss.

%% file: sec/3_method.tex
\section{Methodology}

\subsection{Overview}
Our proposed method addresses the challenging problem of PBR materials 
 decomposition from a set of posed images with strong shadows and indirect illumination. As shown in Fig. \ref{fig:pipeline}, our framework utilizes a robust training scheme consisting of four sequential phases. First, we train NeuS $S(x, \omega)$ as the representation of the scene. Second, we refine and smooth the noisy normal field obtained from the NeuS (Sec. \ref{sec:normal-map}). Third, we employ MLP to learn a compact visibility representation, which we then utilize to learn a masked indirect radiance field (Sec. \ref{sec:segment_indir}). Finally, we decompose the color into albedo, roughness, environment light, and other components with the help of the rendering equation and a non-linearly mapped indirect radiance field (Sec. \ref{sec:hdr_light}). Additionally, we fine-tune the previously learned visibility and normal through regularized estimation, which reduces geometric errors and improves decomposition stability (Sec. \ref{sec:advopt}).

\subsection{Optimize Noisy Normal Field}
\label{sec:normal-map}
\par In our framework, the accuracy of normal vectors is crucial for reliable visibility prediction, precise indirect illumination modeling, and effective decomposition of materials. However, we observed that normals estimated from NeuS tend to be noisy, which compromises the accuracy in subsequent stages of our model. To overcome this, we drew inspiration from Ref-NeRF \cite{verbin2022ref}. Specifically, we predict cleaner normals for each point \textbf{x} using a spatial MLP, aiming to reduce noises. 
We align these predicted normals with the density gradient normals obtained from NeuS using a combination of $\mathcal L_2$ and smooth loss:
\begin{equation}
    \label{equ:normal}
    \mathcal L_{norm} = \dfrac{1}{N}( \Vert n - \hat{n}\Vert_2^2 + \Vert n - n^{\prime}\Vert_2^2), 
\end{equation}
where $n$ denotes the normal learned by MLP, $\hat{n}$ denotes the supervision normal value from NeuS, and $n^{\prime}$ is a smoothing term outputted by the normal MLP, obtained by adding $0.02\times$ Gaussian noises to the input \textbf{x}.

\subsection{Visibility and Masked Indirect Illumination}
\label{sec:segment_indir}
\par Following InvRender \cite{zhang2022modeling}, we model the indirect radiance field $L_I(x, \omega_i)$ at the intersection point $x$ using SGs. This is achieved by firstly performing octree tracing along direction $\omega_i$ to get the second intersection point $\hat x$. Then the indirect radiance field is supervised by the out-going radiance $S_o (\hat x, -\omega_i)$ at $\hat{x}$.
However, this approach often results in an indirect illumination representation that is too smooth to accurately capture the nuances of the actual lighting field. A visualization and detailed analysis can be found in the supplementary materials. In this modeling method, the parts that hit the second intersection point are given supervisory values, while the parts that do not hit are assigned a value of 0, resulting in discontinuity. This causes the SG modeling to be inaccurate, leading to residual indirect illumination in the decomposed albedo.

\par To address the aforementioned issue, 
we propose a masked method that separates the indirect radiance field into occlusion and non-occlusion parts using a learnable visibility network. Non-occlusion parts refer to the rays that pass through an object with only one bounce.  By only optimizing the occlusion part of the indirect radiance field, we can model indirect illumination more accurately.

\par The calculation of visibility is a critical step in indirect illumination training. However, performing sphere tracing at surface points requires a significant amount of time and memory.
To overcome this limitation, we use an octree tracer extracted from SDF to accelerate the tracing and obtain more precise intersection results. 
In addition, we followed \cite{zhang2022modeling} to improve efficiency by compressing the visibility field into an MLP 
that maps the surface point $\mathbf{x}$ and direction $\boldsymbol{\omega}$ to visibility $V(\mathbf x, \boldsymbol{\omega})$, providing a compact and continuous representation. Then, the indirect light $L_I$ with a mixture of $M=24$ SGs can be divided by visibility:
\begin{equation}
    \label{equ:indirect-specular}
    \begin{aligned}
        L_I(\mathbf x, \boldsymbol{\omega}; \Gamma) &= (1 - V(\mathbf x, \boldsymbol{\omega}))\sum_{j=1}^M G(\boldsymbol\omega; \Gamma_j(\mathbf x, \gamma)), \\
        G(\boldsymbol{\omega}; \boldsymbol{\xi}, \lambda, \boldsymbol{\mu}) &= \boldsymbol{\mu} e^{\lambda(\boldsymbol{\omega}\cdot \boldsymbol{\xi} - 1)},
    \end{aligned}
\end{equation}
where we use MLP $\Gamma$ to output the $j$th SG parameters ($\boldsymbol{\xi} \in R^3$ is the lobe axis, $\lambda \in R^1$ is the lobe sharpness, and $\boldsymbol{\mu} \in R^3$ is the lobe amplitude), $G$ denotes function of spherical Gaussians, and $\gamma$ denotes the scale factor, which will be illustrated in Sec. \ref{sec:hdr_light}. 

\par The indirect radiance field is supervised by the color of the second intersection sample $\hat x$ from the prior learned radiance fields $ S(\hat x, \omega)$ from NeuS. To enhance the convergence stability, we use \emph{softplus} as the activation function. $\mathcal L_{indir}$ and $\mathcal L_{vis}$ are optimized by $\mathcal L_1$ and binary cross entropy loss as follows:
\begin{equation}
\label{equ:loss-indirect}
\begin{aligned}
\mathcal L_{indir} &= \dfrac{1}{N} \Vert \hat L_{I} - L_{I} \Vert_1, \\
\mathcal L_{vis} &= \dfrac{1}{N}\sum_{i=1}^{N} \Psi(V(\mathbf x, \boldsymbol{\omega}), \hat V(\mathbf x, \boldsymbol{\omega})),
\end{aligned}
\end{equation}
where $\Psi(p_i \ \Vert \ y_i)$ represents the binary cross-entropy (BCE) loss, $\hat L_{I}$ is the radiance value at the second intersection point $\hat x$ obtained by querying NeuS, and $\hat V(\mathbf x, \boldsymbol{\omega})$ is calculated using  an octree tracer from point $\mathbf{x}$ along direction $\boldsymbol{\omega}$.

\subsection{Decomposition with Non-linearly Mapped Indirect Radiance Field} \label{sec:hdr_light}
\par During the scene decomposition stage, we use differentiable rendering to factorize the scene's materials and adopt $M=128$ learnable SGs to model direct illumination. However,
previous approaches tend to leave shadow and indirect illumination in albedo under scenes with high light intensity, which necessitate decreasing and increasing illumination in the corresponding regions to eliminate.
Therefore, we need to apply a non-linear mapping to the light, making the light intensity weaker in shadowed areas and stronger in reflection areas.
Inspired by \cite{mildenhall2022nerf}, we apply HDR tone mapping, which is a non-linear curve, to the masked indirect radiance field and the direct illumination learned at this stage will be transformed into the same value domain as the non-linear mapping of the indirect radiance field.

\paragraph{HDR Tone Mapping.}
\par Several recent works \cite{huang2022hdr, mildenhall2022nerf} have incorporated HDR into NeRF for specific applications. We adopt the widely used Academy Color Encoding System (ACES) \cite{arrighetti2017academy} tone mapping. Specifically, we apply the ACES tone mapping function $\mathcal F$ to the input HDR color $e$, which is formulated as:

\begin{equation}
\label{equ:aces}
\mathcal F(e) = \frac{(2.51 e + 0.03) e}{(2.43 e + 0.59) e + 0.14},
\end{equation}
whereas for the input LDR color $c$, we use the ACES inverse tone mapping function $\mathcal F_{\rm I}$, which is given by:

\begin{footnotesize}
\begin{equation}
\label{equ:aces-inverse}
\mathcal F_{\rm I}(c) = \frac{0.59 c-0.03+\sqrt{-1.0127 c^{2}+1.3702 c+0.0009}}{2(2.51-2.43 c)}.
\end{equation}
\end{footnotesize}

\paragraph{Automatic ACES mapping search.}
\par Given that the light intensity varies across different scenes, applying ACES tone mapping universally is not feasible. To remedy this, we introduce an additional learnable parameter, $\gamma$, which ranges within (0, 1]. This parameter modifies the ACES tone mapping curve, enabling it to automatically adapt to each scene's unique illumination intensity. The resulting deformed tone mapping function is defined as follows:
\begin{equation}
\label{equ:hdr-aces}
\begin{aligned}
\mathcal F^\gamma(e) &= \max(0, \min(1, \gamma^{-0.2} \mathcal F( e))), \\
\mathcal F^\gamma_{\rm I}(c) &= \max(0, \min(1, \mathcal F_{\rm I}( c \cdot \gamma^{0.2}))).
\end{aligned}
\end{equation}

\par 
However, simultaneous training of the indirect radiance field with varying $\gamma$ and the material decomposition model is challenging. To overcome this issue, we separate the scene-specific $\gamma$ learning into two stages. 
In the first stage, as shown in Equ. \eqref{equ:indirect-specular}, we train the indirect radiance field by treating $\gamma$ as an explicit input, randomly sampling all possible values of $\gamma$. Consequently, the loss function $\mathcal L_{indir}$ in Equ. \eqref{equ:loss-indirect} is then revised to include $\gamma$ as follows:
\begin{equation}
\label{equ:loss-hdr-indirect-light}
\begin{aligned}
\mathcal L_{indir} &= \dfrac{1}{N} \Vert \mathcal F^\gamma_{\rm I}(\hat L_{I}) - L_{I}\Vert_1.
\end{aligned}
\end{equation}

Up to this point, we obtain the non-linearly mapped indirect radiance field with coefficient $\gamma$. We then stop training the indirect radiance field and treat $\gamma$ as a learnable parameter. The optimal $\gamma$  for the current scene will be determined as the decomposition model converges.

\paragraph{Material Decomposition.}
\par So far, we have faithfully reconstructed the geometry, visibility and the indirect radiance field of the scene. We aim to accurately evaluate the rendering equation in order to precisely estimate the surface BRDF i.e. albedo $a$, and roughness $r$. Following \cite{zhang2022modeling}, we represent PBR materials using an encoder-decoder network. 
The network initially encodes the input surface point \textbf{x} into its corresponding latent code \textbf{z} and then decodes it into albedo and roughness.
To simplify the computation, we assume dielectric materials with a fixed Fresnel term value of $F_0 = 0.02$. 

\par Combining Equ. \eqref{equ:rendering-eqation} and Equ. \eqref{equ: disney-brdf}, and following \cite{zhang2021physg}, we can convert the specular component $f_s$ and cosine factor $\omega_i \cdot \textbf{n}$ to a single SG respectively and approximate the diffuse and specular color from direct illumination at surface point \textbf{x} as the inner product ($\otimes$) of SGs:
\begin{equation}
\begin{aligned}
\label{equ:spec-diffuse}
    c(\textbf{x}) &= c_{\text{diffuse}} + c_{\text{specular}}, \\
    c_{\text{diffuse}} &= \frac{a}{\pi} \sum^{M}_{j=1} G_j(\omega_i; \boldsymbol{\xi}, \lambda, \eta \boldsymbol{\mu}) \otimes (\omega_i \cdot \textbf{n}), \\
    c_{\text{specular}} &= \sum^{M}_{j=1} f_s \otimes G_j(\omega_i; \boldsymbol{\xi}, \lambda, \eta^{\prime} \boldsymbol{\mu}) \otimes (\omega_i \cdot \textbf{n}),
\end{aligned}
\end{equation}
where $G$ is the direct SGs learned in this stage, $\eta = \frac{\sum_{i=0}^{S} G\left(\boldsymbol{\omega_i}\right) V\left(\mathbf{x}, \boldsymbol{\omega_i}\right)}{\sum_{i=0}^{S} G\left(\boldsymbol{\omega_i}\right)}$ signifies the visibility ratio for direct SGs obtained by randomly sampling $S$ directions, $\eta^{\prime}$ is determined by specular SG to accurately model the specular color. For more detailed about $c_{\text{specular}}$ and SG convertion, please see the supplementary materials. The rendering of indirect illumination is similar to Equ. \eqref{equ:spec-diffuse}, except that the SGs queried are masked the indirect SGs modeled in Sec. \ref{sec:segment_indir}.
To further reduce the noises in materials, we also add the same smooth loss in Equ. \eqref{equ:normal} to the albedo and roughness.
\subsection{Regularized Visibility Estimation}
\label{sec:advopt}
\par One of our primary goals is to achieve clean albedo with no residual shadows, which are typically caused by direct lighting and inaccurate visibility. Despite all efforts of the previous stages, a small amount of stubborn visibility errors caused by reflectance still exist, affecting the actual contribution of direct light.

\par To this end, we introduce a prior assumption stating that "the visibility ratio of a specific point x from a given direct SG is smooth" to further optimize visibility, and present regularized visibility estimation (RVE) that utilizes an intermediate layer $\tilde Q(\mathbf x, \boldsymbol{\tau})$ to jointly optimize against the previously learned visibility network $V(\mathbf x, \boldsymbol{\omega})$.
Specifically, 
$\tilde Q(\mathbf x, \boldsymbol{\tau})$ is a visibility prediction network learned from scratch, indicating the visibility ratio of point $\mathbf x$ to the direct SG, while $\boldsymbol{\tau}$ represents the one-hot embedding of each direct SG. Since visibility errors primarily occur at the edges and boundaries, which are also sparse in the scene, we leverage the sparse loss to make the residual sparse:
\begin{equation}
    \label{equ:return}
    \begin{aligned}
        \mathcal L_{sparse} &= \operatorname{KL}(\tilde Q(\mathbf x, \boldsymbol{\tau}) - \eta(\mathbf x) \Vert \epsilon),
    \end{aligned}
\end{equation}
where $\operatorname{KL}(\rho\  \Vert \  \hat{\rho}) = \rho log\frac{\rho}{\hat \rho} + (1 - \rho)log\dfrac{1-\rho}{1- \hat \rho}$ represents Kullback-Leibler divergence loss that measures the relative entropy of two probability distributions, and $\epsilon$ is set to $0.01$. 
\begin{figure}[ht]
    \centering
    \includegraphics[width=0.45\textwidth]{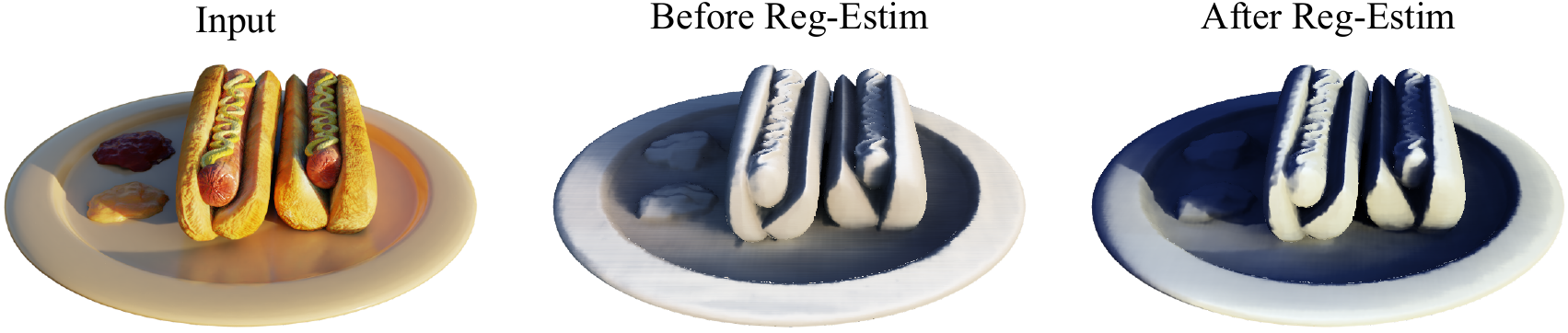}
    \caption{\textbf{Visualization of masked direct light SGs.} We run experiments to visualize masked direct light SG before and after regularized visibility estimation. Persistent visibility errors result in incorrect masks at shadow boundaries, which are critical for removing shadows in albedo. We optimize visibility errors by using regularized visibility estimation to provide direct light SGs with more accurate masks for decomposition.}
    \label{fig:adv-vis}
\end{figure}

\par The regularized visibility estimation creates a gradient between $Q(\mathbf x, \boldsymbol{\tau})$ and the octree-based visibility network $V(\mathbf x, \boldsymbol{\omega})$. Furthermore, it makes training more flexible and less sensitive to regularization weights with two different optimizing directions. As the two components gradually converge through regularized visibility estimation, a more accurate visibility $V(\mathbf x, \boldsymbol{\omega})$ can be obtained for decomposition (See in Fig. \ref{fig:adv-vis}).
We also apply the same regularized estimation strategy to the normal field learned in the normal optimizing stage to obtain a more accurate normal.

\par After incorporating regularized visibility estimation into inverse rendering, our final loss function in the decomposition stage is:
\begin{equation}
    \mathcal{L}=\lambda_{rgb}\mathcal{L}_{rgb}+\lambda_{sm}\mathcal{L}_{sm}+\lambda_{KL}\mathcal{L}_{KL},
\end{equation}
where $\mathcal L_{rgb}$ is the $\mathcal L_2$ loss between PBR rendering and gt,  $\mathcal{L}_{sm}$ is the same smooth $\mathcal L_1$ loss as Equ. \eqref{equ:normal} for albedo and roughness, and $\mathcal{L}_{KL}$ is the KL divergence loss on the latent code \textbf{z} of BRDF. See more in the supplementary materials. 

\begin{figure*}
    \centering
    \addtolength{\tabcolsep}{-6.5pt}
    \footnotesize{
        \setlength{\tabcolsep}{1pt} 
        \begin{tabular}{p{8.2pt}ccccccc}
            & Ours & Invrender & TensoIR & NeRO & NeRFactor & NVDiffrec & GT   \\
        \raisebox{30pt}{\rotatebox[origin=c]{90}{albedo}}&
             \includegraphics[width=0.13\textwidth]{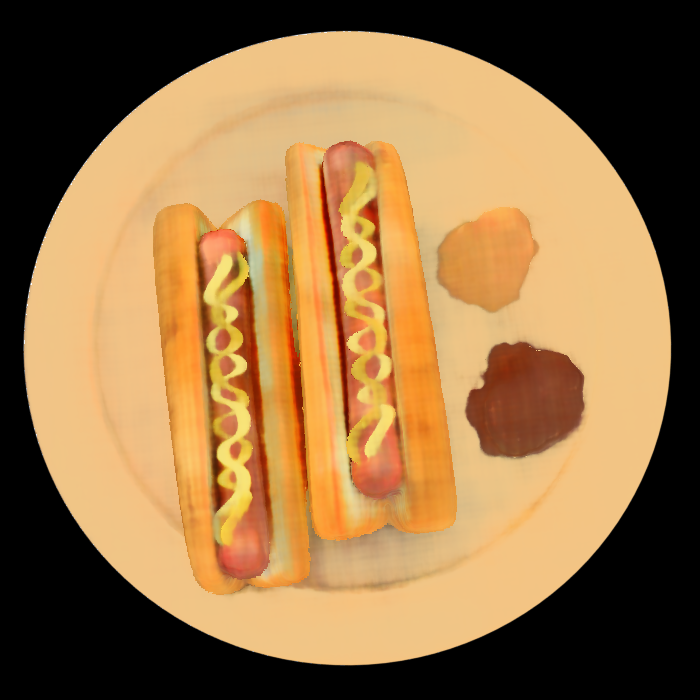} &
            \includegraphics[width=0.13\textwidth]{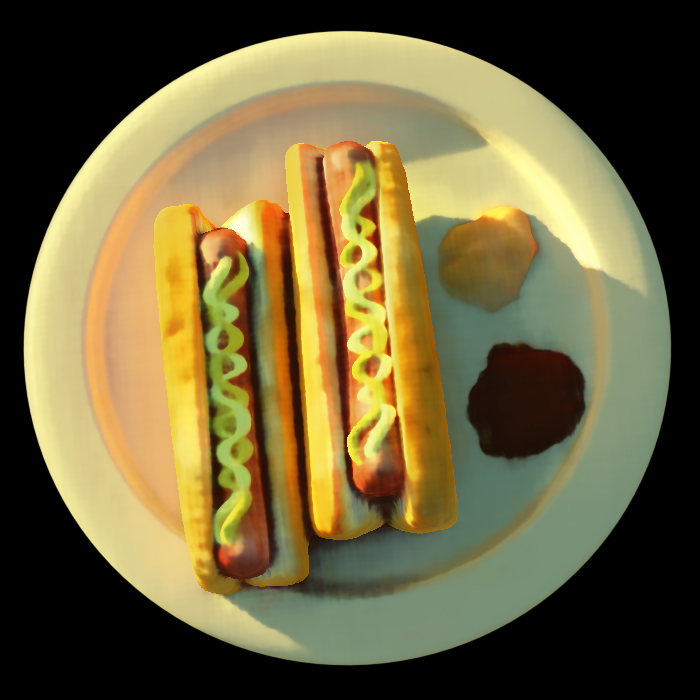} &
            \includegraphics[width=0.13\textwidth]{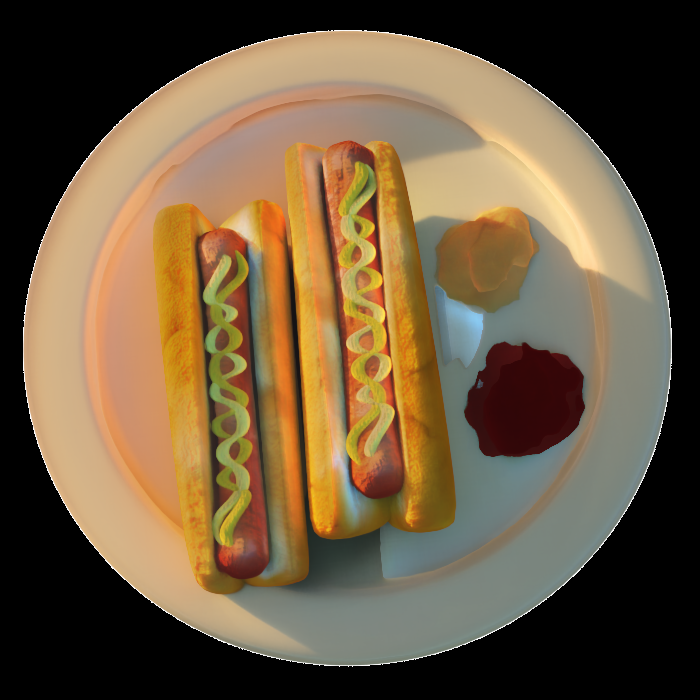} &
            \includegraphics[width=0.13\textwidth]{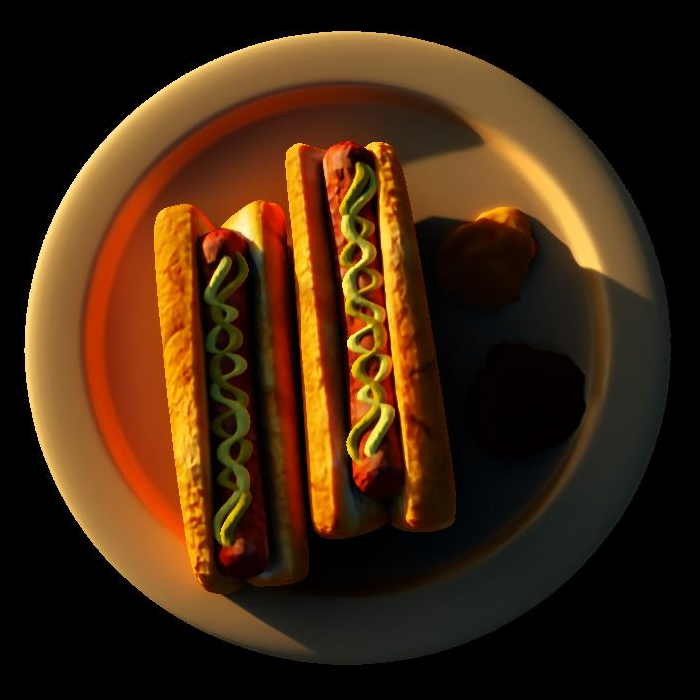} &
            \includegraphics[width=0.13\textwidth]{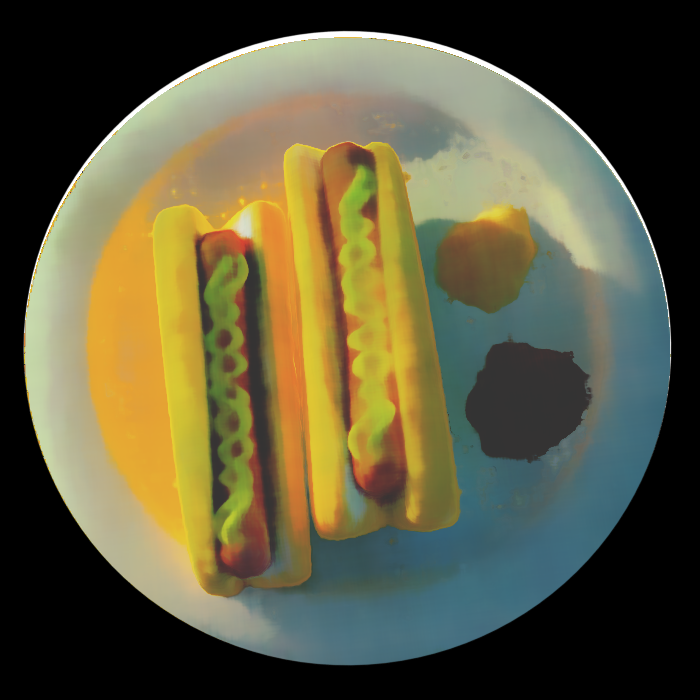} &
            \includegraphics[width=0.13\textwidth]{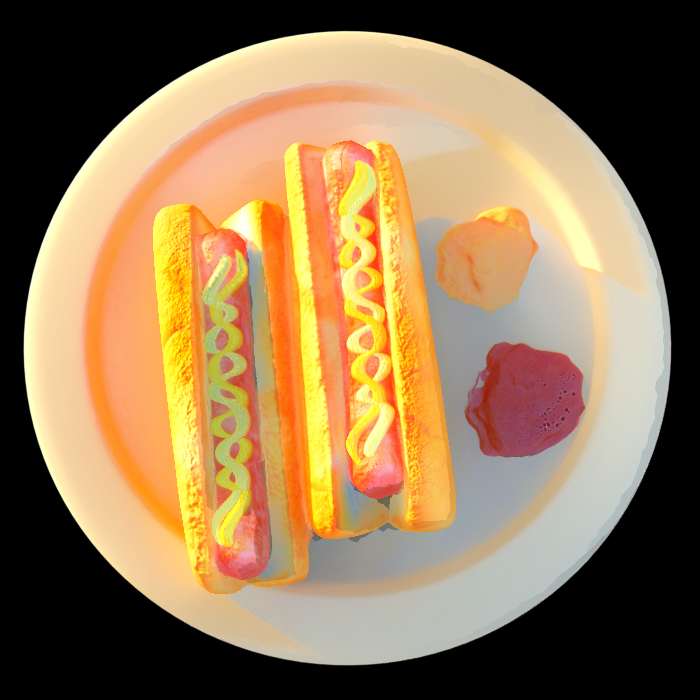} &
            \includegraphics[width=0.13\textwidth]{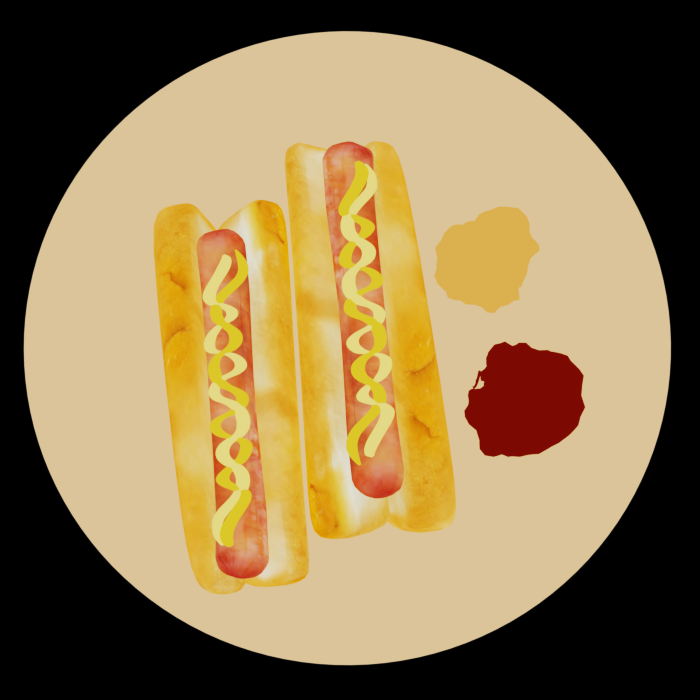}
             \\
        \raisebox{30pt}{\rotatebox[origin=c]{90}{roughness}}&
             \includegraphics[width=0.13\textwidth]{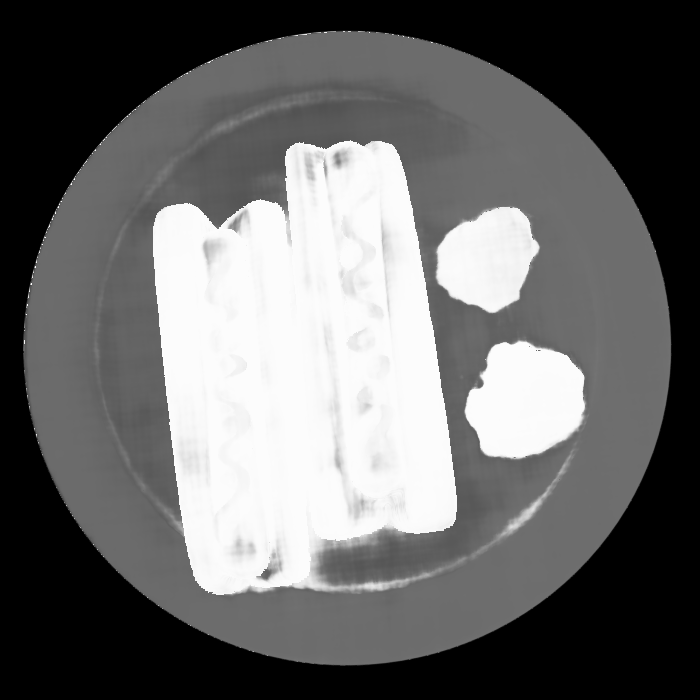} &
            \includegraphics[width=0.13\textwidth]{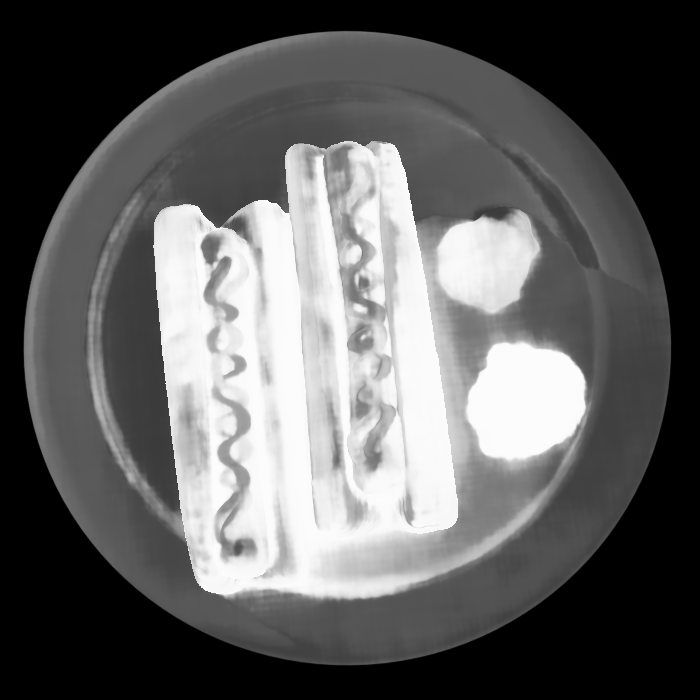} &
            \includegraphics[width=0.13\textwidth]{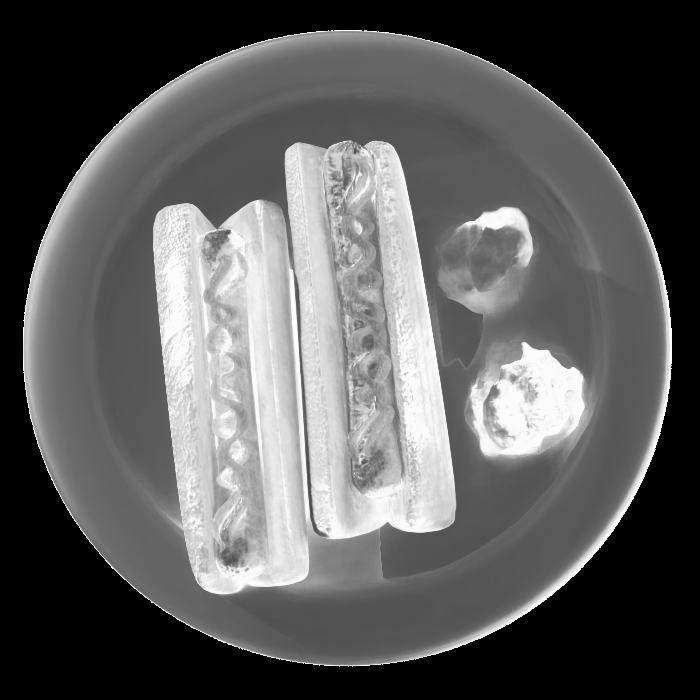} &
            \includegraphics[width=0.13\textwidth]{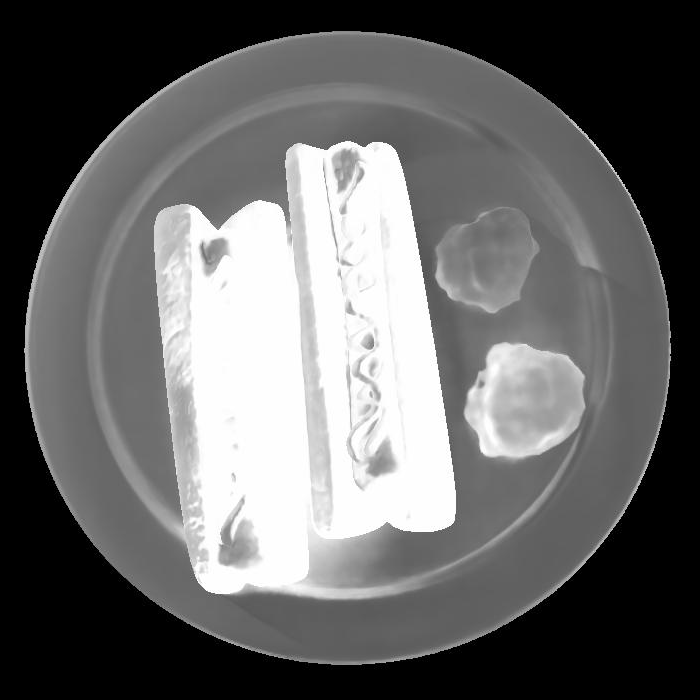} &
            \includegraphics[width=0.13\textwidth]{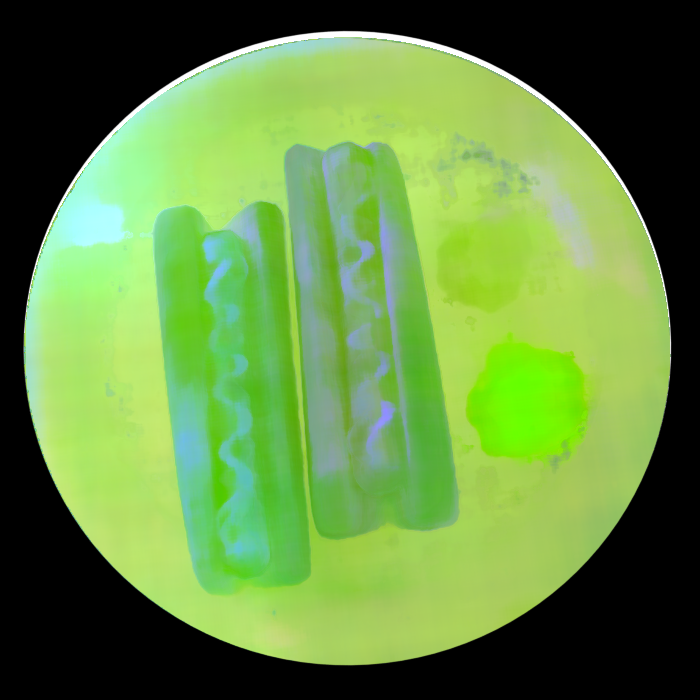} &
            \includegraphics[width=0.13\textwidth]{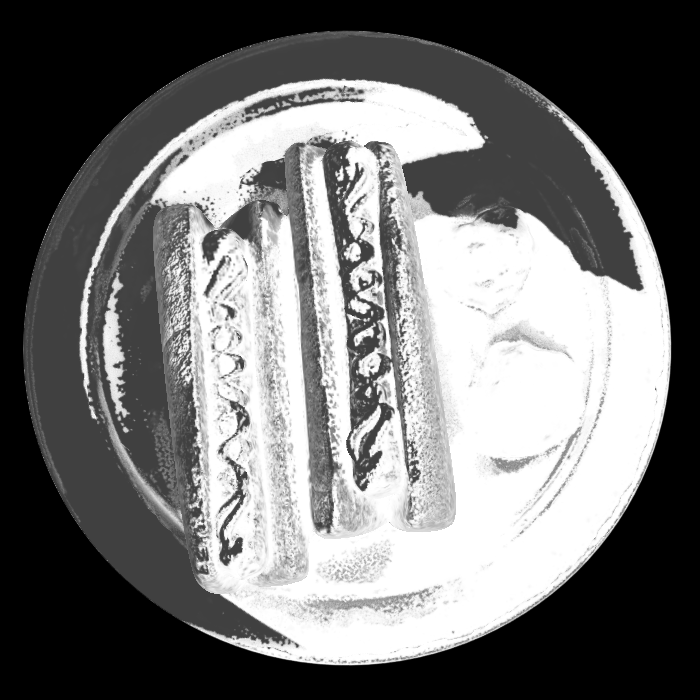} &
            \includegraphics[width=0.13\textwidth]{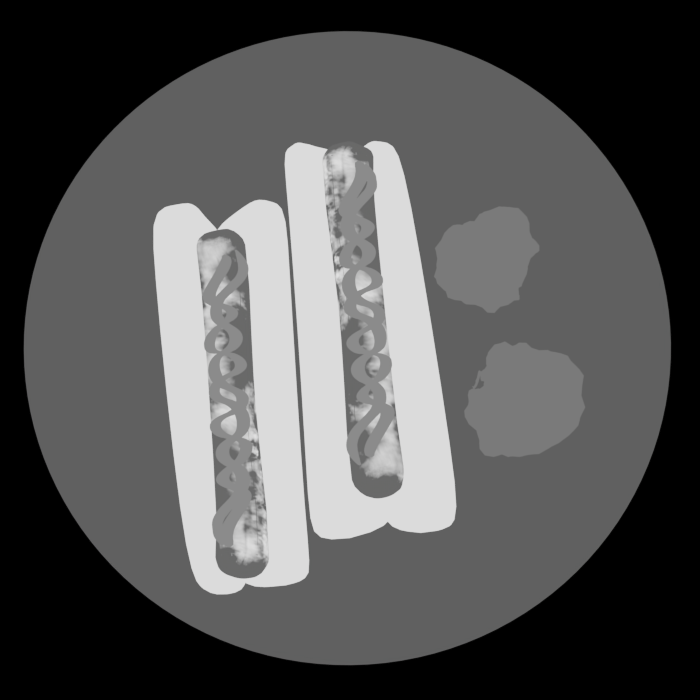}
            \\
        \raisebox{30pt}{\rotatebox[origin=c]{90}{albedo}}&
             \includegraphics[width=0.13\textwidth]{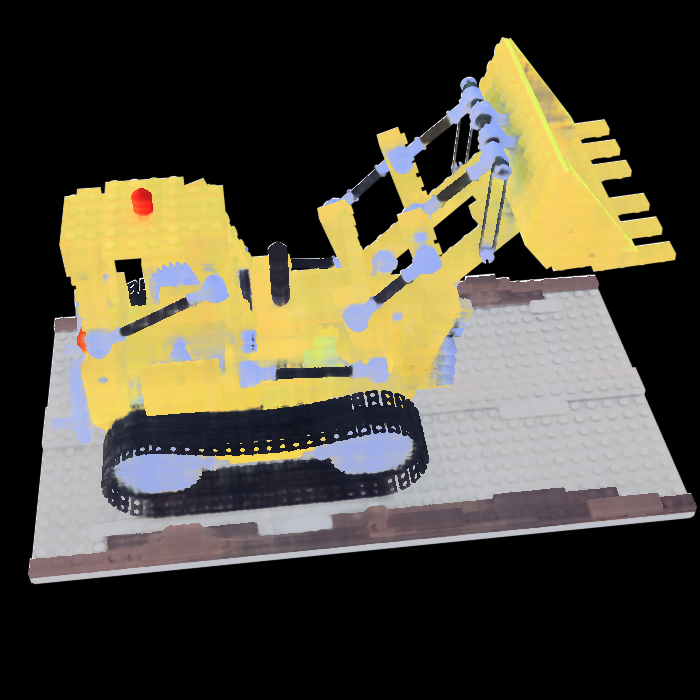} &
            \includegraphics[width=0.13\textwidth]{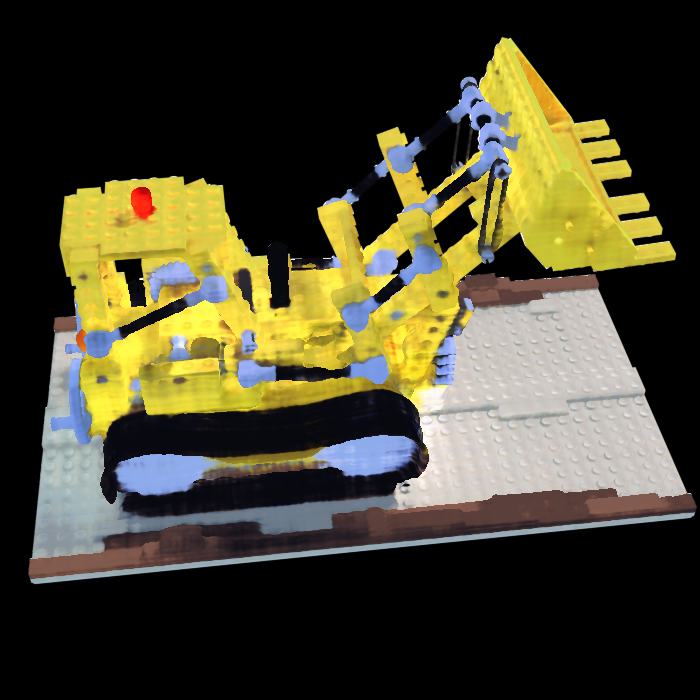} &
            \includegraphics[width=0.13\textwidth]{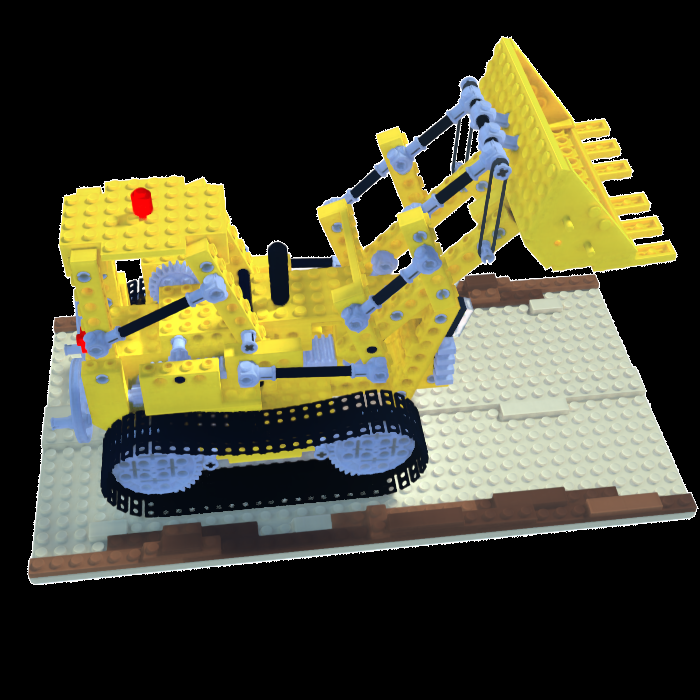} &
            \includegraphics[width=0.13\textwidth]{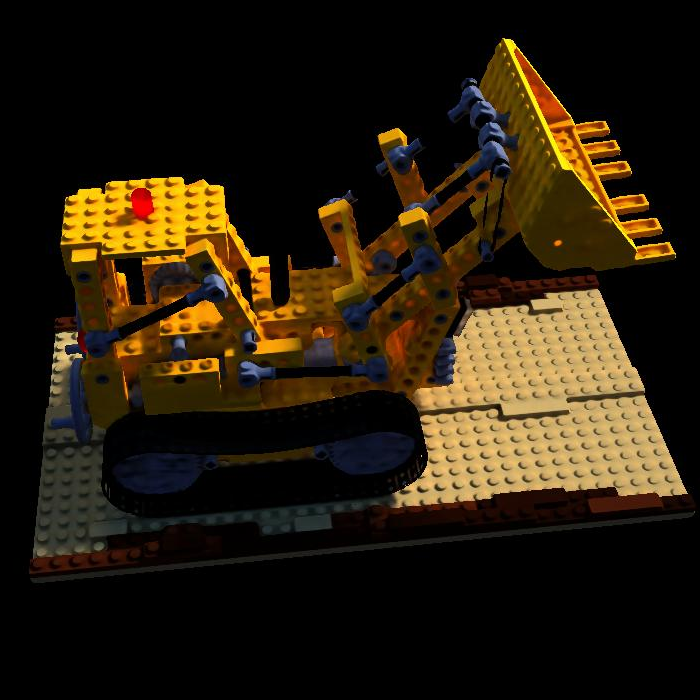} &
            \includegraphics[width=0.13\textwidth]{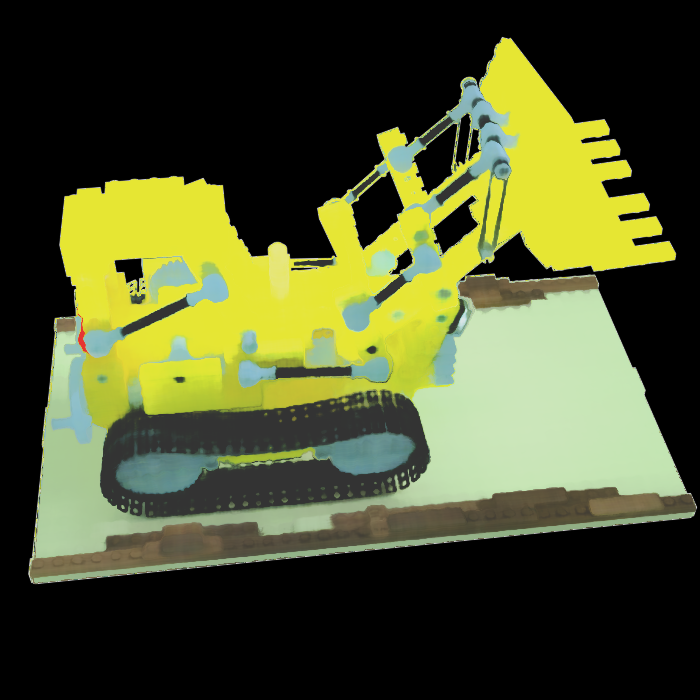} &
            \includegraphics[width=0.13\textwidth]{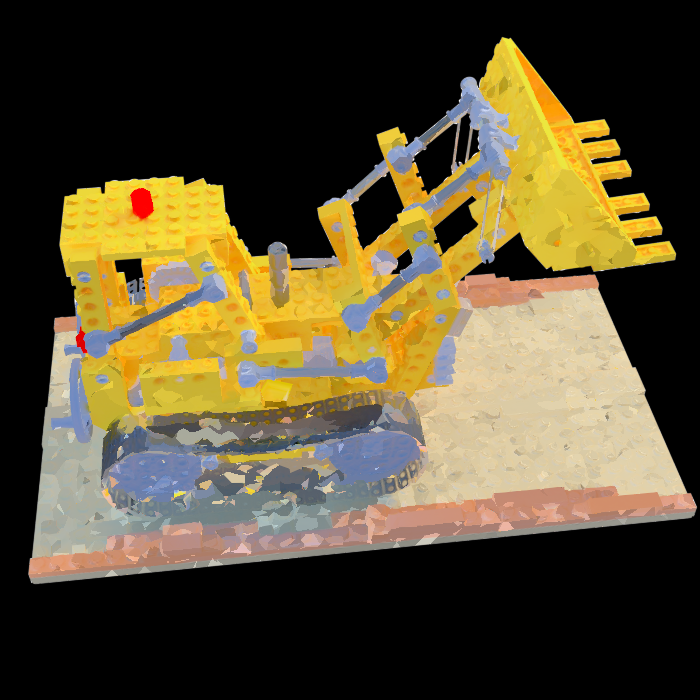} &
            \includegraphics[width=0.13\textwidth]{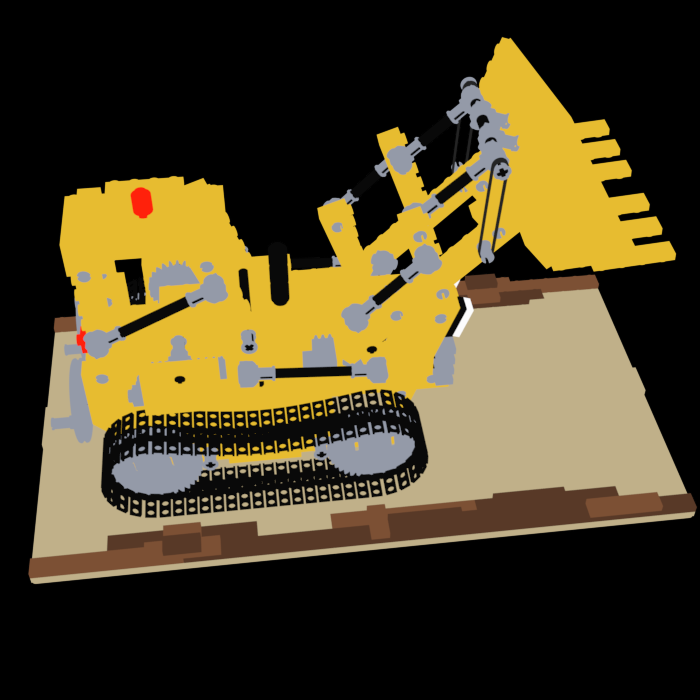}
            \\
        \raisebox{30pt}{\rotatebox[origin=c]{90}{roughness}}&
             \includegraphics[width=0.13\textwidth]{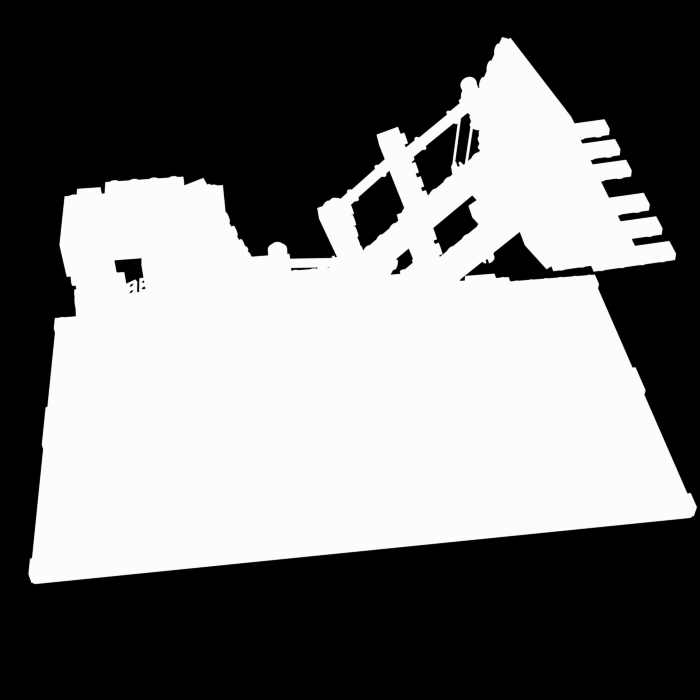} &
            \includegraphics[width=0.13\textwidth]{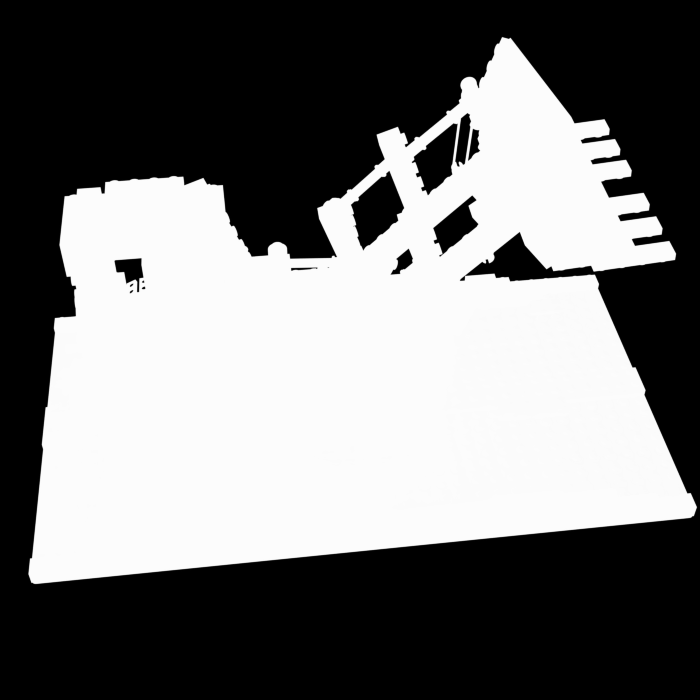} &
            \includegraphics[width=0.13\textwidth]{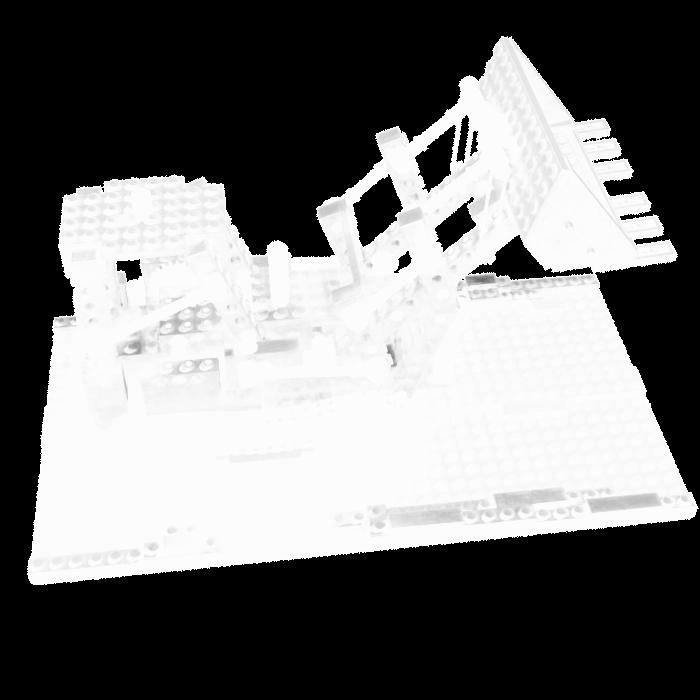} &
            \includegraphics[width=0.13\textwidth]{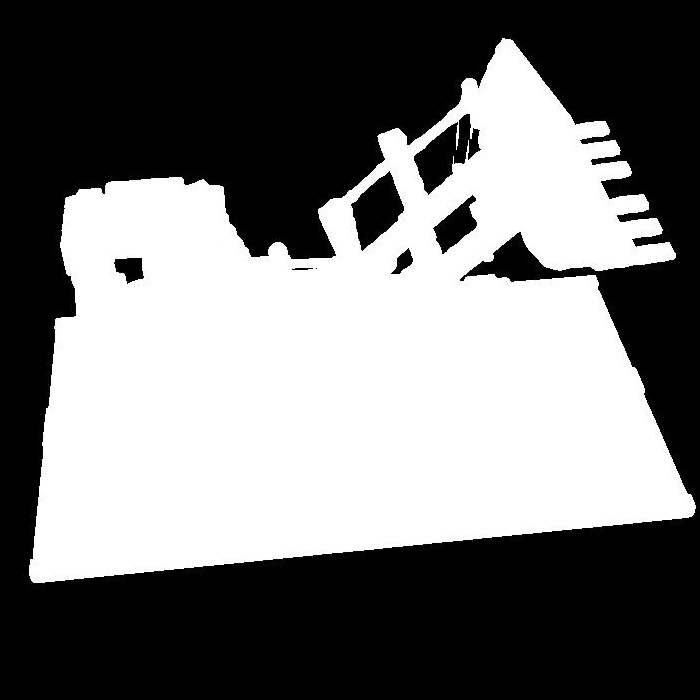} &
            \includegraphics[width=0.13\textwidth]{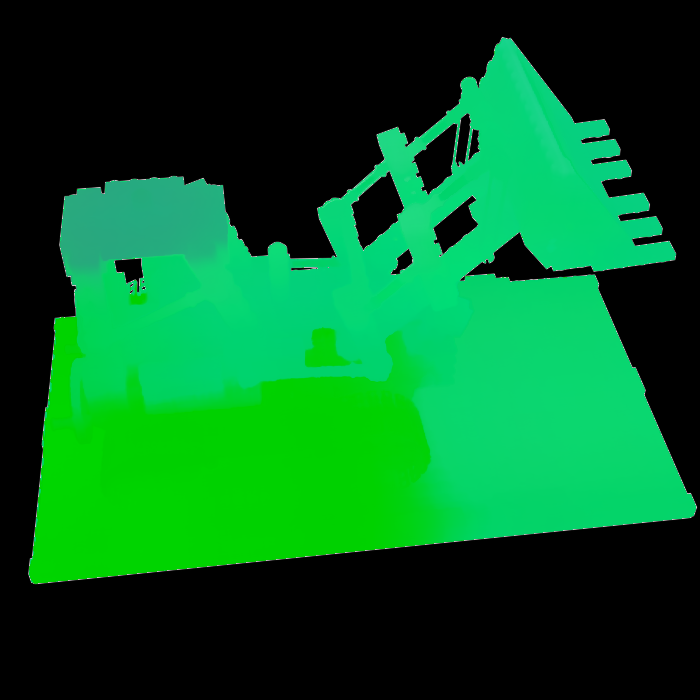} &
            \includegraphics[width=0.13\textwidth]{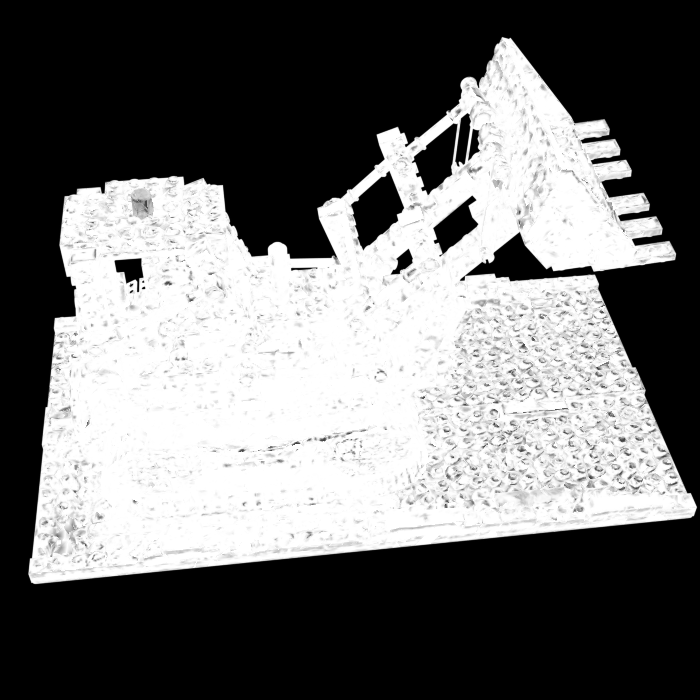} &
            \includegraphics[width=0.13\textwidth]{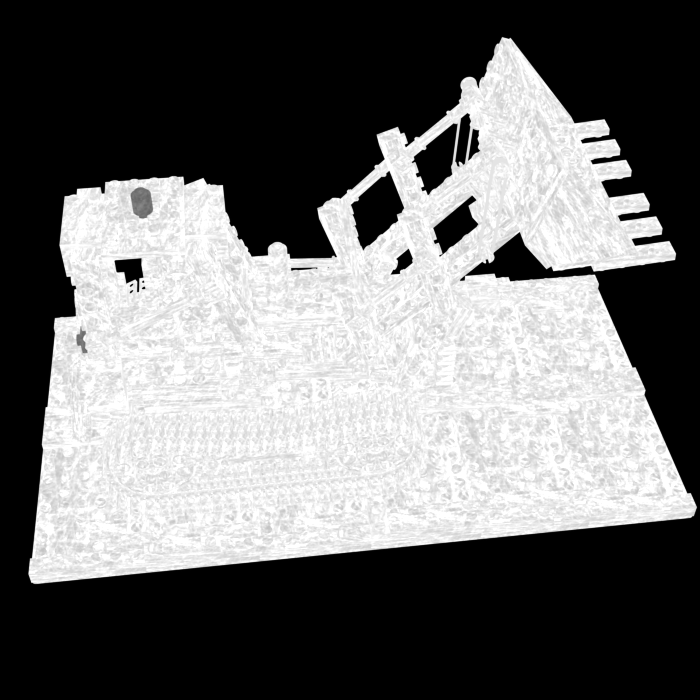}
        \end{tabular}
    }
\vspace{-5pt}
	\caption{\textbf{Comparisons to state-of-the-art methods.} 
We compare our method to several previous approaches, including InvRender \cite{zhang2022modeling}, TensoIR \cite{Jin2023TensoIR}, NeRO \cite{liu2023nero}, NeRFactor \cite{zhang2021nerfactor}, and NVDiffrec \cite{hasselgren2022nvdiffrecmc}. The results show that our method outperforms previous approaches in the accuracy of albedo and roughness generation. We should note that NeRFactor does not explicitly model roughness, so we represent it with the corresponding BRDF latent code. Furthermore, we train NVDiffrec from scratch rather than using a pre-trained mesh for initialization. 
}\label{fig:ogre}
\end{figure*}

%% file: sec/4_experiment.tex
\section{Experiments}
\par 
In this section, we present the experimental evaluation of our methods. To assess the effectiveness of our approach, we collect synthetic and real datasets from NeRF and NeuS \textbf{without any post-processing}. In addition, we use Blender to render our own datasets to further demonstrate the superiority of our methods. The collected datasets are used to evaluate the performance of our methods in terms of reconstruction accuracy and decomposition quality. 

\par Our model hyperparameters consisted of a batch size of 1024, with each stage trained for 100 epochs and 200k iterations for the NeuS training. For regularized visibility estimation, we initialized $\tilde Q$ for the first 5 epochs. The model was implemented in PyTorch and optimized with the Adam optimizer at a learning rate of $5e^{-4}$. All tests were conducted on a single Tesla V100 GPU with 32GB memory. The training time without NeuS is around 5 hours.

\subsection{Decomposition Results and Comparisons} \label{sec:exp-compare}
\par We evaluate the performance of our proposed methods by comparing them to some closely related inverse rendering methods that all decompose scenes under unknown illumination conditions: InvRender \cite{zhang2022modeling}, TensoIR \cite{Jin2023TensoIR}, NeRO \cite{liu2023nero}, NeRFactor \cite{zhang2021nerfactor}, and NVDiffrec \cite{munkberg2022extracting}.

\par 
As shown in Fig. \ref{fig:ogre}, our method removes shadows more effectively and produces cleaner albedo and roughness. 
In comparison with other methods, our method produces smoother results without losing details. Quantitative evaluations provided in Tab. \ref{tab:quantitative} show the precision of the albedo and roughness. The term "Log" refers to the use of sigmoid mapping instead of ACES during these evaluations. Note that since the albedo and roughness generated by each method have inconsistent hues, comparing PSNR is not as meaningful as comparing SSIM and LPIPS. We have shown more complete quantitative comparison results in the supplementary materials. Overall, our approach ensures satisfactory performance in both reconstruction and decomposition quality. 

\par
\begin{figure}
    \centering
    \footnotesize{
        \setlength{\tabcolsep}{0.5pt} 
        \begin{tabular}{ccccc}
        (a) nvdiffrec & (b) InvRender & (c) TensoIR & (d) ours & (e) gt \\
        \includegraphics[width=0.095\textwidth]{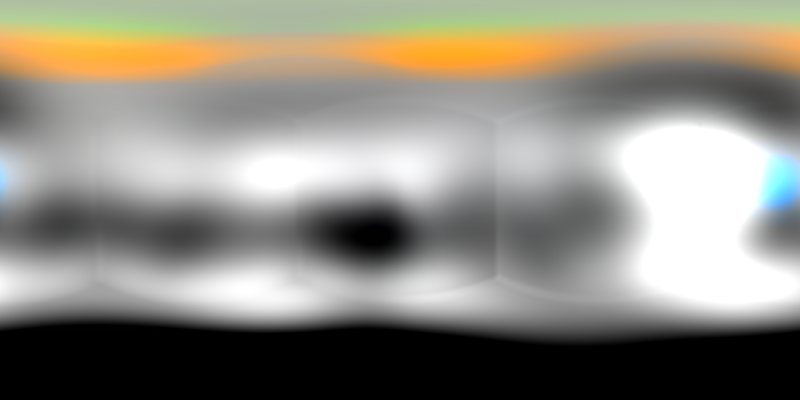} &
        \includegraphics[width=0.095\textwidth]{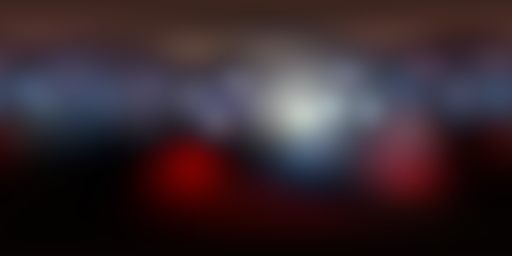} &
        \includegraphics[width=0.095\textwidth]{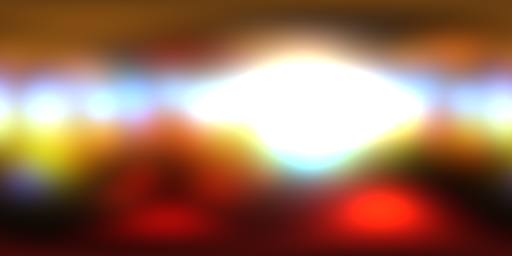} &
        \includegraphics[width=0.095\textwidth]{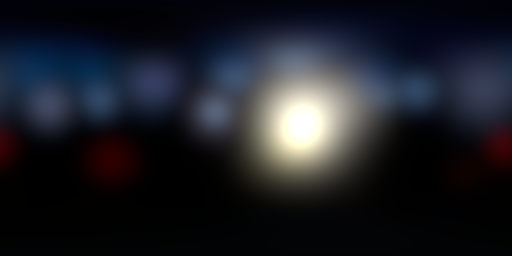} &
        \includegraphics[width=0.095\textwidth]{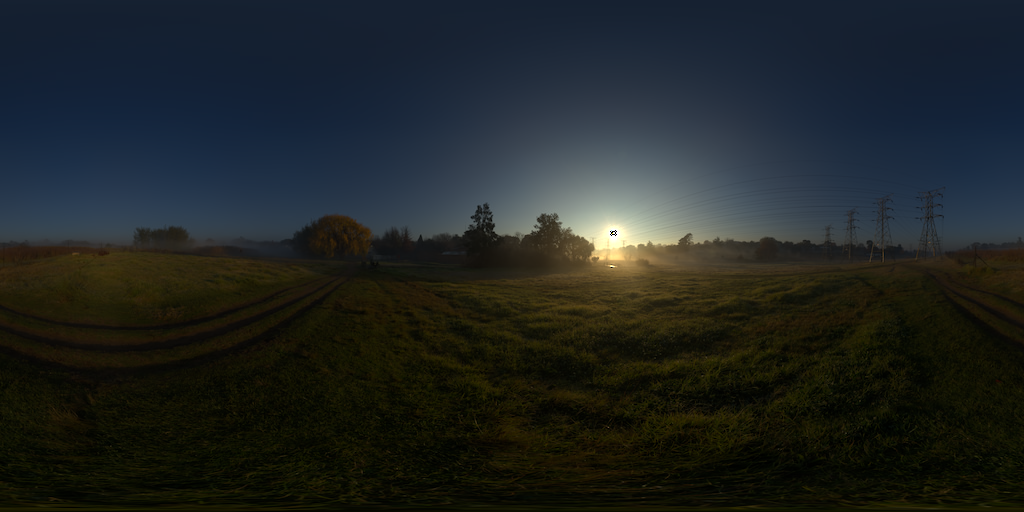} \\
        \includegraphics[width=0.095\textwidth]{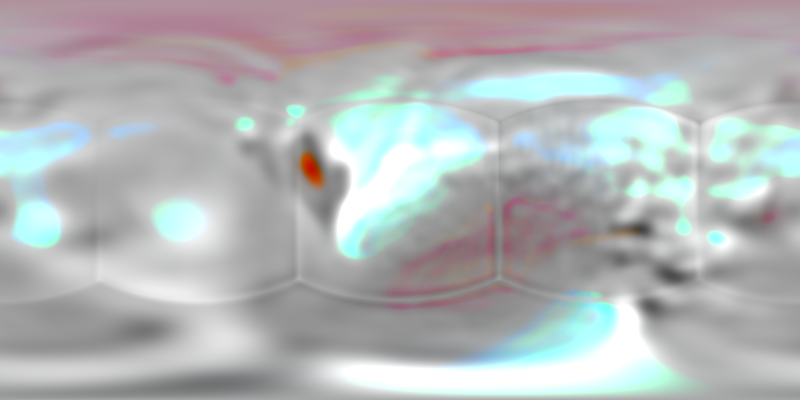} &
        \includegraphics[width=0.095\textwidth]{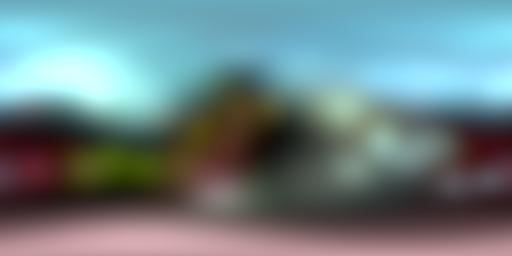} &
        \includegraphics[width=0.095\textwidth]{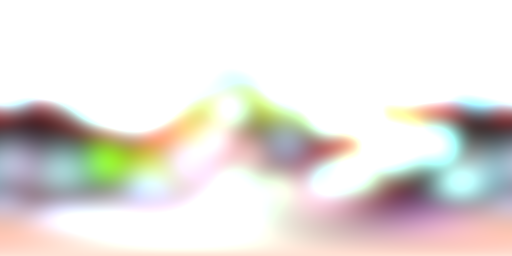} &
        \includegraphics[width=0.095\textwidth]{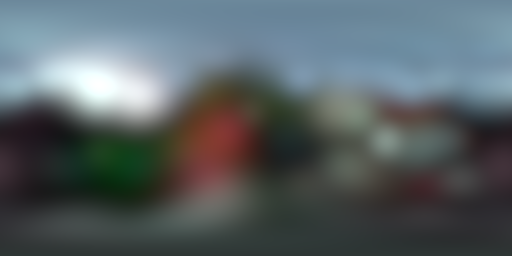} &
        \includegraphics[width=0.095\textwidth]{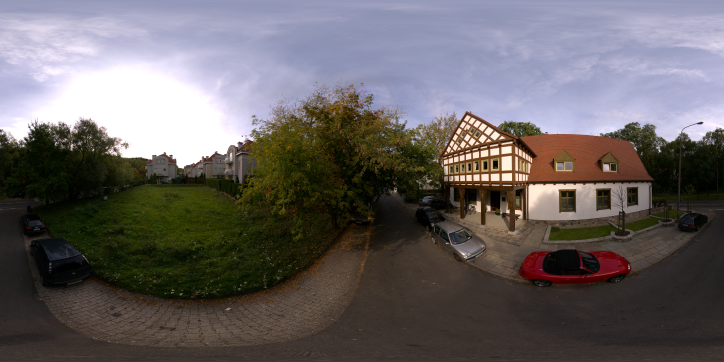} \\
        \end{tabular}
    }
\vspace{-5pt}
	\caption{\textbf{Reconstructed environment light.} We present the reconstructed environment maps of the hotdog (upper) and the helmet (bottom). Compared to existing approaches, our method produces superior environment light, indicating its effectiveness in modeling non-linearly mapped radiance fields.}\label{fig:envmaps}
\end{figure}

 The reconstructed environment light maps are shown in Fig. \ref{fig:envmaps}, where the lighting is shifted by a constant on exposure to visualize the HDR values. 
Our method can more accurately estimate the position of the light source and generate higher and more precise light intensity in HDR environment light.

\par We also perform experiments on real-world captured scenes. As illustrated in the supplementary materials, our method is capable of decomposing real-world objects into plausible geometry, albedo, and roughness. Decomposed components can be used to aid in downstream tasks such as realistically relighting real-world scenes under arbitrary lighting conditions for free-viewpoint navigation.

\subsection{Ablation Studies}
\label{sec:exp-ablation}
\par
\begin{figure}
    \centering
    \addtolength{\tabcolsep}{-6.5pt}
    \footnotesize{
        \setlength{\tabcolsep}{1.5pt} 
        \begin{tabular}{cccccc}
        Ours & Log Tone & No ACES & No RVE & InvRender & GT \\ 
        \includegraphics[width=0.071\textwidth]{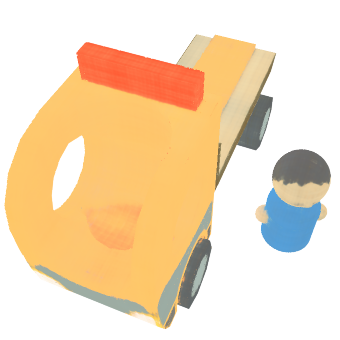} &
        \includegraphics[width=0.071\textwidth]{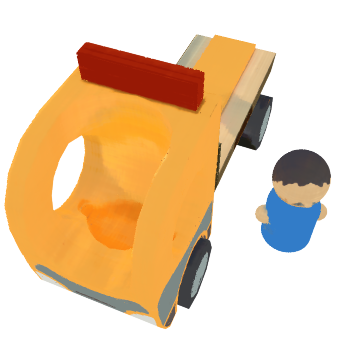} &
        \includegraphics[width=0.071\textwidth]{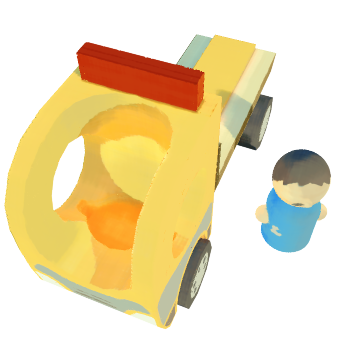} &
        \includegraphics[width=0.071\textwidth]{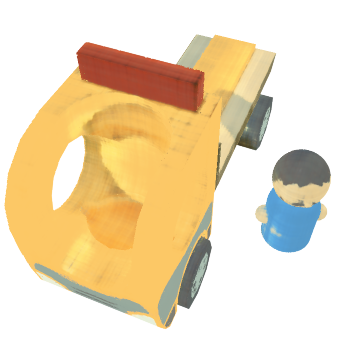} &
        \includegraphics[width=0.071\textwidth]{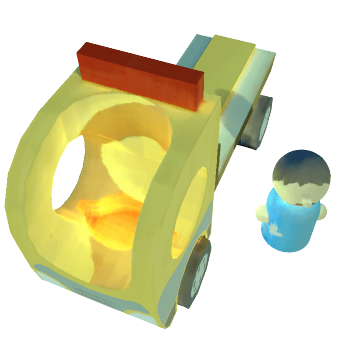} &
        \includegraphics[width=0.071\textwidth]{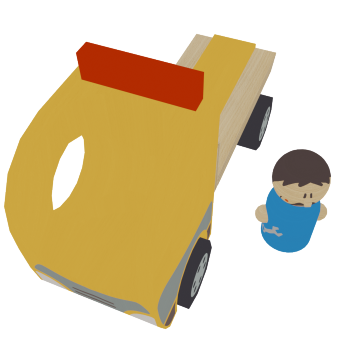} \\
        \includegraphics[width=0.071\textwidth]{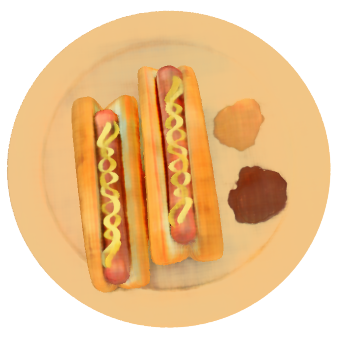} &
        \includegraphics[width=0.071\textwidth]{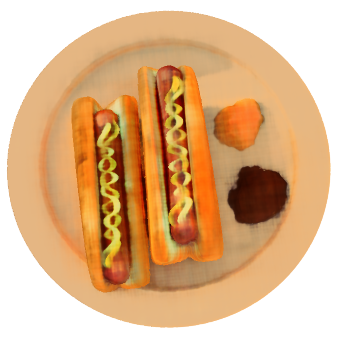} &
        \includegraphics[width=0.071\textwidth]{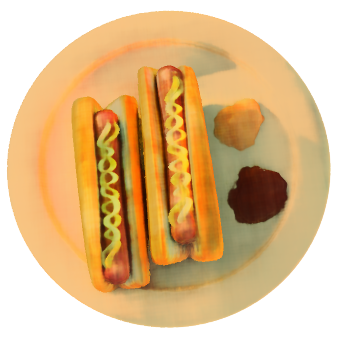} &
        \includegraphics[width=0.071\textwidth]{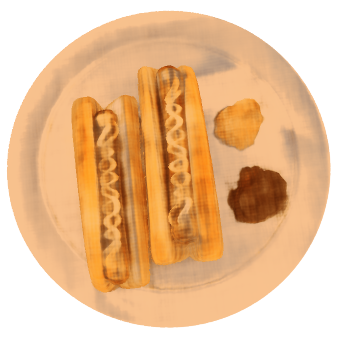} &
        \includegraphics[width=0.071\textwidth]{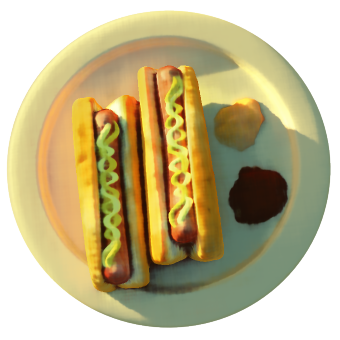} &
        \includegraphics[width=0.071\textwidth]{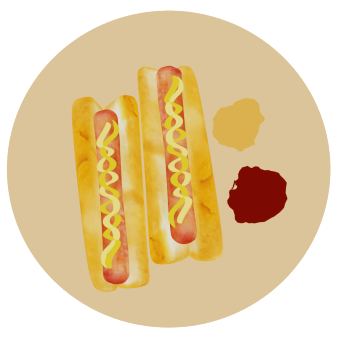} \\ 
        \end{tabular}
    }
\vspace{-5pt}
    \caption{\textbf{Ablation Study.} 
    We conduct ablation experiments on the key components in the Scene Decomposition Stage, with "Log Tone" referring to the use of sigmoid tone mapping as opposed to ACES tone mapping. The ablation results not only validate the effectiveness of ACES non-linear tone mapping but also emphasize the critical importance of each component in our proposed framework for attaining high-quality albedo.}\label{fig:ablation}
\end{figure}

 We perform an ablation study to analyze the importance of the key components in our proposed \emph{SIRe-IR} methods. As illustrated in Fig. \ref{fig:ablation}, we observe that InvRender produces poor decomposition results under intense lighting conditions. 
This is due to the lack of non-linear mapping for the radiance fields in InvRender, which inhibits it from effectively balancing the light intensity, thereby leaving residual shadows and indirect illumination.
In the absence of ACES tone mapping, our method is unable to eliminate both shadows and indirect illumination. Without regularized visibility estimation, the training process is frequently unstable and the resulting albedo may contain shadows in the corners. The "Log Tone" result indicates that ACES offers a more effective non-linear mapping than the sigmoid function within our framework. Finally, our full method can correctly decompose light into albedo and roughness, resulting in the best performance.

\begin{table*}[]
\centering
\resizebox{0.95\textwidth}{!}{
\begin{tabular}{r|cccc|cccc|cccc}
\hline
\multicolumn{1}{c}{} & \multicolumn{4}{|c|}{Albedo (SSIM) $\uparrow$} & \multicolumn{4}{c|}{Albedo (LPIPS) $\downarrow$} & \multicolumn{4}{c}{Roughness (SSIM) $\uparrow$} \\ \hline
Scene    & Hotdog & Lego & Helmet & Chess & Hotdog & Lego  & Helmet & Chess & Hotdog & Lego  & Helmet & Chess              \\ \hline
ours      & \cellcolor{yzybest}0.9283 & \cellcolor{yzybest}0.8941  & \cellcolor{yzybest}0.9461 & \cellcolor{yzybest}0.8535  & \cellcolor{yzybest}0.1296 & \cellcolor{yzybest}0.1361      & \cellcolor{yzysecond}0.1160    & \cellcolor{yzybest}0.1898 & \cellcolor{yzybest}0.9190 & \cellcolor{yzybest}0.8823 & \cellcolor{yzythird}0.8583 & \cellcolor{yzythird}0.8101   \\
ours-Log    & \cellcolor{yzythird}0.9046          & 0.8863                & \cellcolor{yzythird}0.9276          & \cellcolor{yzythird}0.8354    & \cellcolor{yzysecond}0.1703 & \cellcolor{yzythird}0.1392             & \cellcolor{yzybest}0.1071 & \cellcolor{yzysecond}0.2013 & 0.8248 & \cellcolor{yzythird}0.8805 & 0.8356 & 0.6837           \\
no aces    & 0.9041          & \cellcolor{yzysecond}0.8933           & 0.9056          & \cellcolor{yzysecond}0.8524   & \cellcolor{yzythird}0.1690          & 0.1402                & 0.1542          & \cellcolor{yzythird}0.2048 & \cellcolor{yzythird}0.8870 & 0.8803 & \cellcolor{yzysecond}0.8819 & 0.6374               \\
no reg-estim    & \cellcolor{yzysecond}0.9157    & \cellcolor{yzythird}0.8871                & \cellcolor{yzysecond}0.9383    & 0.8198    & 0.1715          & \cellcolor{yzysecond}0.1376         & \cellcolor{yzythird}0.1242          & 0.2382 & \cellcolor{yzysecond}0.8985 & 0.8804 & 0.8441 & \cellcolor{yzybest}0.8779          \\ \hline
InvRender  & 0.8762          & 0.8833                & 0.9115          & 0.8126      & 0.2275          & 0.1551                 & 0.1509          & 0.2259 & 0.8842 & \cellcolor{yzysecond}0.8807 & \cellcolor{yzybest}0.8900 & 0.7083  \\
nvdiffrec & 0.8377          & 0.7872          & 0.7859          & 0.7363    & 0.3649          & 0.2785         & 0.3454          & 0.4060 & 0.7351 & 0.8148 & 0.7796 & \cellcolor{yzysecond}0.8194    \\
nerfactor & 0.8238          & 0.8386              & -               & -   & 0.3318          & 0.2208        & -         & - & - & - & - & -                     \\ \hline
\end{tabular}}
\caption{\label{tab:quantitative} \textbf{Quantitative evaluations.} 
We present the results of four synthetic scenes. We color each cell as \colorbox{yzybest}{best}, \colorbox{yzysecond}{second best}, and \colorbox{yzythird}{third best}. Due to specific component naming rules in the Blender scene, we only compared datasets that NeRFactor supports. In summary, our method can produce high-quality albedo and roughness while maintaining the reconstruction fidelity. See more in the supplementary materials. }

\end{table*}

\subsection{Application}
\label{sec:application}
\paragraph{De-shadowing.}
\par
\begin{figure}
    \centering
    \footnotesize{
        \setlength{\tabcolsep}{7pt} 
        \begin{tabular}{ccc}
        \includegraphics[width=0.12\textwidth]{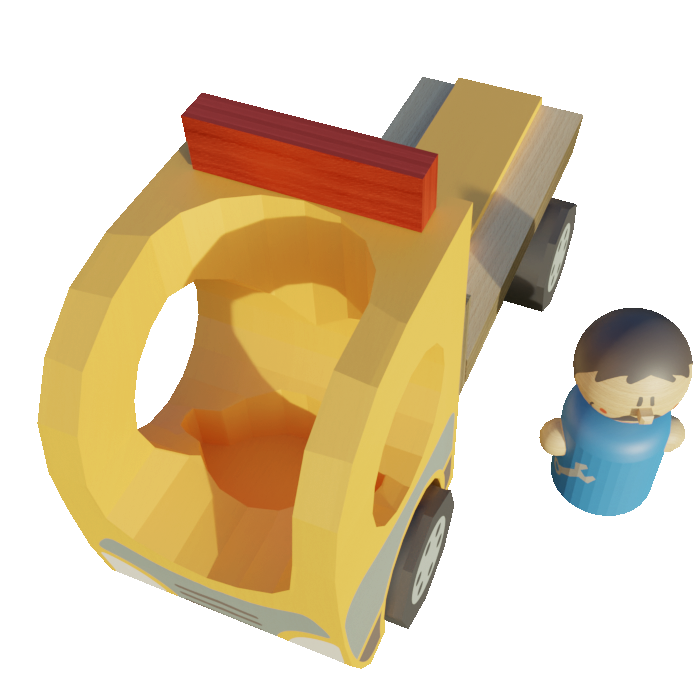} &
        \includegraphics[width=0.12\textwidth]{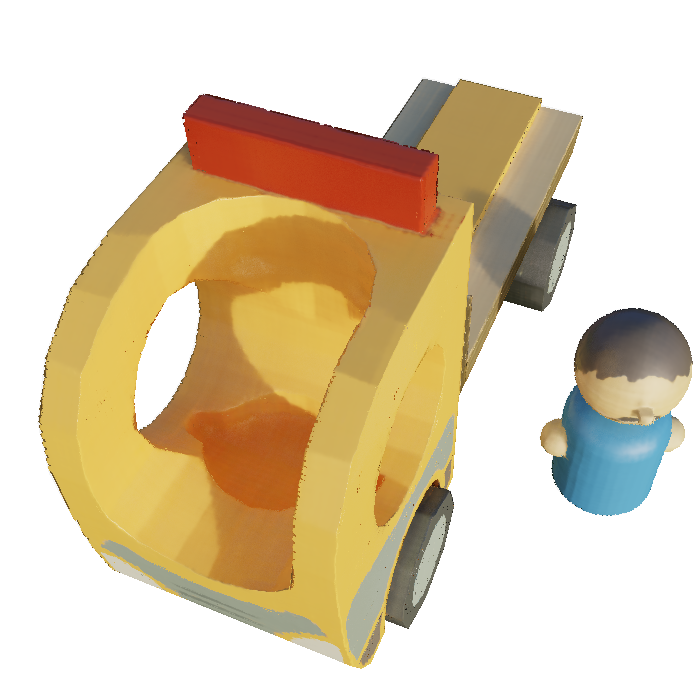} &
        \includegraphics[width=0.12\textwidth]{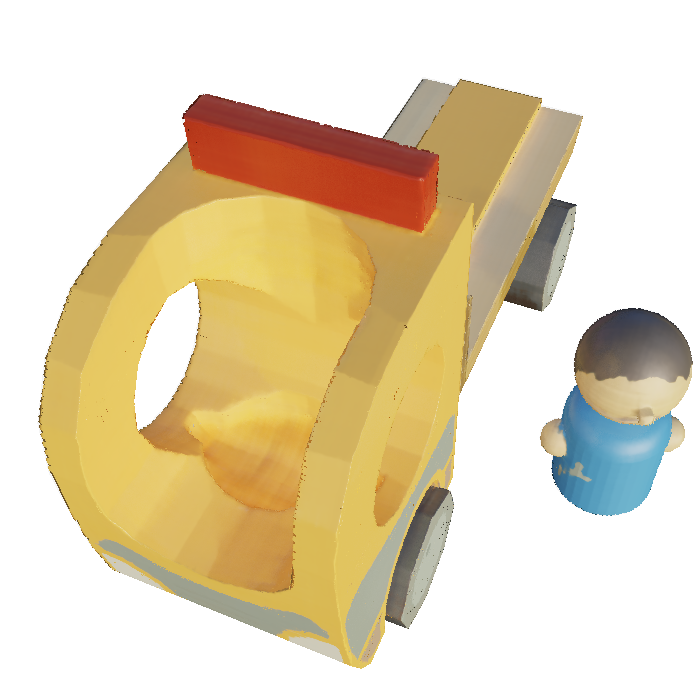} \\
        \includegraphics[width=0.12\textwidth]{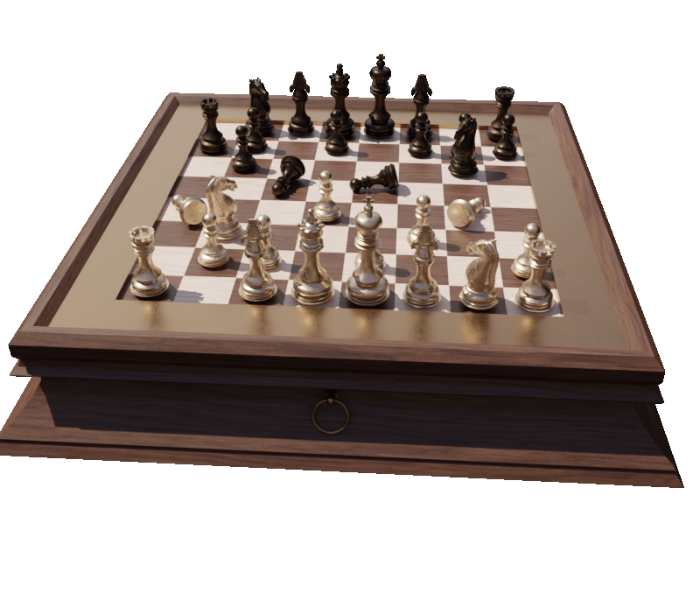} &
        \includegraphics[width=0.12\textwidth]{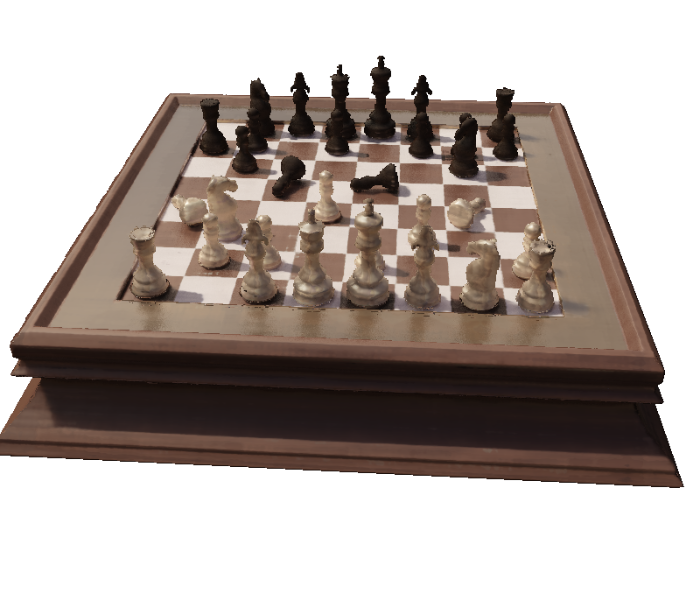} &
        \includegraphics[width=0.12\textwidth]{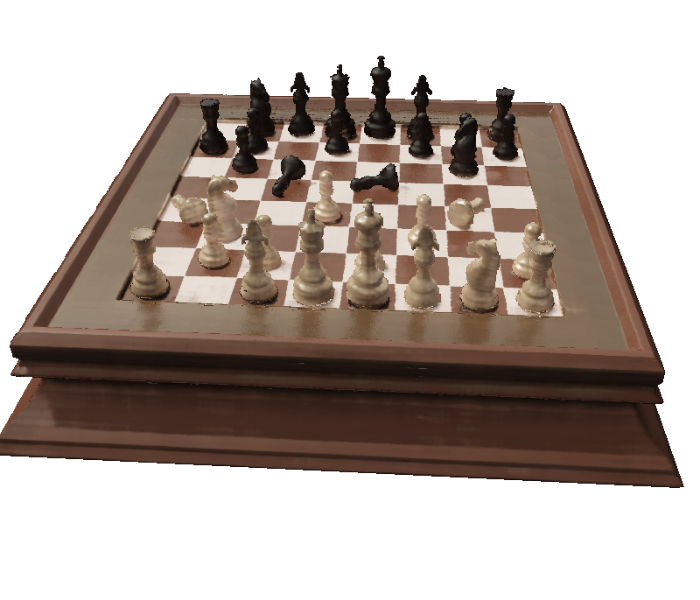} \\ 
        (a) Input View & (b) Rendering & (c) Deshadow \\ 
        \end{tabular}
    }
    \caption{\textbf{De-shadowing.} We showcase the effectiveness of our method in de-shadowing applications. Given an input image captured from a specific viewpoint (a), our proposed approach can accurately remove shadows caused by direct light occlusion (c) even in challenging scenes with intense illumination, while preserving high-fidelity in the re-rendered results (b). }
    \label{fig:deshadow}
\end{figure}

 De-shadowing is a challenging task in the field of inverse rendering, often requiring strong priors and large data-driven models. Our proposed method correctly understands various lighting effects and is capable of effectively eliminating strong and irregular shadows, particularly in scenes with intense lighting.
As shown in Fig. \ref{fig:deshadow}, by setting the visibility ratio $\eta$ to 1, we remove the shadowed portions caused by direct light occlusion. It should be noted that our method \textbf{cannot remove the areas with reflections and the dark regions caused by the backlighting phenomenon}. But this also to some extent demonstrates the accuracy of our method's visibility. These results also demonstrate the ability of our model to accurately identify and remove unwanted shadows.

\begin{figure}
    \centering
    \footnotesize{
        \setlength{\tabcolsep}{1pt} 
        \begin{tabular}{ccc}
        \includegraphics[width=0.145\textwidth]{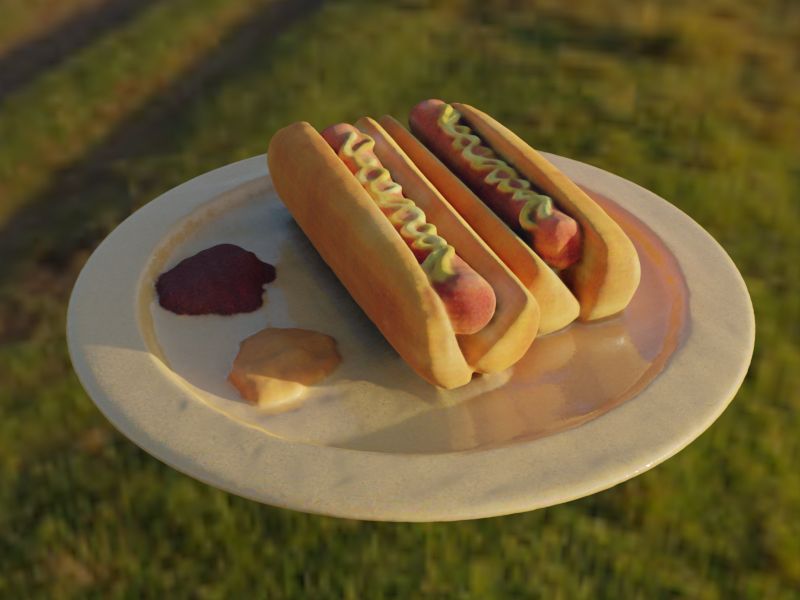} &
        \includegraphics[width=0.145\textwidth]{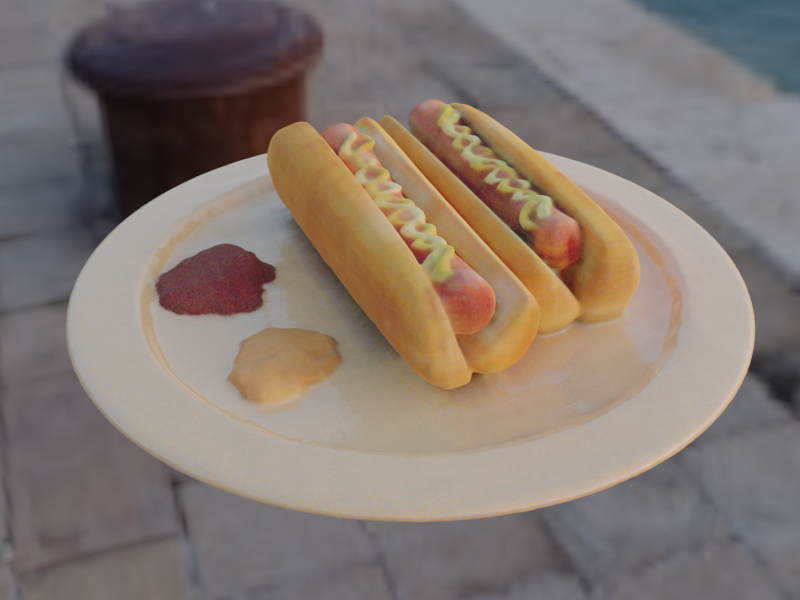} &
        \includegraphics[width=0.145\textwidth]{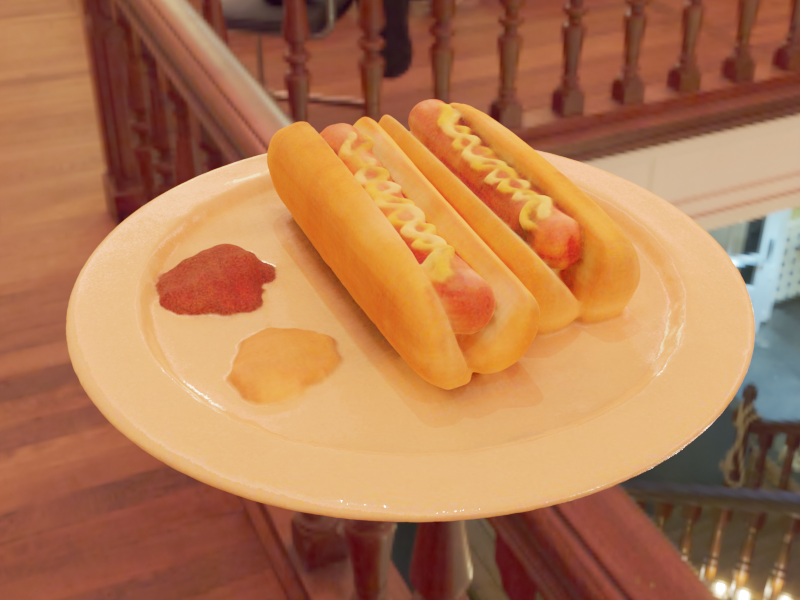} \\
        \includegraphics[width=0.145\textwidth]{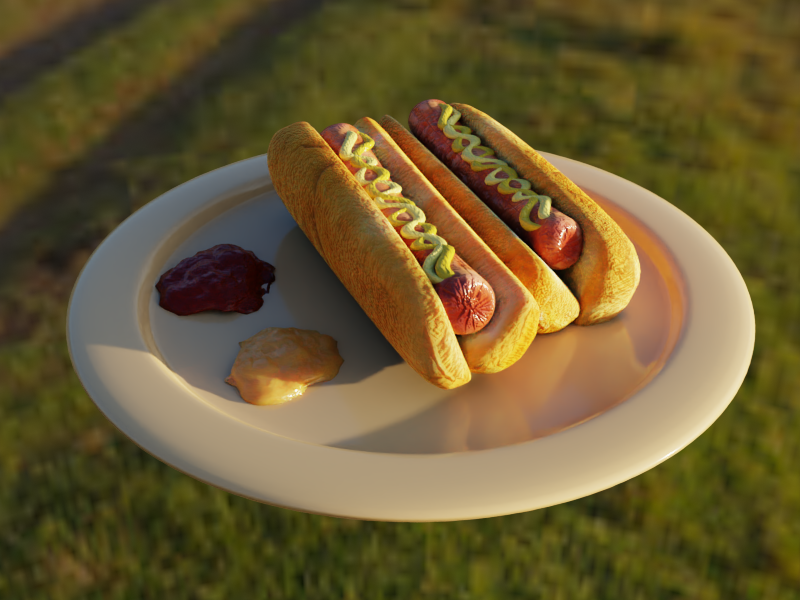} &
        \includegraphics[width=0.145\textwidth]{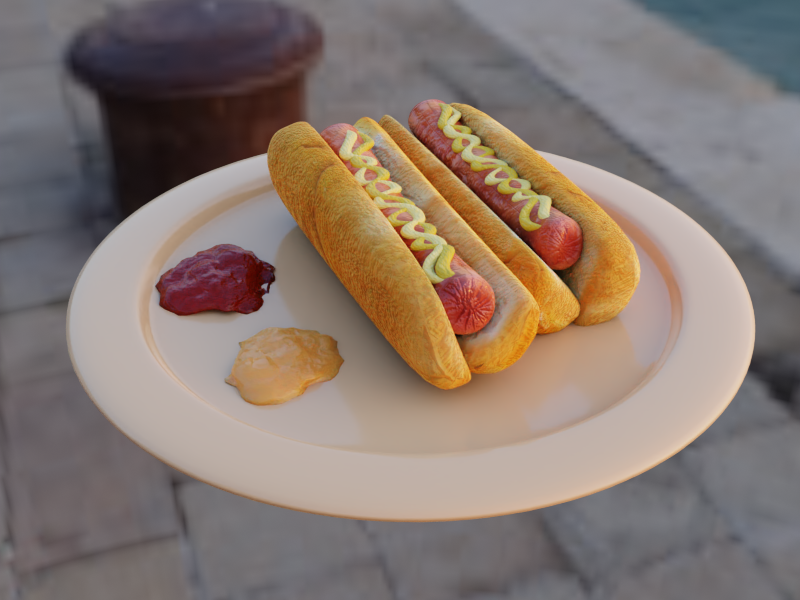} &
        \includegraphics[width=0.145\textwidth]{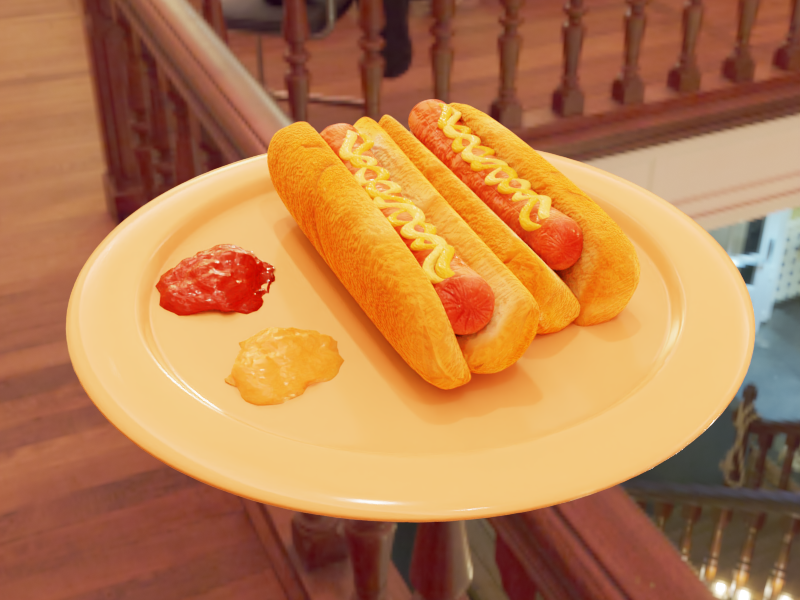} \\ 
        (a) Light 0 & (b) Light 1 & (c) Light 2 \\ 
        \end{tabular}
    }
    \caption{\textbf{Relighting.} We employ Blender to relight the extracted mesh, where the first row denotes our predictions, and the second row represents GT. Our approach allows for relighting the scene with decomposed components, providing flexibility to modify the lighting conditions as desired.}
    \label{fig:relight}
\end{figure}

\paragraph{Relighting.}
\par To demonstrate the practical utility of the materials from our method, we conducted relighting experiments. 
As shown in Fig. \ref{fig:relight}, it illustrates that our decomposition results can be accurately relighted in various lighting environments without shadow or illumination artifacts.

%% file: sec/5_conclusion.tex
\section{Conclusions and Discussions} 
\par We presented a novel inverse rendering framework for extracting high-quality albedo and roughness by removing shadows and indirect illumination. 
The key innovation lies in the use of non-linear mapping (ACES tone mapping) for illumination, which eliminates shadows and indirect illumination at the same time. In addition, masked indirect illumination and regularized visibility estimation are employed to ensure the high quality of decomposition. Experiment results on both synthetic and real-world data show that our full framework outperforms previous work in eliminating shadows and indirect illumination in PBR materials. Furthermore, scene components such as albedo, roughness, normal, and environment light produced in our method can be directly used in the traditional render pipeline.

\par Currently, the proposed method has some limitations. First, areas with strong reflections pose a significant challenge for accurate processing, which leads to artifacts in the corresponding regions in albedo and normal. Second, non-solid, translucent, and thin objects cannot be correctly handled due to the limitations of NeuS. Third, the employment of SGs to model both direct and indirect lighting presents challenges in dealing with anisotropic objects, consequently leading to our method's deficiency in incorporating the metallic learnable parameters present in the Disney BRDF model. Finally, we have not considered scenes with dynamic lighting. We can draw inspiration from \cite{martin2021nerf, sun2022neural} in the future work.

%% file: sec/X_supple.tex
\appendix

\section{Overview}
This supplementary document provides some implementation details and further results that accompany the paper. 

\begin{itemize}
    \item Section~\ref{sec:detail} introduces more details of our approach.
    \item Section~\ref{sec: add-results} provides additional results, including more visualizations and results on more datasets.

\end{itemize}

\begin{figure*}
    \centering
    \addtolength{\tabcolsep}{-6.5pt}
    \footnotesize{
        \setlength{\tabcolsep}{1pt} 
        \begin{tabular}{lccccccc}
        \raisebox{32pt}{\rotatebox[origin=c]{90}{Hotdog}}&
        \includegraphics[width=0.14\textwidth]{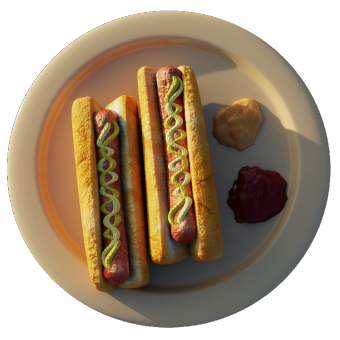} &
        \includegraphics[width=0.14\textwidth]{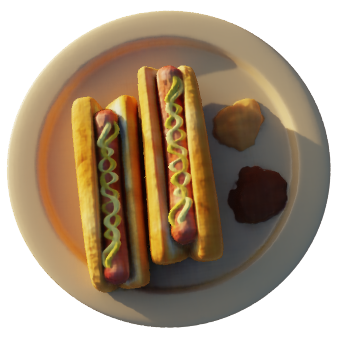} &
        \includegraphics[width=0.14\textwidth]{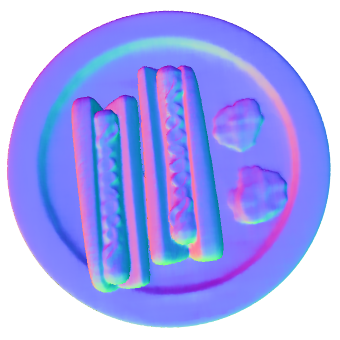} &
        \includegraphics[width=0.14\textwidth]{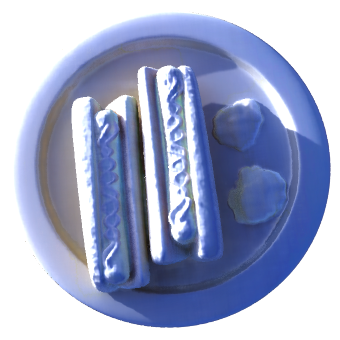} &
        \includegraphics[width=0.14\textwidth]{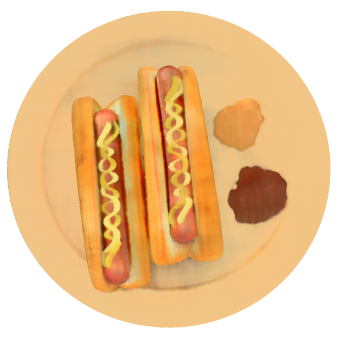} &
        \includegraphics[width=0.14\textwidth]{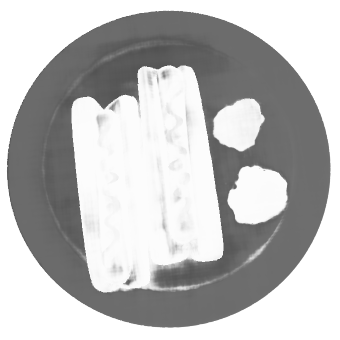} \\
        \raisebox{32pt}{\rotatebox[origin=c]{90}{Truck}}&
        \includegraphics[width=0.14\textwidth]{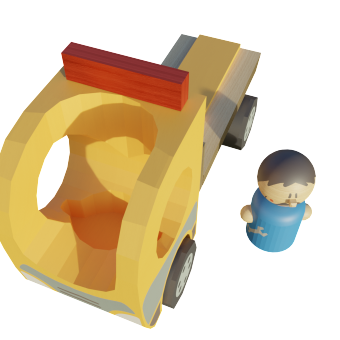} &
        \includegraphics[width=0.14\textwidth]{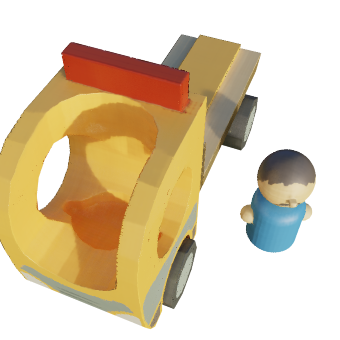} &
        \includegraphics[width=0.14\textwidth]{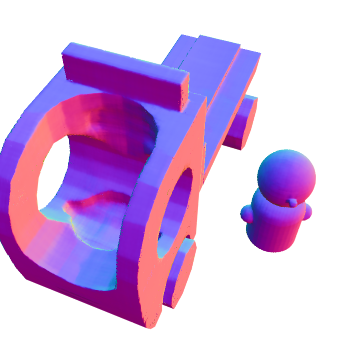} &
        \includegraphics[width=0.14\textwidth]{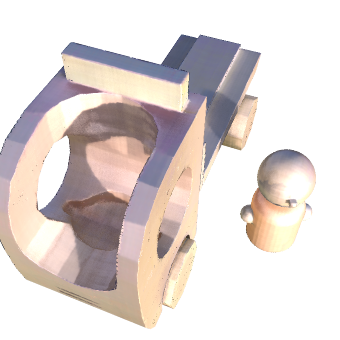} &
        \includegraphics[width=0.14\textwidth]{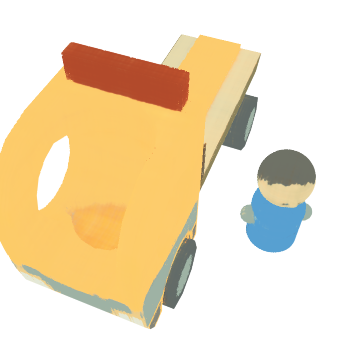} &
        \includegraphics[width=0.14\textwidth]{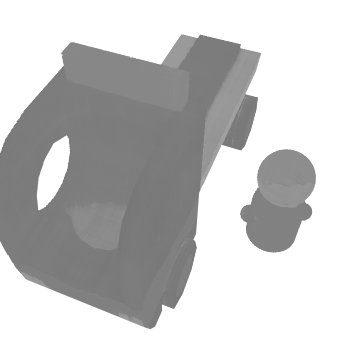} \\
        \raisebox{32pt}{\rotatebox[origin=c]{90}{Chess}}&
        \includegraphics[width=0.14\textwidth]{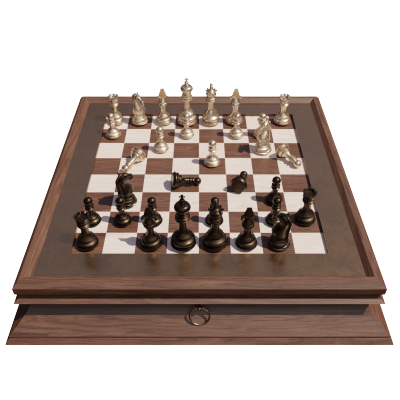} &
        \includegraphics[width=0.14\textwidth]{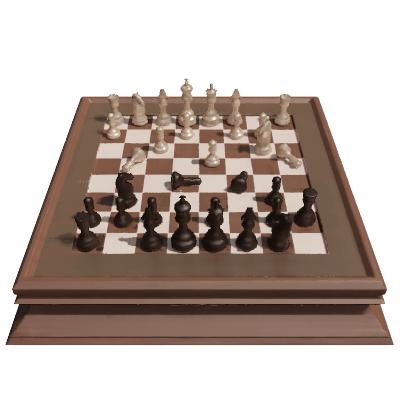} &
        \includegraphics[width=0.14\textwidth]{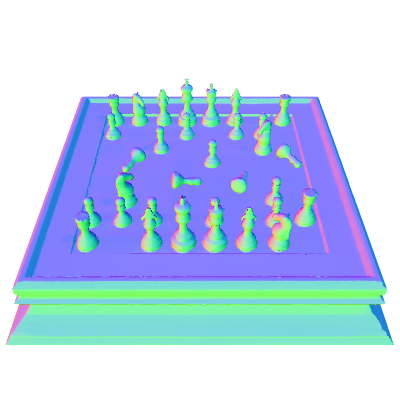} &
        \includegraphics[width=0.14\textwidth]{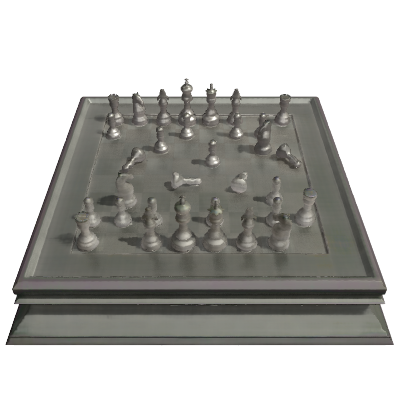} &
        \includegraphics[width=0.14\textwidth]{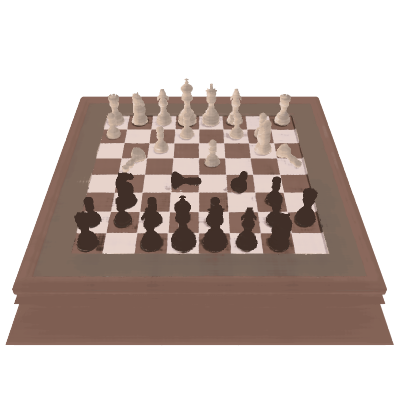} &
        \includegraphics[width=0.14\textwidth]{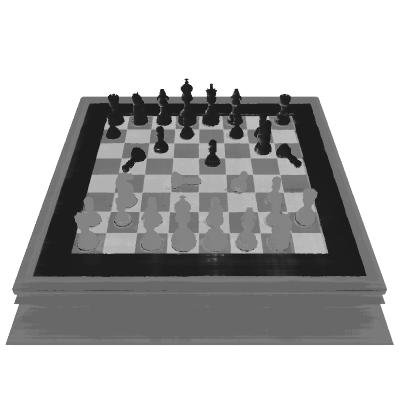} \\
        \raisebox{32pt}{\rotatebox[origin=c]{90}{Car}}&
        \includegraphics[width=0.14\textwidth]{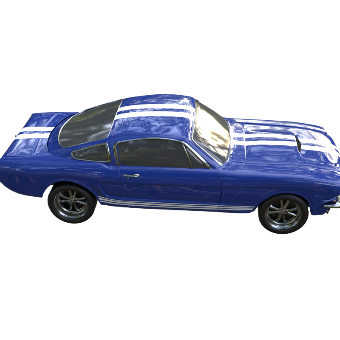} &
        \includegraphics[width=0.14\textwidth]{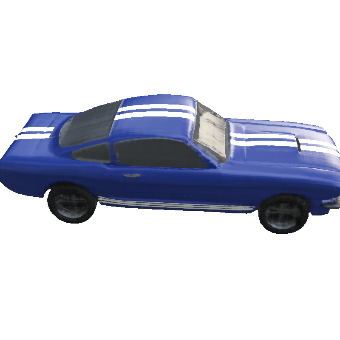} &
        \includegraphics[width=0.14\textwidth]{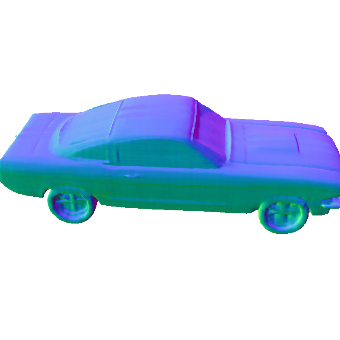} &
        \includegraphics[width=0.14\textwidth]{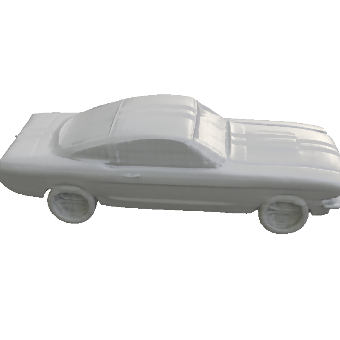} &
        \includegraphics[width=0.14\textwidth]{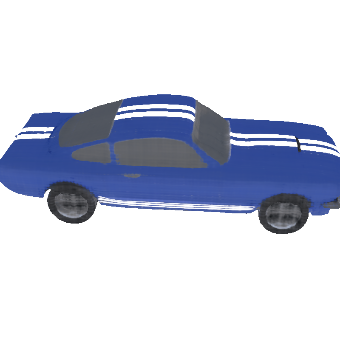} &
        \includegraphics[width=0.14\textwidth]{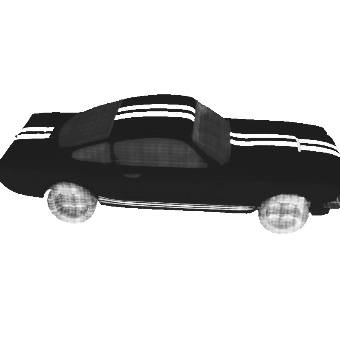} \\
        \raisebox{32pt}{\rotatebox[origin=c]{90}{Helmet}}&
        \includegraphics[width=0.14\textwidth]{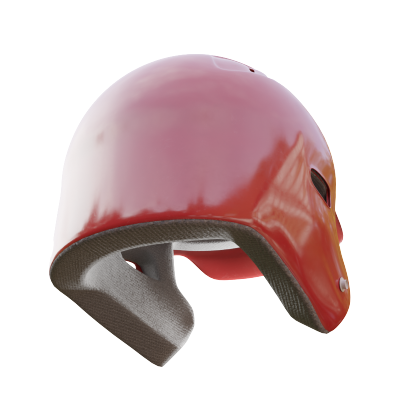} &
        \includegraphics[width=0.14\textwidth]{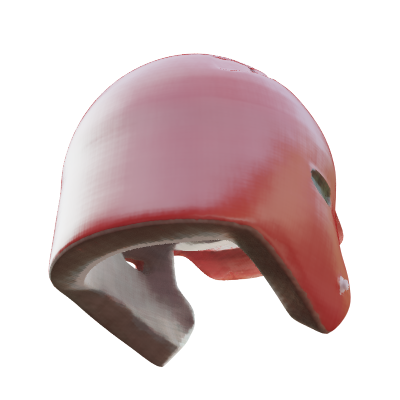} &
        \includegraphics[width=0.14\textwidth]{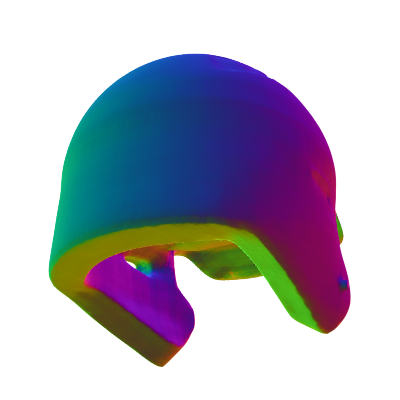} &
        \includegraphics[width=0.14\textwidth]{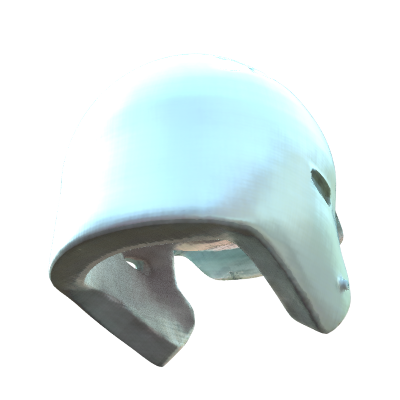} &
        \includegraphics[width=0.14\textwidth]{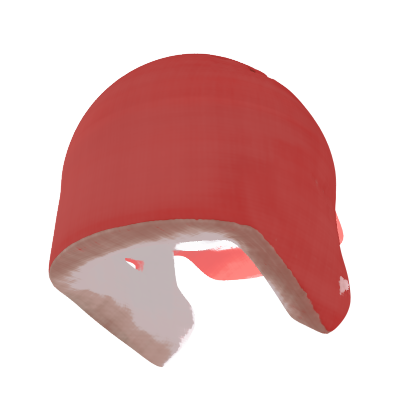} &
        \includegraphics[width=0.14\textwidth]{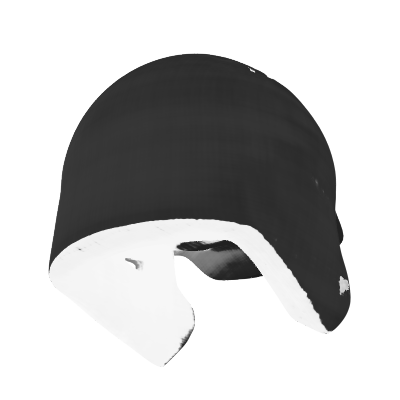} \\
        & (a) gt & (b) rendering & (c) normal & (d) light & (e) albedo & (f) roughness \\
        \end{tabular}
    }
\vspace{-5pt}
	\caption{\textbf{Results on synthetic datasets.} We present additional results obtained from our method on several synthetic scenes, where we visualize the normal (c), light (d), albedo (e), and roughness (f) components, respectively. The fidelity of our method is further demonstrated by comparing the rendered results (b) with the ground truth images (a).}\label{fig:decomp}
\end{figure*}

\section{Implementation Details} \label{sec:detail}
\paragraph{Rendering equation details.} As described in Sec. 3.4, our approach divides the simplified Disney BRDF \cite{burley2012physically, li2018learning, bi2020deep} into specular and diffuse components. Following the methodology from \cite{zhang2021physg}, we employ the inner product of SGs to approximate the computation of the rendering equation. The position \textbf{x} is dropped in the following equation due to the distant illumination assumption. Specifically, the term $\boldsymbol{\omega}_{i} \cdot \textbf{n}$ is approximated by a SG as follows:

\begin{equation}
\boldsymbol{\omega}_{i} \cdot \mathbf{n} \approx G\left(\boldsymbol{\omega}_{i} ; 0.0315, \mathbf{n}, 32.7080\right)-31.7003.
\end{equation}

For the specular component $f_s$, we utilize the following simplified Disney BRDF model:
\begin{equation}
\begin{aligned}
f_{s}\left(\boldsymbol{\omega}_{o}, \boldsymbol{\omega}_{i}\right)&=\mathcal{M}\left(\boldsymbol{\omega}_{o}, \boldsymbol{\omega}_{i}\right) \mathcal{D}(\mathbf{h}), \\ \mathbf{h}&=\frac{\boldsymbol{\omega}_{o}+\boldsymbol{\omega}_{i}}{\Vert \boldsymbol{\omega}_{o}+\boldsymbol{\omega}_{i} \Vert_{2}}, 
\end{aligned}
\end{equation}
where $\mathcal{M}$ represents the Fresnel with shadowing effects, and $\mathcal{D}$ is the normalized distribution function. The $\mathcal{D}$ function is also represented using a single SG:
\begin{equation}
\mathcal{D}(\mathbf{h})=G(\mathbf{h} ; \boldsymbol{\xi}, \lambda, \boldsymbol{\mu}).
\end{equation}

To simplify the computation, we assume an isotropic specular BRDF, aligning $\boldsymbol{\xi}$ with the surface normal $\mathbf{n}$, and adopt the monochrome assumption that makes $\boldsymbol{\mu}$ an identical vector. Based on these assumptions, we then adapt $\mathcal{D}$ and $\mathcal{M}$ as follows:

\begin{equation}
\begin{aligned}
\mathcal{D}_{\mathbf{x}}(\mathbf{h}) &= G\left(\mathbf{h} ; \mathbf{n}, \frac{\lambda}{4 \mathbf{h} \cdot \boldsymbol{\omega}_{o}}, \boldsymbol{\mu}\right), \\
\mathcal{M}_{\mathbf{x}}\left(\boldsymbol{\omega}_{o}, \boldsymbol{\omega}_{i}\right) &\approx \mathcal{M}\left(\boldsymbol{\omega}_{o}, 2\left(\boldsymbol{\omega}_{o} \cdot \mathbf{n}\right) \mathbf{n}-\boldsymbol{\omega}_{o}\right).
\end{aligned}
\end{equation}
where $\mathcal{D}$ is spherically warped, and $\mathcal{M}$ is approximated by a constant.

\par Finally, we can compute the $c_{\text{diffuse}}$ and $c_{\text{specular}}$ in Equ. (10) in the main text through the fast inner product of SGs \cite{Meder2018HemisphericalGF}.

\paragraph{Our simplified specular BRDF model.} 
\par 
Our Physically-Based Rendering (PBR) network outputs albedo $a \in \mathcal{R}{+}$, roughness $r \in \mathcal{R}{+}$, and a learnable parameter for specular reflectance $s \in [0,1]^{3}$. Utilizing $s$ and $r$, we compute the specular BRDF $f_s$ following the simplified Disney BRDF model, as in previous works \cite{zhang2021physg, zhang2022modeling}:
\begin{equation*}
\begin{aligned}
f_{s}\left(\boldsymbol{\omega}_{o}, \boldsymbol{\omega}_{i}\right)&=\mathcal{M}\left(\boldsymbol{\omega}_{o}, \boldsymbol{\omega}_{i}\right) \mathcal{D}(\mathbf{h}), \\
\mathbf{h}&=\frac{\boldsymbol{\omega}_{o}+\boldsymbol{\omega}_{i}}{\left\|\boldsymbol{\omega}_{o}+\boldsymbol{\omega}_{i}\right\|_{2}}, \\
\mathcal{M}\left(\boldsymbol{\omega}_{o}, \boldsymbol{\omega}_{i}\right)&=\frac{\mathcal{F}\left(\boldsymbol{\omega}_{o}, \boldsymbol{\omega}_{i}\right) \mathcal{G}\left(\boldsymbol{\omega}_{o}, \boldsymbol{\omega}_{i}\right)}{4\left(\mathbf{n} \cdot \boldsymbol{\omega}_{o}\right)\left(\mathbf{n} \cdot \boldsymbol{\omega}_{i}\right)} \\
\mathcal{F}\left(\boldsymbol{\omega}_{o}, \boldsymbol{\omega}_{i}\right)&=\boldsymbol{s}+(1-\boldsymbol{s}) \cdot 2^{-\left(5.55473 \boldsymbol{\omega}_{o} \cdot \mathbf{h}+6.8316\right)\left(\boldsymbol{\omega}_{o} \cdot \mathbf{h}\right)}, \\
\mathcal{G}\left(\boldsymbol{\omega}_{o}, \boldsymbol{\omega}_{i}\right)&=\frac{\boldsymbol{\omega}_{o} \cdot \mathbf{n}}{\boldsymbol{\omega}_{o} \cdot \mathbf{n}(1-k)+k} \cdot \frac{\boldsymbol{\omega}_{i} \cdot \mathbf{n}}{\boldsymbol{\omega}_{i} \cdot \mathbf{n}(1-k)+k}, \\
k&=\frac{(r+1)^{2}}{8}, \\
\mathcal{D}(\mathbf{h})&=G\left(\mathbf{h} ; \mathbf{n}, \frac{2}{r^{4}}, \frac{1}{\pi r^{4}}\right).
\end{aligned}
\end{equation*}

\begin{figure*}
    \centering
    \addtolength{\tabcolsep}{-6.5pt}
    \footnotesize{
        \setlength{\tabcolsep}{1pt} 
        \begin{tabular}{cccccccc}
        \raisebox{23pt}{\rotatebox[origin=c]{90}{Ficus}}&
            \includegraphics[width=0.11\textwidth]{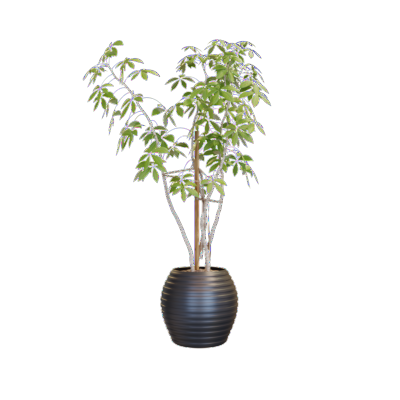} &
            \includegraphics[width=0.11\textwidth]{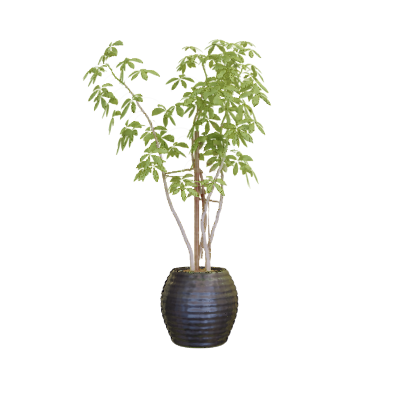} &
            \includegraphics[width=0.11\textwidth]{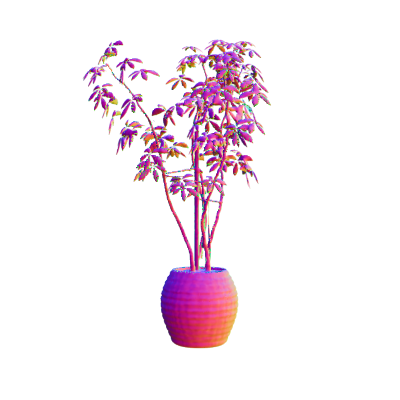} &
            \includegraphics[width=0.11\textwidth]{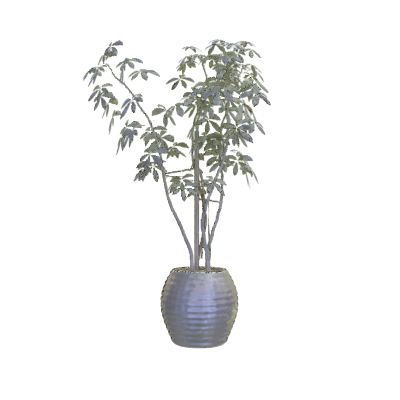} &
            \includegraphics[width=0.11\textwidth]{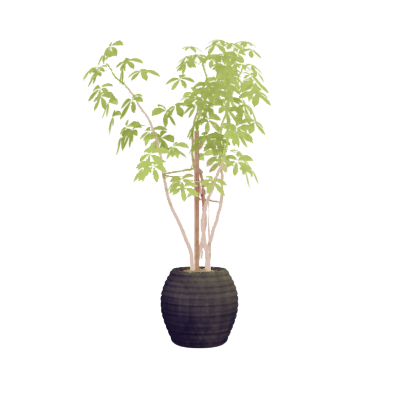} &
            \includegraphics[width=0.11\textwidth]{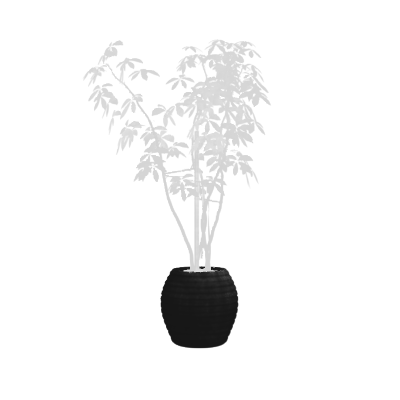} &
            \includegraphics[width=0.22\textwidth]{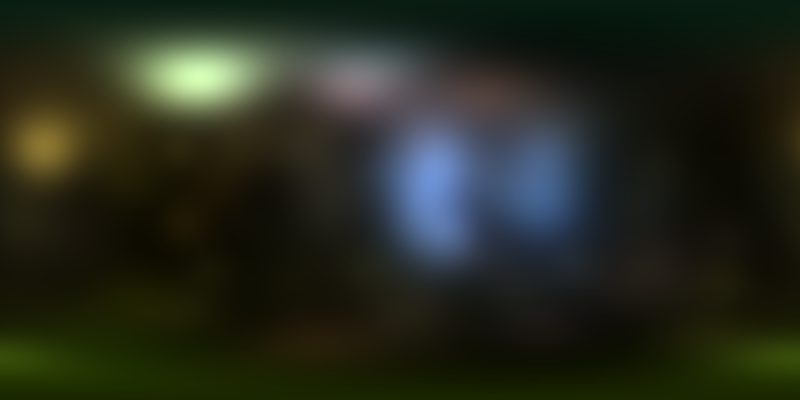} \\
            \raisebox{23pt}{\rotatebox[origin=c]{90}{Stool}}&
            \includegraphics[width=0.11\textwidth]{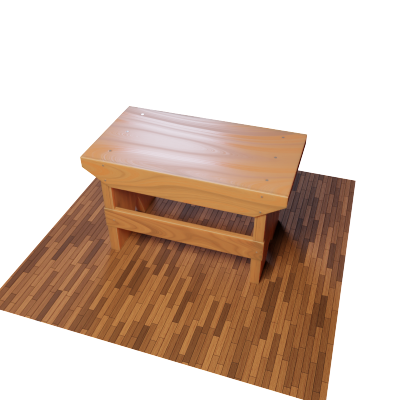} &
            \includegraphics[width=0.11\textwidth]{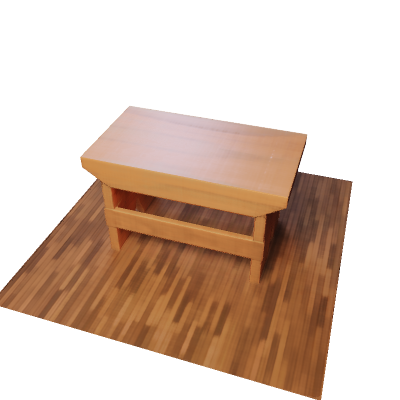} &
            \includegraphics[width=0.11\textwidth]{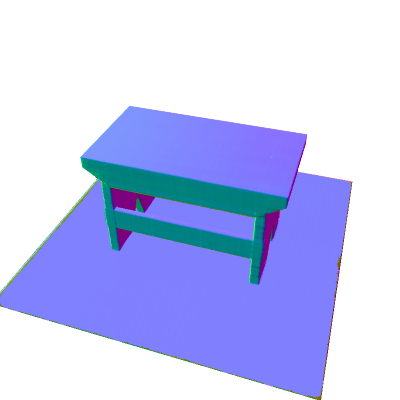} &
            \includegraphics[width=0.11\textwidth]{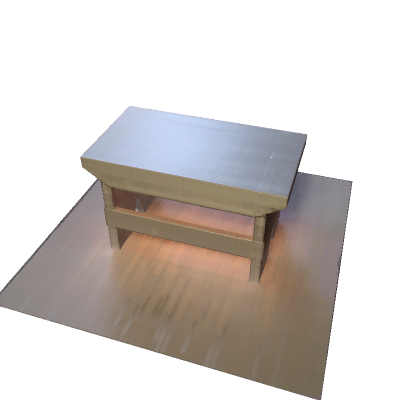} &
            \includegraphics[width=0.11\textwidth]{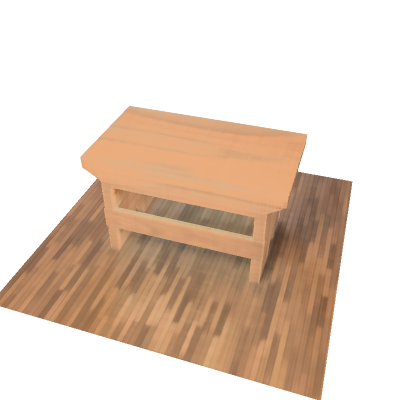} &
            \includegraphics[width=0.11\textwidth]{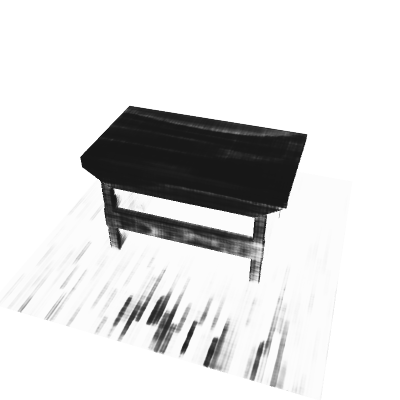} &
            \includegraphics[width=0.22\textwidth]{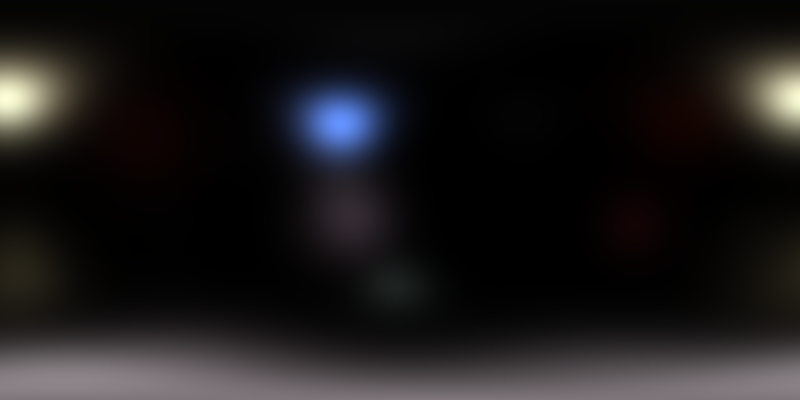} \\
            \raisebox{23pt}{\rotatebox[origin=c]{90}{Mic}}&
            \includegraphics[width=0.11\textwidth]{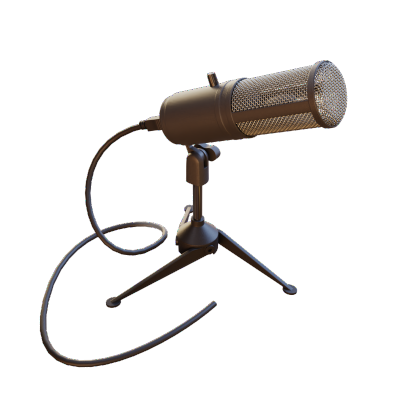} &
            \includegraphics[width=0.11\textwidth]{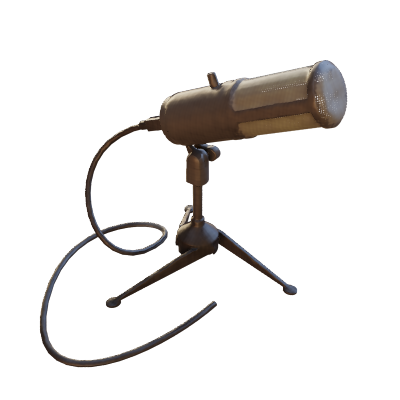} &
            \includegraphics[width=0.11\textwidth]{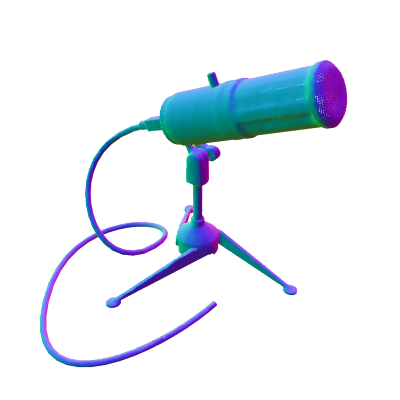} &
            \includegraphics[width=0.11\textwidth]{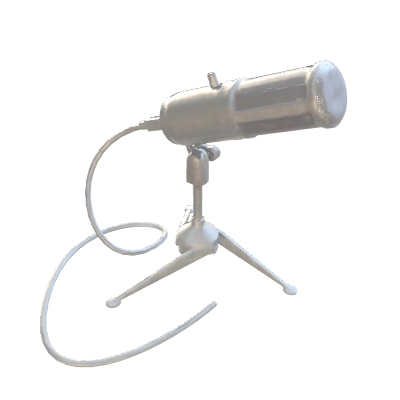} &
            \includegraphics[width=0.11\textwidth]{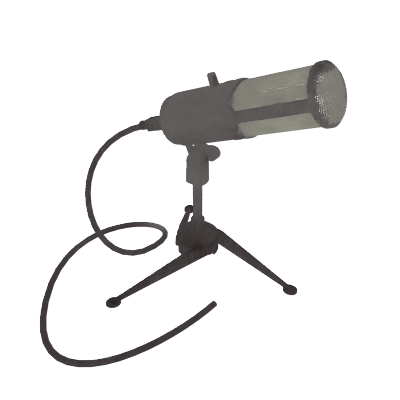} &
            \includegraphics[width=0.11\textwidth]{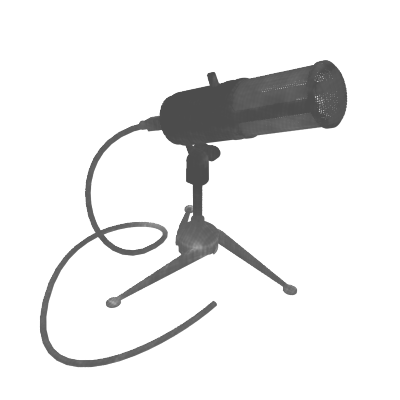} &
            \includegraphics[width=0.22\textwidth]{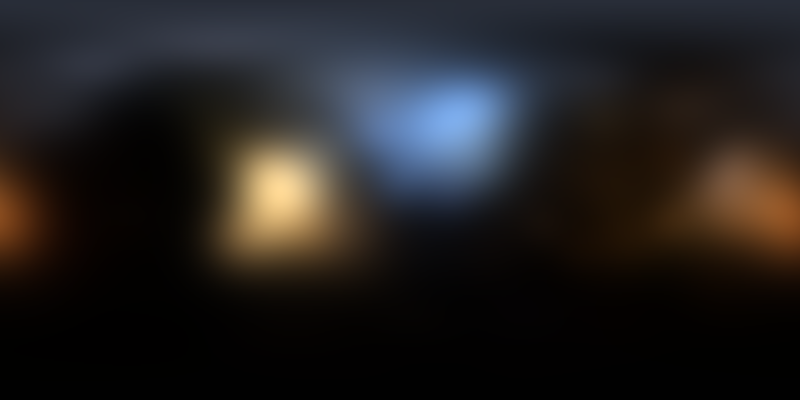} \\
            \raisebox{23pt}{\rotatebox[origin=c]{90}{Headset}}&
            \includegraphics[width=0.11\textwidth]{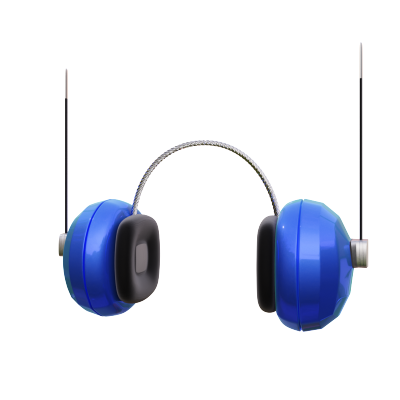} &
            \includegraphics[width=0.11\textwidth]{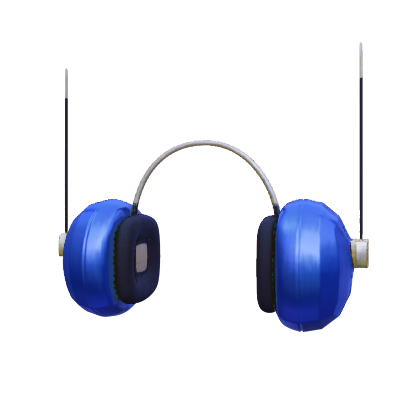} &
            \includegraphics[width=0.11\textwidth]{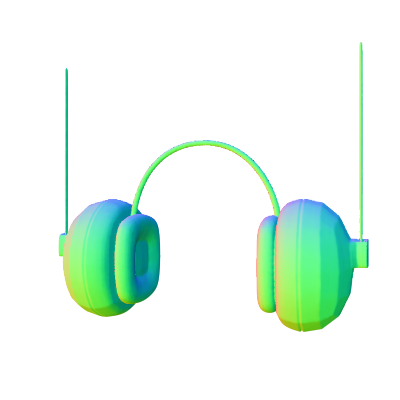} &
            \includegraphics[width=0.11\textwidth]{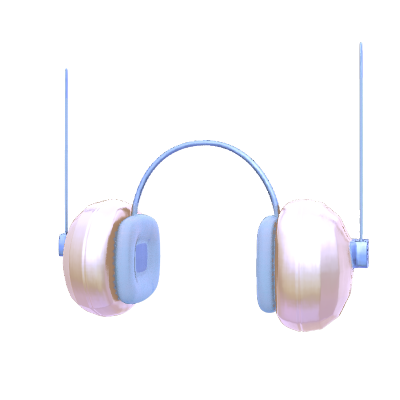} &
            \includegraphics[width=0.11\textwidth]{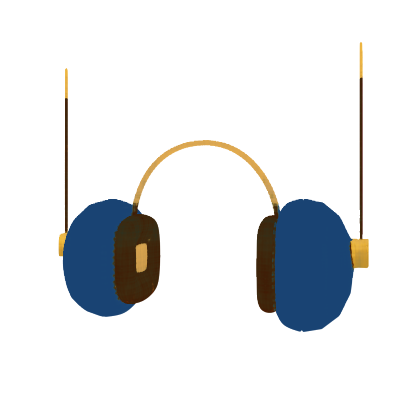} &
            \includegraphics[width=0.11\textwidth]{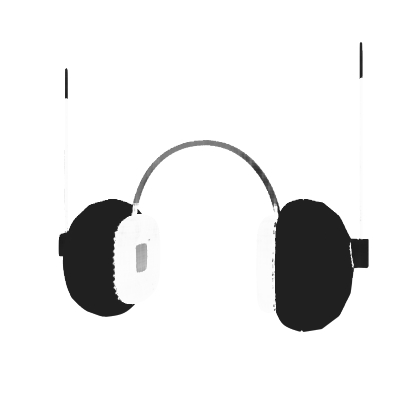} &
            \includegraphics[width=0.22\textwidth]{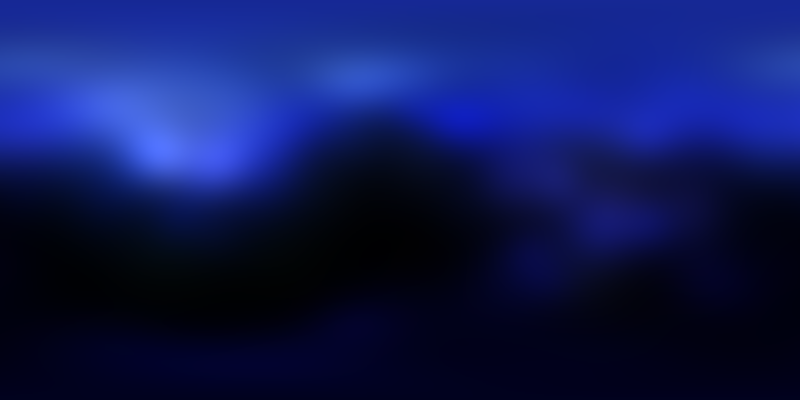} \\
            \raisebox{23pt}{\rotatebox[origin=c]{90}{Jade}}&
            \includegraphics[width=0.11\textwidth]{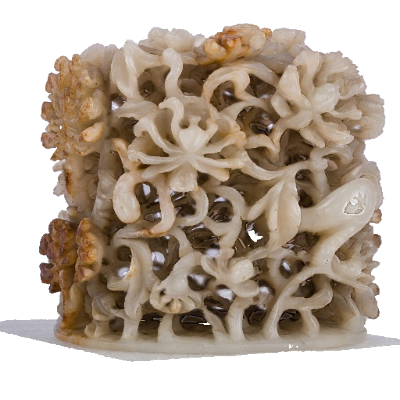} &
            \includegraphics[width=0.11\textwidth]{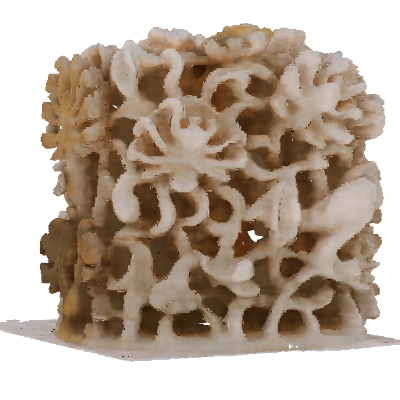} &
            \includegraphics[width=0.11\textwidth]{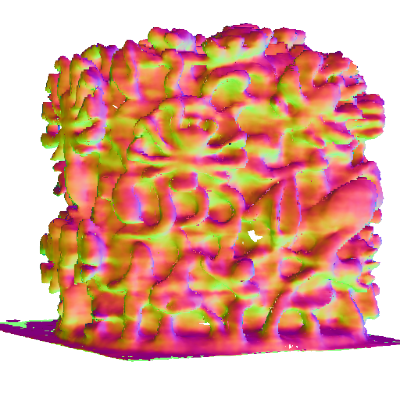} &
            \includegraphics[width=0.11\textwidth]{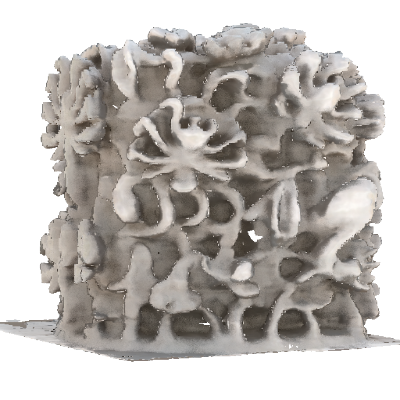} &
            \includegraphics[width=0.11\textwidth]{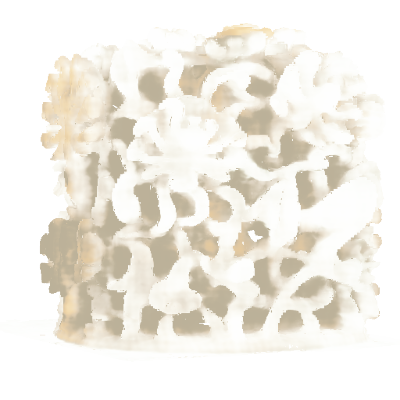} &
            \includegraphics[width=0.11\textwidth]{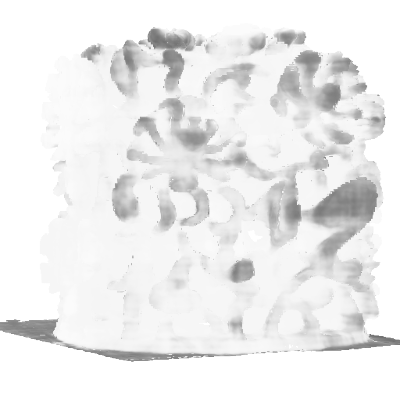} &
            \includegraphics[width=0.22\textwidth]{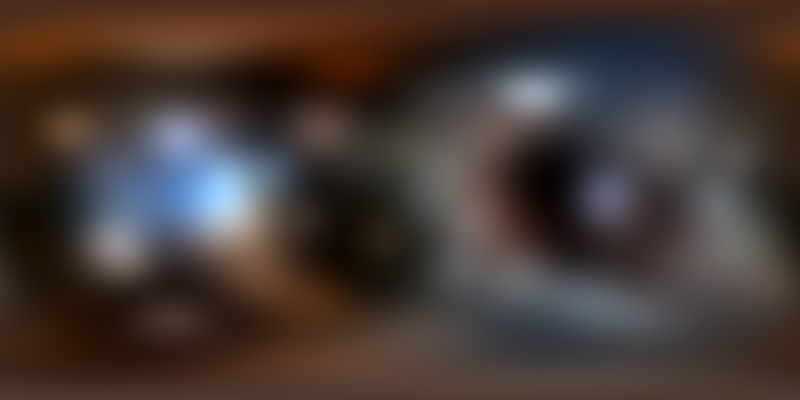} \\
            \raisebox{23pt}{\rotatebox[origin=c]{90}{Clock}}&
            \includegraphics[width=0.11\textwidth]{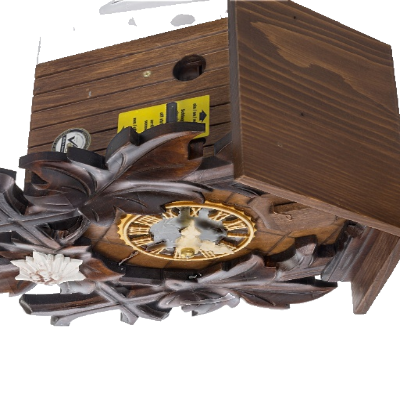} &
            \includegraphics[width=0.11\textwidth]{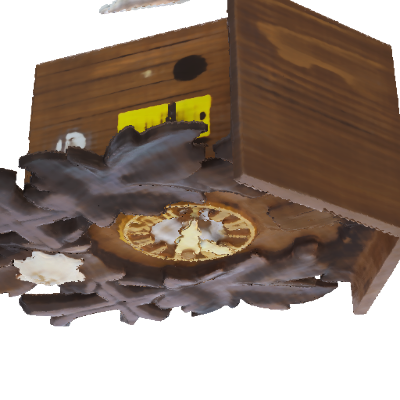} &
            \includegraphics[width=0.11\textwidth]{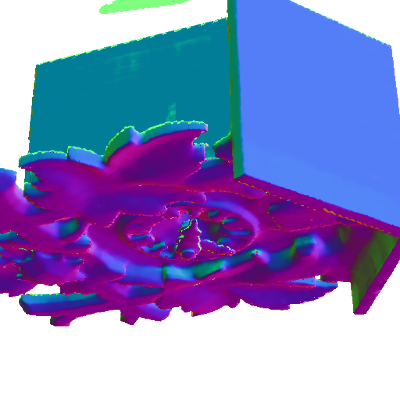} &
            \includegraphics[width=0.11\textwidth]{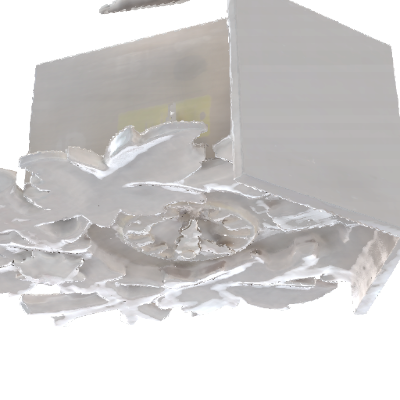} &
            \includegraphics[width=0.11\textwidth]{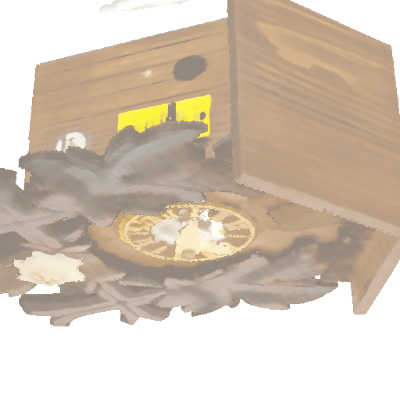} &
            \includegraphics[width=0.11\textwidth]{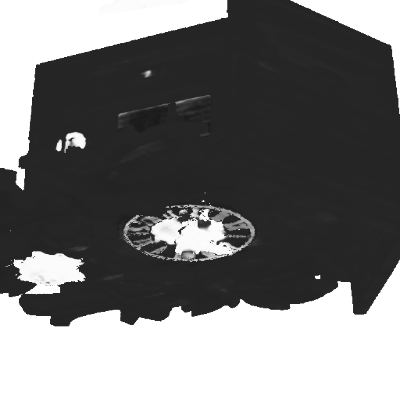} &
            \includegraphics[width=0.22\textwidth]{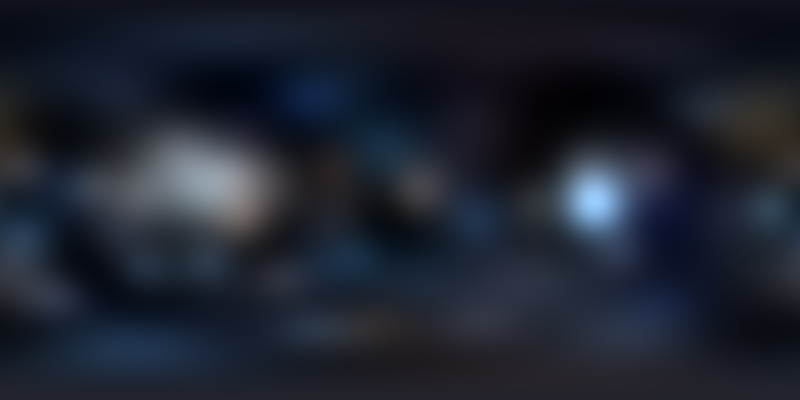} \\ 
            \raisebox{23pt}{\rotatebox[origin=c]{90}{Toy}}&
            \includegraphics[width=0.11\textwidth]{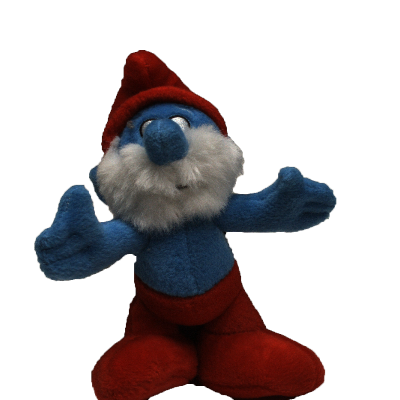} &
            \includegraphics[width=0.11\textwidth]{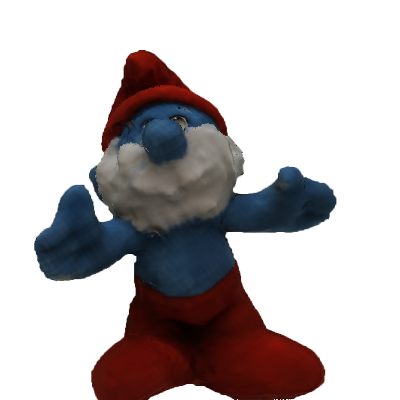} &
            \includegraphics[width=0.11\textwidth]{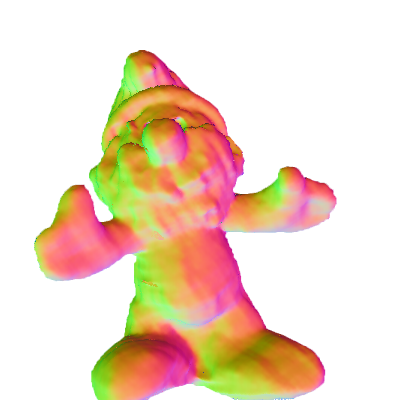} &
            \includegraphics[width=0.11\textwidth]{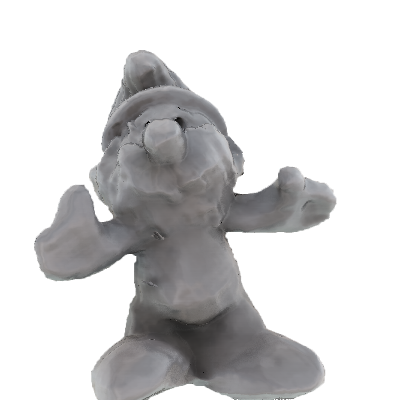} &
            \includegraphics[width=0.11\textwidth]{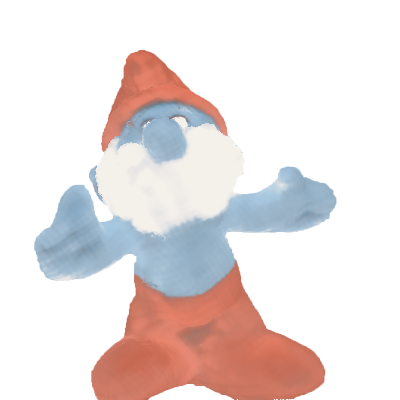} &
            \includegraphics[width=0.11\textwidth]{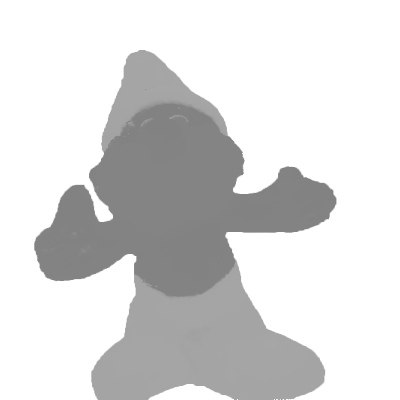} &
            \includegraphics[width=0.22\textwidth]{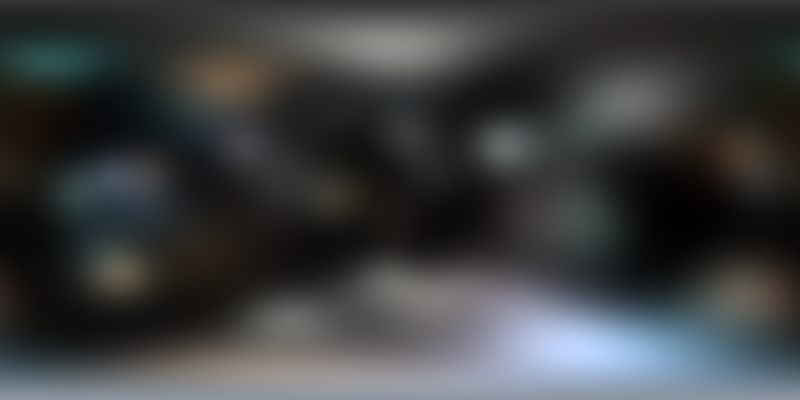} \\ 
            \raisebox{23pt}{\rotatebox[origin=c]{90}{Bear}}&
            \includegraphics[width=0.11\textwidth]{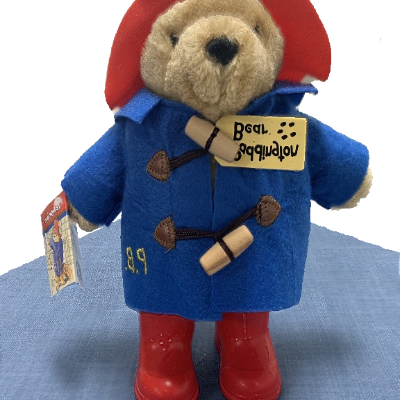} &
            \includegraphics[width=0.11\textwidth]{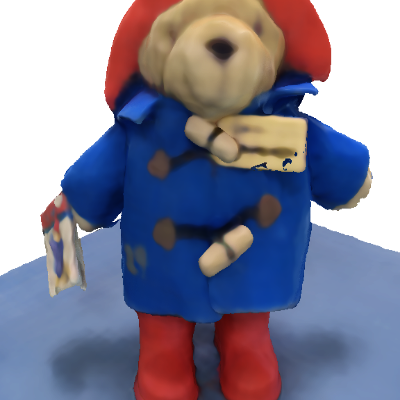} &
            \includegraphics[width=0.11\textwidth]{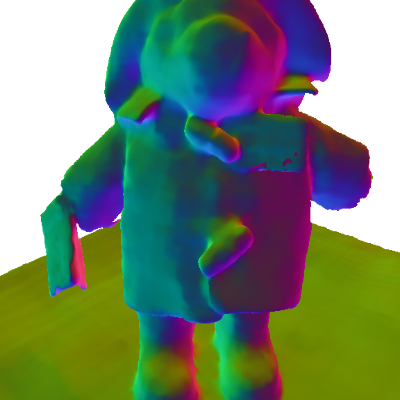} &
            \includegraphics[width=0.11\textwidth]{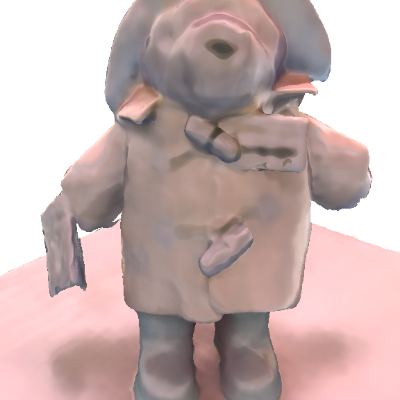} &
            \includegraphics[width=0.11\textwidth]{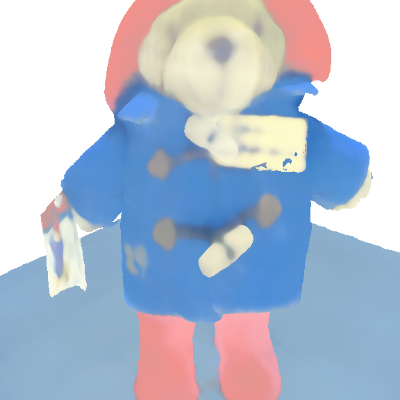} &
            \includegraphics[width=0.11\textwidth]{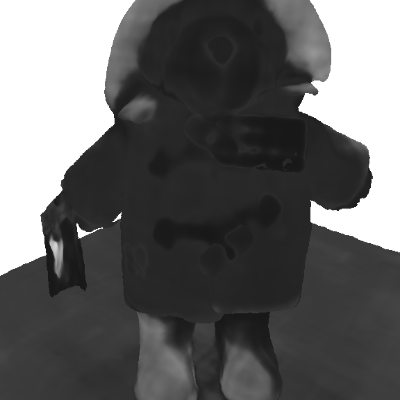} &
            \includegraphics[width=0.22\textwidth]{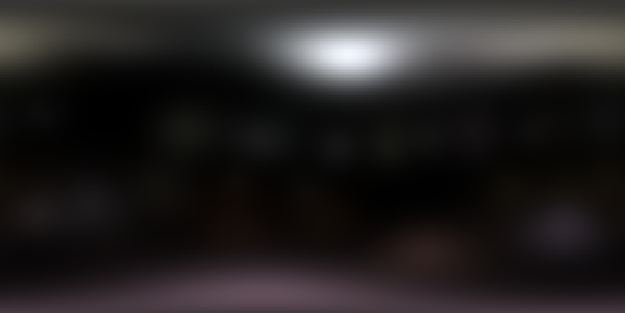} \\
            \raisebox{23pt}{\rotatebox[origin=c]{90}{Sculp}}&
            \includegraphics[width=0.11\linewidth]{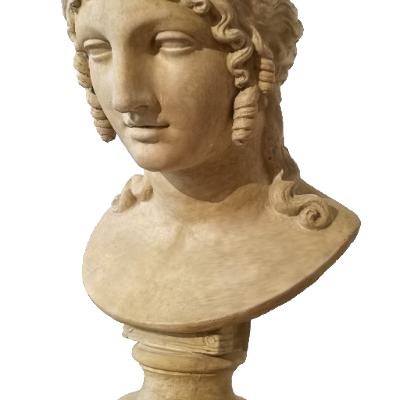} &
            \includegraphics[width=0.11\linewidth]{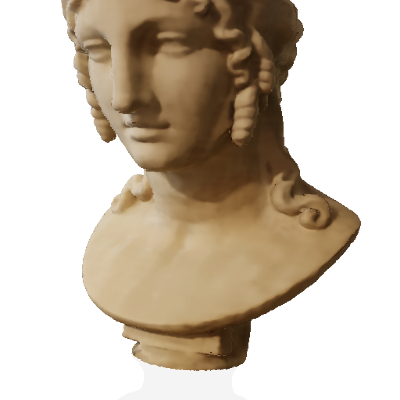} &
            \includegraphics[width=0.11\linewidth]{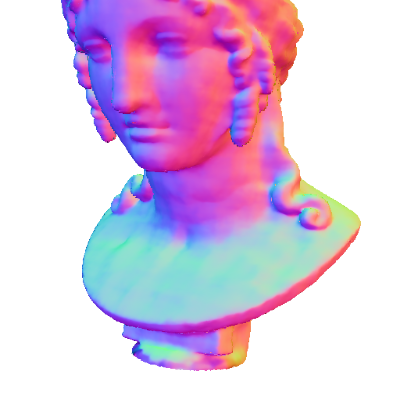} &
            \includegraphics[width=0.11\linewidth]{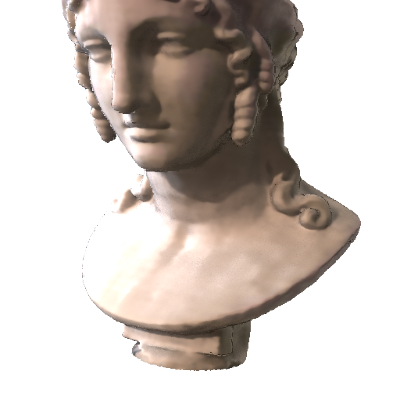} &
            \includegraphics[width=0.11\linewidth]{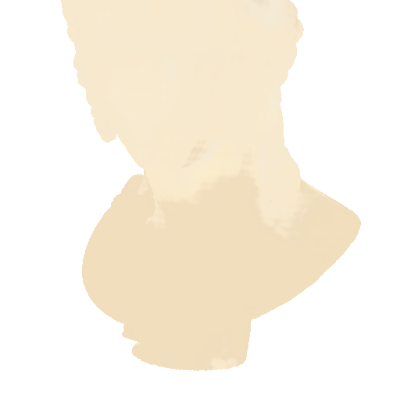} &
            \includegraphics[width=0.11\linewidth]{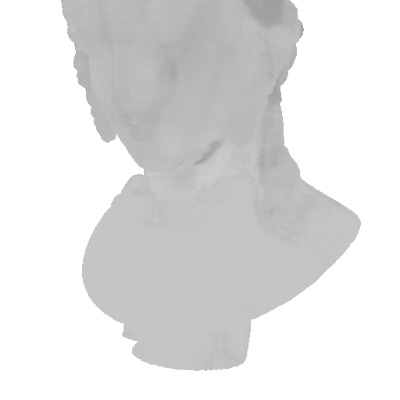} &
            \includegraphics[width=0.22\linewidth]{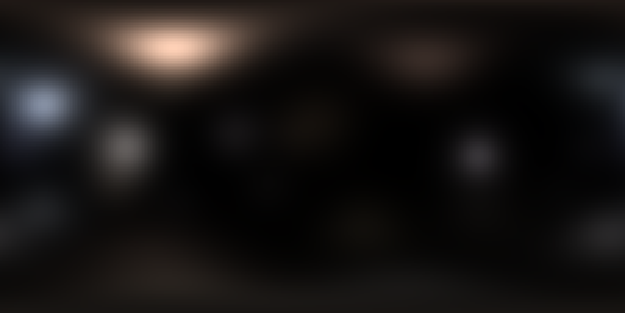} \\ 
            & (a) gt & (b) rendering & (c) normal & (d) light & (e) albedo & (f) roughness & (g) env-map \\
        \end{tabular}
    }
\vspace{-5pt}
	\caption{\textbf{Other datasets' results.}
 We present experimental results on various datasets, including four synthetic scenes and five real scenes. The first four rows in the results show the performance of our method in synthetic scenes, while the last five rows demonstrate its effectiveness on real scenes. In each dataset, we present the input ground-truth image (a), our rendering result (b), normal (c), light (d), albedo (e), and roughness (f) obtained through our method. These experiments illustrate the generalizability of our method across diverse datasets and demonstrate its ability to produce high-quality results.
 }
\label{fig:other-decomp}
\end{figure*}

\paragraph{Details of loss function in PBR decomposition.} As described in Sec. 3.5, our final loss function is the combination of pixel loss, smooth loss, and KL sparse loss:
\begin{equation}
\begin{aligned}
\mathcal{L}&=\lambda_{rgb}\mathcal{L}_{rgb}+\lambda_{sm}\mathcal{L}_{sm}+\lambda_{KL}\mathcal{L}_{KL}, \\
\mathcal{L}_{sm} &= \frac{1}{N} \Vert c(x), c^{\prime}(x) \Vert_2^2, \\
\mathcal{L}_{KL} &= \frac{1}{N} \sum_{k=1}^N (\rho log\frac{\rho}{\hat{\mathbf{z}}_k} + (1 - \rho)log\dfrac{1-\rho}{1- \hat{\textbf{z}}_k}),
\end{aligned}
\end{equation}
where $\mathcal{L}_{rgb}$ refers to the $\mathcal L_2$ loss between the color obtained through BRDF rendering and the ground truth. $\mathcal L_{sm}$ refers to the smooth loss of the output materials albedo and roughness, where $c(x)$ is the direct output of the PBR network, and $c^{\prime}(x)$ is the result obtained by adding 0.01x Gaussian noises to the latent code \textbf{z} of the PBR network. $\mathcal L_{KL}$ is the sparsity constraint on the latent code \textbf{z} of the PBR network, with $\hat{\mathbf{z}}$ being the average value of each channel of \textbf{z}, and $\rho$ set to 0.05.

\begin{figure*}
    \centering
    \addtolength{\tabcolsep}{-6.5pt}
    \footnotesize{
        \setlength{\tabcolsep}{2pt} 
        \begin{tabular}{cccc}
        \multicolumn{2}{c}{Random Sampling} & \multicolumn{2}{c}{Directional Sampling} \\
        \\
        \includegraphics[width=0.1\textwidth]{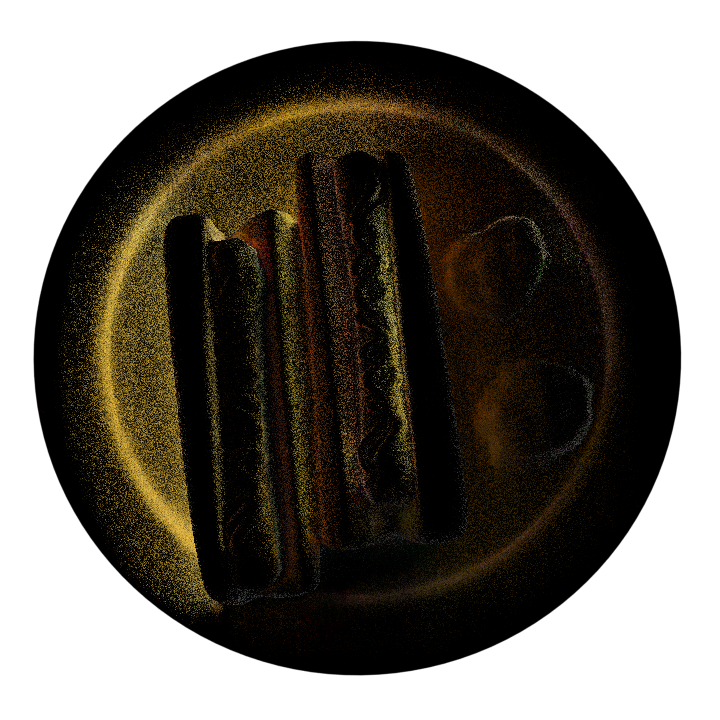} &
        \includegraphics[width=0.1\textwidth]{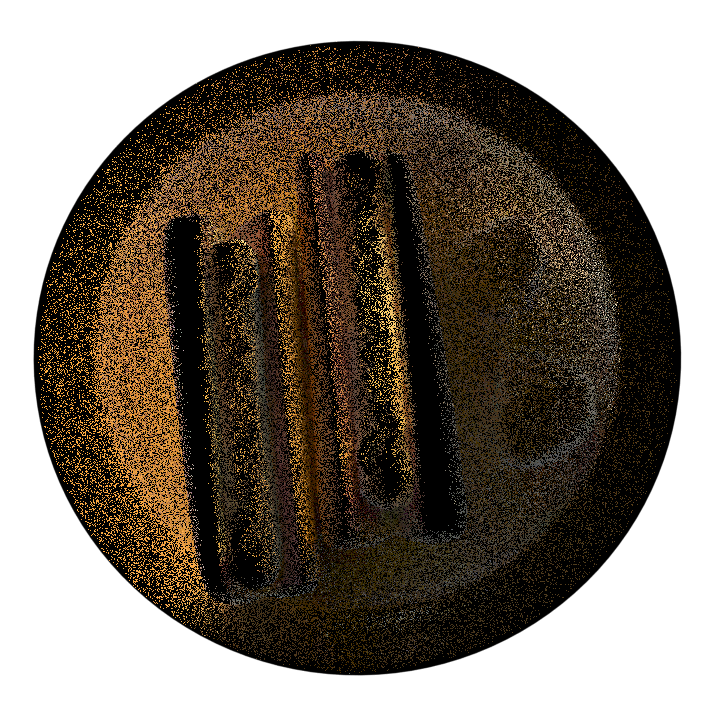} &
        \includegraphics[width=0.1\textwidth]{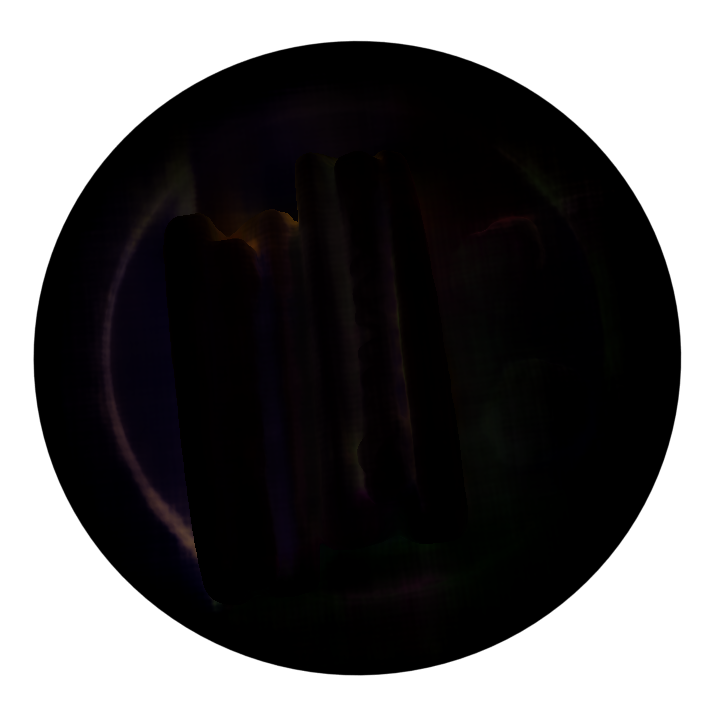} &
        \includegraphics[width=0.1\textwidth]{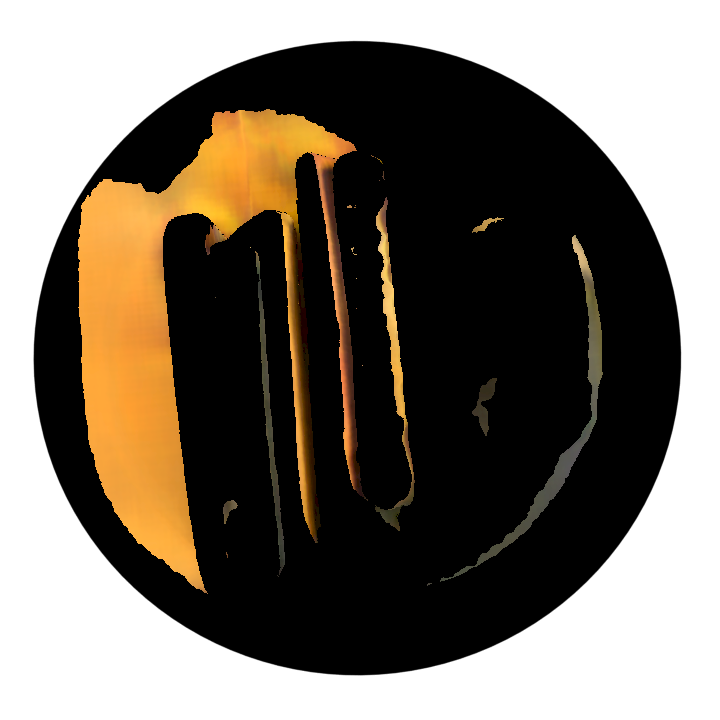} \\
        (a) InvRender & (b) Ours & (c) InvRender & (d) Ours  \\
        \end{tabular}
    }
\vspace{-5pt}
    \caption{\textbf{Visualization of masked indirect illumination.} Our method can model the indirect radiance field more robustly, making the indirect light at edges more accurate, thereby removing artifacts baked into the PBR material.
    } \label{fig:seg_inidrect}
    \label{fig:segment-indir}
\end{figure*}

\begin{figure*}
    \centering
    \addtolength{\tabcolsep}{-6.5pt}
    \footnotesize{
        \setlength{\tabcolsep}{2pt} 
        \begin{tabular}{cccccc}
        $\gamma = 0.01$ & $\gamma = 0.1$ & $\gamma = 0.4$ & $\gamma = 0.7$ & $\gamma = 1.0$ & LDR \\
        \includegraphics[width=0.12\textwidth]{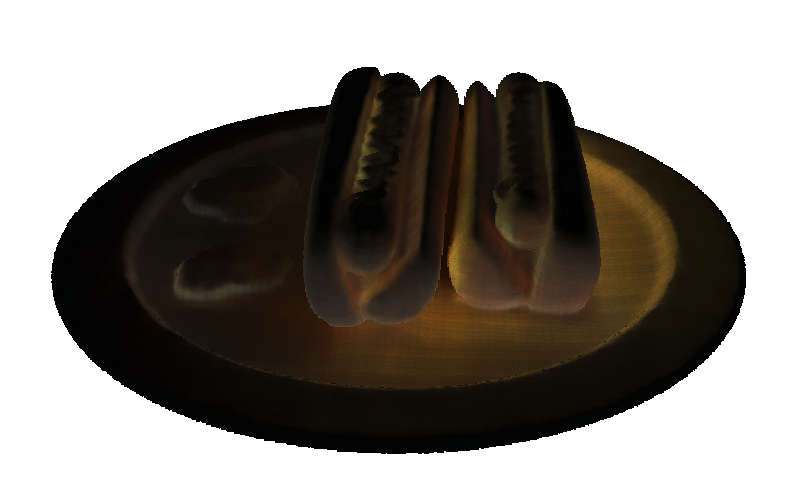} &
        \includegraphics[width=0.12\textwidth]{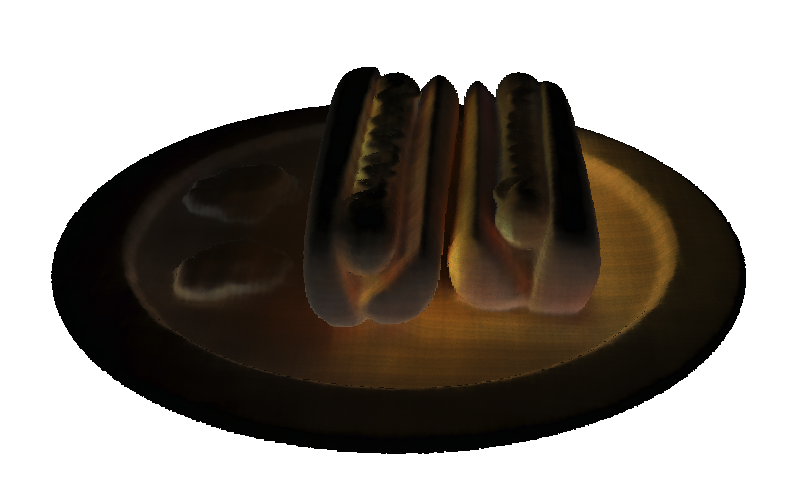} &
        \includegraphics[width=0.12\textwidth]{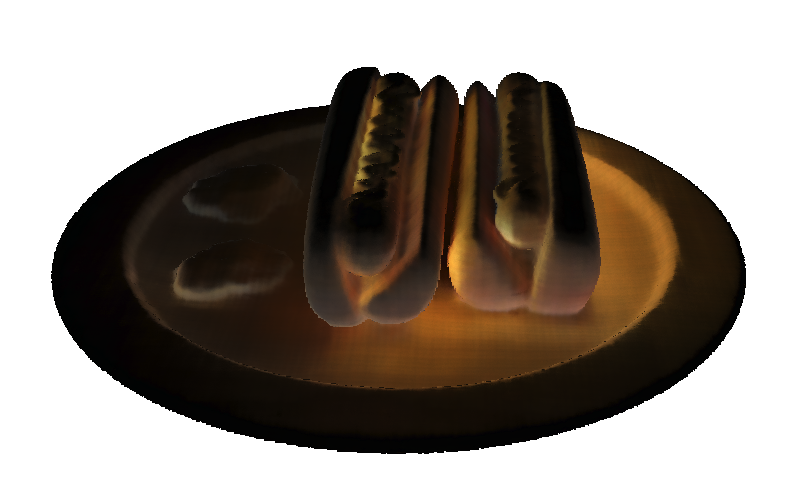} &
        \includegraphics[width=0.12\textwidth]{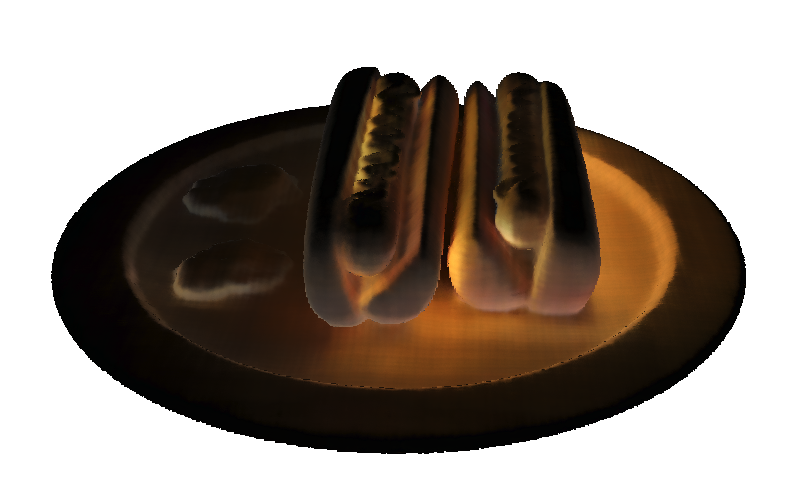} &
        \includegraphics[width=0.12\textwidth]{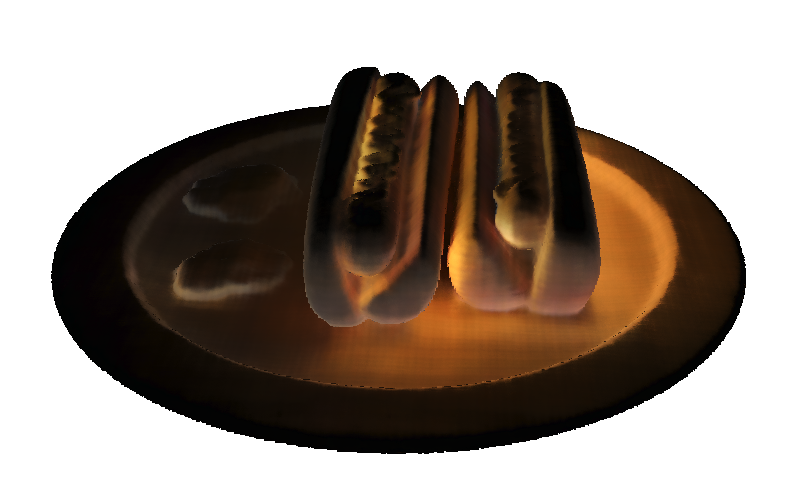} &
        \includegraphics[width=0.12\textwidth]{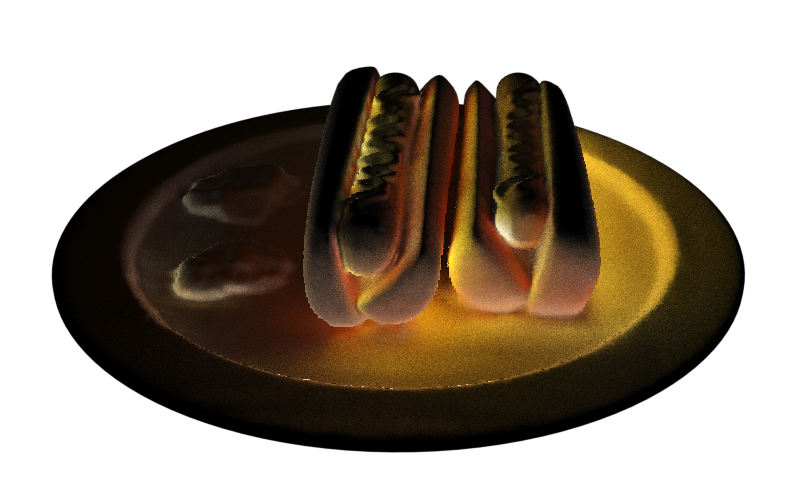} \\
        \end{tabular}
    }
\vspace{-5pt}
	\caption{\textbf{Visualization of non-linearly mapped indirect radiance field.} We conducted experiments to visualize the ACES-mapped indirect radiance field using different values of $\gamma$. Our results demonstrate that our method can accurately represent high-quality indirect illumination in scenes with natural light intensity, outperforming the vanilla LDR indirect radiance field. Furthermore, our deformable ACES tone mapping method can be applied to scenes under various lighting conditions, adapting to different $\gamma$. This showcases the versatility and robustness of our method in handling diverse scenarios, enhancing the overall quality of inverse rendering.
 }\label{fig:hdr-indir}
\end{figure*}

\section{Additional Results} \label{sec: add-results}
\paragraph{More qualitative results.} Our method can effectively remove shadows and indirect illumination baked into albedo and roughness, thanks to our accurate modeling of each decomposition component. Therefore, our method can certainly handle scenes with less intense lighting. Fig. \ref{fig:other-decomp} shows the results of our method on real-world datasets and some synthetic datasets, including scenes with shadows and specular, as well as diffuse objects. Our method can robustly perform inverse rendering in any situation without baking shadows and illumination into PBR materials. For example, in the penultimate real-world scene, our method achieved clean albedo without baking shadow. Fig. \ref{fig:decomp} shows the complete results of our method on the synthetic dataset.
\paragraph{More quantitative results.} We present the complete metrics of our method compared to other methods on a given synthetic dataset in Tab. \ref{tab:psnr-albedo}-Tab. \ref{tab:lpips-roughness}. Our method's albedo and roughness also surpass existing SOTA methods in quantitative metrics. However, it should be noted that since inverse rendering is a highly ill-posed problem, the hues of the PBR materials decomposed by different methods also vary. Our quantitative metrics are only for reference. The quality should be judged based on whether the qualitative results successfully remove shadows, lighting, and other artifacts.

\begin{table}[]
\centering
\resizebox{0.45\textwidth}{!}{
\begin{tabular}{lccccccc}
\hline
& hotdog           & lego           & ficus      & stool           & helmet         & chess            & mean           \\
\hline
ours       & \cellcolor{yzybest}22.65 & 19.44 & \cellcolor{yzybest}20.79 & 17.45 & \cellcolor{yzysecond}20.99 & \cellcolor{yzybest}18.62 & \cellcolor{yzybest}19.99  \\ 
ours-log   & 18.86 & \cellcolor{yzysecond}20.35 & 18.20 & \cellcolor{yzybest}18.89 & \cellcolor{yzybest}21.25 & 17.38 & \cellcolor{yzythird}19.15 \\ 
no aces & \cellcolor{yzythird}18.90 & 19.38 & 16.95 & \cellcolor{yzythird}18.14 & 14.28 & \cellcolor{yzysecond}18.22 & 17.65 \\
no reg-estim & \cellcolor{yzysecond}22.00 & \cellcolor{yzybest}20.96 & \cellcolor{yzythird}19.15 & \cellcolor{yzysecond}18.77 & \cellcolor{yzythird}19.60 & \cellcolor{yzythird}17.69 & \cellcolor{yzysecond}19.70 \\
\hline
invrender & 15.66 & \cellcolor{yzythird}19.80 & 17.41 & 17.62 & 19.00 & 17.67 & 17.86 \\
nvdiffrec & 12.10 & 11.83 & \cellcolor{yzysecond}20.56 & 10.19 & 10.27 & 8.52 & 12.25 \\
\hline
\end{tabular}}
\caption{\textbf{Per-scene PSNR results of the albedo.}}
\label{tab:psnr-albedo}
\end{table}

\begin{table}[]
\centering
\resizebox{0.45\textwidth}{!}{
\begin{tabular}{lccccccc}
\hline
& hotdog           & lego           & ficus      & stool           & helmet         & chess            & mean           \\
\hline
ours       & \cellcolor{yzybest}0.928 & \cellcolor{yzybest}0.894 & \cellcolor{yzybest}0.957 & 0.881 & \cellcolor{yzybest}0.946 & \cellcolor{yzybest}0.854 & \cellcolor{yzybest}0.910 \\ 
ours-log   & \cellcolor{yzythird}0.905 & 0.886 & \cellcolor{yzythird}0.928 & \cellcolor{yzybest}0.899 & 0.928 & \cellcolor{yzythird}0.835 & \cellcolor{yzythird}0.897 \\ 
no aces & 0.904 & \cellcolor{yzysecond}0.893 & 0.911 & \cellcolor{yzythird}0.888 & 0.906 & \cellcolor{yzysecond}0.852 & 0.892 \\
no reg-estim & \cellcolor{yzysecond}0.916 & \cellcolor{yzythird}0.887 & \cellcolor{yzythird}0.949 & 0.878 & \cellcolor{yzysecond}0.938 & 0.820 & \cellcolor{yzysecond}0.898 \\
\hline
invrender & 0.876 & 0.883 & 0.921 & \cellcolor{yzysecond}0.897 & 0.911 & 0.813 & 0.884 \\
nvdiffrec & 0.838 & 0.787 & \cellcolor{yzysecond}0.950 & 0.780 & 0.786 & 0.736 & 0.813 \\
\hline
\end{tabular}}
\caption{\textbf{Per-scene SSIM results of the albedo.}}
\label{tab:ssim-albedo}
\end{table}

\begin{table}[]
\centering
\resizebox{0.45\textwidth}{!}{
\begin{tabular}{lccccccc}
\hline
& hotdog           & lego           & ficus      & stool           & helmet         & chess            & mean           \\
\hline
ours       & \cellcolor{yzybest}0.130 & \cellcolor{yzysecond}0.136 & \cellcolor{yzybest}0.078 & \cellcolor{yzysecond}0.163 & \cellcolor{yzysecond}0.116 & \cellcolor{yzybest}0.190 & \cellcolor{yzybest}0.135 \\ 
ours-log   & \cellcolor{yzythird}0.170 & \cellcolor{yzysecond}0.139 & 0.109 & \cellcolor{yzythird}0.183 & \cellcolor{yzybest}0.107 & \cellcolor{yzysecond}0.201 & \cellcolor{yzysecond}0.152 \\ 
no aces & \cellcolor{yzysecond}0.169 & 0.140 & 0.123 & 0.190 & 0.154 & \cellcolor{yzythird}0.205 & 0.164 \\
no reg-estim & 0.172 & \cellcolor{yzybest}0.138 & \cellcolor{yzythird}0.096 & 0.200 & \cellcolor{yzythird}0.124 & 0.238 & \cellcolor{yzythird}0.161 \\
\hline
invrender & 0.228 & 0.155 & 0.106 & \cellcolor{yzybest}0.132 & 0.151 & 0.226 & 0.166 \\
nvdiffrec & 0.365 & 0.279 & \cellcolor{yzysecond}0.087 & 0.434 & 0.345 & 0.406 & 0.319 \\
\hline
\end{tabular}}
\caption{\textbf{Per-scene LPIPS (VGG) results of the albedo.}}
\label{tab:lpips-albedo}
\end{table}

\begin{table}[]
\centering
\resizebox{0.45\textwidth}{!}{
\begin{tabular}{lccccccc}
\hline
& hotdog           & lego           & ficus      & stool           & helmet         & chess            & mean           \\
\hline
ours       & \cellcolor{yzybest}18.85 & \cellcolor{yzysecond}28.10 & \cellcolor{yzybest}20.98 & \cellcolor{yzysecond}14.27 & 16.47 & \cellcolor{yzybest}17.28 & \cellcolor{yzybest}19.33 \\ 
ours-log   & 14.71 & \cellcolor{yzybest}28.13 & 19.66 & 13.57 & 15.84 & 11.93 & 17.31 \\ 
no aces & \cellcolor{yzythird}17.06 & \cellcolor{yzysecond}28.10 & \cellcolor{yzythird}20.45 & 13.57 & \cellcolor{yzythird}16.79 & 10.62 & 17.77 \\
no reg-estim & \cellcolor{yzysecond}18.33 & 28.08 & 19.51 & 13.24 & 16.15 & \cellcolor{yzysecond}16.73 & \cellcolor{yzysecond}18.67 \\
\hline
invrender & 17.14 & 28.12 & 15.53 & \cellcolor{yzythird}14.24 & \cellcolor{yzysecond}17.77 & 13.05 & 17.64 \\
nvdiffrec & 12.08 & 25.63 & \cellcolor{yzysecond}20.94 & \cellcolor{yzybest}16.82 & \cellcolor{yzybest}18.72 & \cellcolor{yzythird}15.73 & \cellcolor{yzythird}18.32 \\
\hline
\end{tabular}}
\caption{\textbf{Per-scene PSNR results of the roughness.}}
\label{tab:psnr-roughness}
\end{table}

\begin{table}[]
\centering
\resizebox{0.45\textwidth}{!}{
\begin{tabular}{lccccccc}
\hline
& hotdog           & lego           & ficus      & stool           & helmet         & chess            & mean           \\
\hline
ours       & \cellcolor{yzybest}0.919 & \cellcolor{yzybest}0.882 & \cellcolor{yzybest}0.970 & \cellcolor{yzybest}0.886 & \cellcolor{yzythird}0.858 & \cellcolor{yzythird}0.810 & \cellcolor{yzybest}0.888 \\ 
ours-log   & 0.825 & \cellcolor{yzysecond}0.881 & \cellcolor{yzythird}0.964 & 0.835 & 0.836 & 0.684 & 0.837 \\ 
no aces & \cellcolor{yzythird}0.887 & 0.880 & 0.963 & 0.850 & \cellcolor{yzysecond}0.882 & 0.637 & 0.850 \\
no reg-estim & \cellcolor{yzysecond}0.898 & 0.880 & \cellcolor{yzysecond}0.968 & 0.844 & 0.844 & \cellcolor{yzybest}0.878 & \cellcolor{yzysecond}0.885 \\
\hline
invrender & 0.884 & \cellcolor{yzysecond}0.881 & 0.914 & \cellcolor{yzythird}0.872 & \cellcolor{yzybest}0.890 & 0.708 & \cellcolor{yzythird}0.858 \\
nvdiffrec & 0.735 & 0.815 & 0.922 & \cellcolor{yzybest}0.886 & 0.780 & \cellcolor{yzysecond}0.819 & 0.826 \\
\hline
\end{tabular}}
\caption{\textbf{Per-scene SSIM results of the roughness.}}
\label{tab:ssim-roughness}
\end{table}

\begin{table}[]
\centering
\resizebox{0.45\textwidth}{!}{
\begin{tabular}{lccccccc}
\hline
& hotdog           & lego           & ficus      & stool           & helmet         & chess            & mean           \\
\hline
ours       & \cellcolor{yzybest}0.143 & \cellcolor{yzythird}0.262 & \cellcolor{yzysecond}0.045 & \cellcolor{yzysecond}0.227 & \cellcolor{yzybest}0.173 & \cellcolor{yzybest}0.198 & \cellcolor{yzybest}0.175 \\ 
ours-log   & 0.179 & \cellcolor{yzysecond}0.257 & 0.052 & 0.243 & 0.293 & \cellcolor{yzysecond}0.204 & 0.205 \\ 
no aces & \cellcolor{yzythird}0.173 & 0.264 & \cellcolor{yzythird}0.048 & 0.239 & \cellcolor{yzysecond}0.176 & 0.271 & \cellcolor{yzythird}0.195 \\
no reg-estim & \cellcolor{yzysecond}0.158 & 0.264 & \cellcolor{yzybest}0.040 & 0.266 & \cellcolor{yzythird}0.195 & \cellcolor{yzythird}0.212 & \cellcolor{yzysecond}0.189 \\
\hline
invrender & 0.175 & 0.263 & 0.115 & \cellcolor{yzythird}0.236 & 0.199 & 0.248 & 0.206 \\
nvdiffrec & 0.317 & \cellcolor{yzybest}0.199 & 0.227 & \cellcolor{yzybest}0.182 & 0.244 & 0.344 & 0.252 \\
\hline
\end{tabular}}
\caption{\textbf{Per-scene LPIPS (VGG) results of the roughness.}}
\label{tab:lpips-roughness}
\end{table}

\paragraph{Visualization of indirect radiance field.}
\par We visualized our indirect radiance field in Fig. \ref{fig:hdr-indir}. LDR represents the indirect radiance field obtained directly under the supervision of NeuS's radiance field, with a value range $\in [0, 1]$. However, this direct representation can lead to residual shadows and indirect illumination in PBR materials. Therefore, we use ACES non-linear mapping for the indirect radiance field, which enhances the contrast in different areas. The introduction of $\gamma$ makes the non-linear mapping adapt to scenes with various lighting intensities, thereby more accurately representing the real indirect illumination in the scene.

\paragraph{Analysis of masked indirect radiance field.} As shown in Fig. \ref{fig:segment-indir}, our experimental results exhibit a notable enhancement in the representation of indirect light when comparing our masked indirect radiance field ($L_I$) to InvRender. We use an MLP $\Gamma$ to learn the parameters of the indirect SGs for each surface point \textbf{x}. While our approach predicts superior indirect lighting at boundaries by randomly sampling directions for indirect SGs, we observed a significant difference under the parallel direction sampling condition, where only a single specific direction is sampled for visualization. This discrepancy arises from the inherent contradiction between the smooth SG basis functions and the discontinuous indirect illumination in our modeling approach. In the absence of masks, directions without indirect illumination will impact the learning of the parameters of SGs at the point \textbf{x}. As a result, there exists a "learning trap", where all SG coefficients diminish to zero. Our proposed approach adeptly addresses this issue, resulting in a more robust and accurate representation of the indirect radiance field.

%% file: main.bbl
\begin{thebibliography}{10}\itemsep=-1pt

\bibitem{arrighetti2017academy}
Walter Arrighetti.
\newblock The academy color encoding system (aces): A professional color-management framework for production, post-production and archival of still and motion pictures.
\newblock {\em Journal of Imaging}, 3(4):40, 2017.

\bibitem{barron2014shape}
Jonathan~T Barron and Jitendra Malik.
\newblock Shape, illumination, and reflectance from shading.
\newblock {\em IEEE Transactions on Pattern Analysis and Machine Intelligence}, 37(8):1670--1687, 2014.

\bibitem{bi2020deep}
Sai Bi, Zexiang Xu, Kalyan Sunkavalli, Milo{\v{s}} Ha{\v{s}}an, Yannick Hold-Geoffroy, David Kriegman, and Ravi Ramamoorthi.
\newblock Deep reflectance volumes: Relightable reconstructions from multi-view photometric images.
\newblock In {\em European Conference on Computer Vision}, pages 294--311. Springer, 2020.

\bibitem{boss2021nerd}
Mark Boss, Raphael Braun, Varun Jampani, Jonathan~T Barron, Ce Liu, and Hendrik Lensch.
\newblock Nerd: Neural reflectance decomposition from image collections.
\newblock In {\em Proceedings of the IEEE/CVF International Conference on Computer Vision}, pages 12684--12694, 2021.

\bibitem{boss2022samurai}
Mark Boss, Andreas Engelhardt, Abhishek Kar, Yuanzhen Li, Deqing Sun, Jonathan Barron, Hendrik Lensch, and Varun Jampani.
\newblock Samurai: Shape and material from unconstrained real-world arbitrary image collections.
\newblock {\em Advances in Neural Information Processing Systems}, 35:26389--26403, 2022.

\bibitem{boss2021neural}
Mark Boss, Varun Jampani, Raphael Braun, Ce Liu, Jonathan Barron, and Hendrik Lensch.
\newblock Neural-pil: Neural pre-integrated lighting for reflectance decomposition.
\newblock {\em Advances in Neural Information Processing Systems}, 34:10691--10704, 2021.

\bibitem{boss2020two}
Mark Boss, Varun Jampani, Kihwan Kim, Hendrik Lensch, and Jan Kautz.
\newblock Two-shot spatially-varying brdf and shape estimation.
\newblock In {\em Proceedings of the IEEE/CVF Conference on Computer Vision and Pattern Recognition}, pages 3982--3991, 2020.

\bibitem{burley2012physically}
Brent Burley and Walt Disney~Animation Studios.
\newblock Physically-based shading at disney.
\newblock In {\em Acm Siggraph}, volume 2012, pages 1--7. vol. 2012, 2012.

\bibitem{Chen2022ECCV}
Anpei Chen, Zexiang Xu, Andreas Geiger, Jingyi Yu, and Hao Su.
\newblock Tensorf: Tensorial radiance fields.
\newblock In {\em European Conference on Computer Vision (ECCV)}, 2022.

\bibitem{chen2022tracing}
Ziyu Chen, Chenjing Ding, Jianfei Guo, Dongliang Wang, Yikang Li, Xuan Xiao, Wei Wu, and Li Song.
\newblock L-tracing: Fast light visibility estimation on neural surfaces by sphere tracing.
\newblock In {\em European Conference on Computer Vision}, pages 217--233. Springer, 2022.

\bibitem{chen2022mobilenerf}
Zhiqin Chen, Thomas Funkhouser, Peter Hedman, and Andrea Tagliasacchi.
\newblock Mobilenerf: Exploiting the polygon rasterization pipeline for efficient neural field rendering on mobile architectures.
\newblock {\em arXiv preprint arXiv:2208.00277}, 2022.

\bibitem{cheng2021multi}
Ziang Cheng, Hongdong Li, Yuta Asano, Yinqiang Zheng, and Imari Sato.
\newblock Multi-view 3d reconstruction of a texture-less smooth surface of unknown generic reflectance.
\newblock In {\em Proceedings of the IEEE/CVF Conference on Computer Vision and Pattern Recognition}, pages 16226--16235, 2021.

\bibitem{garbin2021fastnerf}
Stephan~J Garbin, Marek Kowalski, Matthew Johnson, Jamie Shotton, and Julien Valentin.
\newblock Fastnerf: High-fidelity neural rendering at 200fps.
\newblock In {\em Proceedings of the IEEE/CVF International Conference on Computer Vision}, pages 14346--14355, 2021.

\bibitem{hasselgren2022nvdiffrecmc}
Jon Hasselgren, Nikolai Hofmann, and Jacob Munkberg.
\newblock {Shape, Light, and Material Decomposition from Images using Monte Carlo Rendering and Denoising}.
\newblock {\em arXiv:2206.03380}, 2022.

\bibitem{hedman2021baking}
Peter Hedman, Pratul~P Srinivasan, Ben Mildenhall, Jonathan~T Barron, and Paul Debevec.
\newblock Baking neural radiance fields for real-time view synthesis.
\newblock In {\em Proceedings of the IEEE/CVF International Conference on Computer Vision}, pages 5875--5884, 2021.

\bibitem{huang2022hdr}
Xin Huang, Qi Zhang, Ying Feng, Hongdong Li, Xuan Wang, and Qing Wang.
\newblock Hdr-nerf: High dynamic range neural radiance fields.
\newblock In {\em Proceedings of the IEEE/CVF Conference on Computer Vision and Pattern Recognition}, pages 18398--18408, 2022.

\bibitem{Jin2023TensoIR}
Haian Jin, Isabella Liu, Peijia Xu, Xiaoshuai Zhang, Songfang Han, Sai Bi, Xiaowei Zhou, Zexiang Xu, and Hao Su.
\newblock Tensoir: Tensorial inverse rendering.
\newblock In {\em Proceedings of the IEEE/CVF Conference on Computer Vision and Pattern Recognition (CVPR)}, 2023.

\bibitem{knodt2021nreural}
Julian Knodt, Joe Bartusek, Seung-Hwan Baek, and Felix Heide.
\newblock Neural ray-tracing: Learning surfaces and reflectance for relighting and view synthesis.
\newblock {\em arXiv preprint arXiv:2104.13562}, 2021.

\bibitem{kuang2022neroic}
Zhengfei Kuang, Kyle Olszewski, Menglei Chai, Zeng Huang, Panos Achlioptas, and Sergey Tulyakov.
\newblock Neroic: Neural rendering of objects from online image collections.
\newblock {\em ACM Trans. Graph.}, 41(4), jul 2022.

\bibitem{lensch2003image}
Hendrik~PA Lensch, Jan Kautz, Michael Goesele, Wolfgang Heidrich, and Hans-Peter Seidel.
\newblock Image-based reconstruction of spatial appearance and geometric detail.
\newblock {\em ACM Transactions on Graphics (TOG)}, 22(2):234--257, 2003.

\bibitem{li2018differentiable}
Tzu-Mao Li, Miika Aittala, Fr{\'e}do Durand, and Jaakko Lehtinen.
\newblock Differentiable monte carlo ray tracing through edge sampling.
\newblock {\em ACM Transactions on Graphics (TOG)}, 37(6):1--11, 2018.

\bibitem{li2020inverse}
Zhengqin Li, Mohammad Shafiei, Ravi Ramamoorthi, Kalyan Sunkavalli, and Manmohan Chandraker.
\newblock Inverse rendering for complex indoor scenes: Shape, spatially-varying lighting and svbrdf from a single image.
\newblock In {\em Proceedings of the IEEE/CVF Conference on Computer Vision and Pattern Recognition}, pages 2475--2484, 2020.

\bibitem{li2018learning}
Zhengqin Li, Zexiang Xu, Ravi Ramamoorthi, Kalyan Sunkavalli, and Manmohan Chandraker.
\newblock Learning to reconstruct shape and spatially-varying reflectance from a single image.
\newblock {\em ACM Transactions on Graphics (TOG)}, 37(6):1--11, 2018.

\bibitem{liu2020neural}
Lingjie Liu, Jiatao Gu, Kyaw Zaw~Lin, Tat-Seng Chua, and Christian Theobalt.
\newblock Neural sparse voxel fields.
\newblock {\em Advances in Neural Information Processing Systems}, 33:15651--15663, 2020.

\bibitem{liu2023nero}
Yuan Liu, Peng Wang, Cheng Lin, Xiaoxiao Long, Jiepeng Wang, Lingjie Liu, Taku Komura, and Wenping Wang.
\newblock Nero: Neural geometry and brdf reconstruction of reflective objects from multiview images.
\newblock In {\em SIGGRAPH}, 2023.

\bibitem{martel2021acorn}
Julien N.~P. Martel, David~B. Lindell, Connor~Z. Lin, Eric~R. Chan, Marco Monteiro, and Gordon Wetzstein.
\newblock Acorn: {Adaptive} coordinate networks for neural scene representation.
\newblock {\em ACM Trans. Graph. (SIGGRAPH)}, 40(4), 2021.

\bibitem{martin2021nerf}
Ricardo Martin-Brualla, Noha Radwan, Mehdi~SM Sajjadi, Jonathan~T Barron, Alexey Dosovitskiy, and Daniel Duckworth.
\newblock Nerf in the wild: Neural radiance fields for unconstrained photo collections.
\newblock In {\em Proceedings of the IEEE/CVF Conference on Computer Vision and Pattern Recognition}, pages 7210--7219, 2021.

\bibitem{Meder2018HemisphericalGF}
Julian Meder and Beat~D. Br{\"u}derlin.
\newblock Hemispherical gaussians for accurate light integration.
\newblock In {\em International Conference on Computer Vision and Graphics}, 2018.

\bibitem{10.1145/3374753}
Abhimitra Meka, Mohammad Shafiei, Michael Zollh\"{o}fer, Christian Richardt, and Christian Theobalt.
\newblock Real-time global illumination decomposition of videos.
\newblock {\em ACM Transactions on Graphics}, 40(3), aug 2021.

\bibitem{mildenhall2022nerf}
Ben Mildenhall, Peter Hedman, Ricardo Martin-Brualla, Pratul~P Srinivasan, and Jonathan~T Barron.
\newblock Nerf in the dark: High dynamic range view synthesis from noisy raw images.
\newblock In {\em Proceedings of the IEEE/CVF Conference on Computer Vision and Pattern Recognition}, pages 16190--16199, 2022.

\bibitem{mildenhall2020nerf}
Ben Mildenhall, Pratul~P. Srinivasan, Matthew Tancik, Jonathan~T. Barron, Ravi Ramamoorthi, and Ren Ng.
\newblock Nerf: Representing scenes as neural radiance fields for view synthesis.
\newblock In {\em ECCV}, 2020.

\bibitem{mueller2022instant}
Thomas M\"uller, Alex Evans, Christoph Schied, and Alexander Keller.
\newblock Instant neural graphics primitives with a multiresolution hash encoding.
\newblock {\em ACM Trans. Graph.}, 41(4):102:1--102:15, July 2022.

\bibitem{munkberg2022extracting}
Jacob Munkberg, Jon Hasselgren, Tianchang Shen, Jun Gao, Wenzheng Chen, Alex Evans, Thomas M{\"u}ller, and Sanja Fidler.
\newblock Extracting triangular 3d models, materials, and lighting from images.
\newblock In {\em Proceedings of the IEEE/CVF Conference on Computer Vision and Pattern Recognition}, pages 8280--8290, 2022.

\bibitem{nimier2021material}
Merlin Nimier-David, Zhao Dong, Wenzel Jakob, and Anton Kaplanyan.
\newblock Material and lighting reconstruction for complex indoor scenes with texture-space differentiable rendering.
\newblock 2021.

\bibitem{oechsle2021unisurf}
Michael Oechsle, Songyou Peng, and Andreas Geiger.
\newblock Unisurf: Unifying neural implicit surfaces and radiance fields for multi-view reconstruction.
\newblock In {\em Proceedings of the IEEE/CVF International Conference on Computer Vision}, pages 5589--5599, 2021.

\bibitem{reiser2021kilonerf}
Christian Reiser, Songyou Peng, Yiyi Liao, and Andreas Geiger.
\newblock Kilonerf: Speeding up neural radiance fields with thousands of tiny mlps.
\newblock In {\em Proceedings of the IEEE/CVF International Conference on Computer Vision}, pages 14335--14345, 2021.

\bibitem{sang2020single}
Shen Sang and Manmohan Chandraker.
\newblock Single-shot neural relighting and svbrdf estimation.
\newblock In {\em European Conference on Computer Vision}, pages 85--101. Springer, 2020.

\bibitem{schmitt2020joint}
Carolin Schmitt, Simon Donne, Gernot Riegler, Vladlen Koltun, and Andreas Geiger.
\newblock On joint estimation of pose, geometry and svbrdf from a handheld scanner.
\newblock In {\em Proceedings of the IEEE/CVF Conference on Computer Vision and Pattern Recognition}, pages 3493--3503, 2020.

\bibitem{sengupta2019neural}
Soumyadip Sengupta, Jinwei Gu, Kihwan Kim, Guilin Liu, David~W Jacobs, and Jan Kautz.
\newblock Neural inverse rendering of an indoor scene from a single image.
\newblock In {\em Proceedings of the IEEE/CVF International Conference on Computer Vision}, pages 8598--8607, 2019.

\bibitem{song2022novel}
Shuang Song and Rongjun Qin.
\newblock A novel intrinsic image decomposition method to recover albedo for aerial images in photogrammetry processing, 2022.

\bibitem{srinivasan2021nerv}
Pratul~P Srinivasan, Boyang Deng, Xiuming Zhang, Matthew Tancik, Ben Mildenhall, and Jonathan~T Barron.
\newblock Nerv: Neural reflectance and visibility fields for relighting and view synthesis.
\newblock In {\em Proceedings of the IEEE/CVF Conference on Computer Vision and Pattern Recognition}, pages 7495--7504, 2021.

\bibitem{sun2022direct}
Cheng Sun, Min Sun, and Hwann-Tzong Chen.
\newblock Direct voxel grid optimization: Super-fast convergence for radiance fields reconstruction.
\newblock In {\em Proceedings of the IEEE/CVF Conference on Computer Vision and Pattern Recognition}, pages 5459--5469, 2022.

\bibitem{sun2022improved}
Cheng Sun, Min Sun, and Hwann-Tzong Chen.
\newblock Improved direct voxel grid optimization for radiance fields reconstruction.
\newblock {\em arXiv preprint arXiv:2206.05085}, 2022.

\bibitem{sun2022neural}
Jiaming Sun, Xi Chen, Qianqian Wang, Zhengqi Li, Hadar Averbuch-Elor, Xiaowei Zhou, and Noah Snavely.
\newblock Neural 3d reconstruction in the wild.
\newblock In {\em ACM SIGGRAPH 2022 Conference Proceedings}, pages 1--9, 2022.

\bibitem{verbin2022ref}
Dor Verbin, Peter Hedman, Ben Mildenhall, Todd Zickler, Jonathan~T Barron, and Pratul~P Srinivasan.
\newblock Ref-nerf: structured view-dependent appearance for neural radiance fields.
\newblock In {\em 2022 IEEE/CVF Conference on Computer Vision and Pattern Recognition (CVPR)}, pages 5481--5490. IEEE, 2022.

\bibitem{vicini2022differentiable}
Delio Vicini, S{\'e}bastien Speierer, and Wenzel Jakob.
\newblock Differentiable signed distance function rendering.
\newblock {\em ACM Transactions on Graphics (TOG)}, 41(4):1--18, 2022.

\bibitem{wang2021neus}
Peng Wang, Lingjie Liu, Yuan Liu, Christian Theobalt, Taku Komura, and Wenping Wang.
\newblock Neus: Learning neural implicit surfaces by volume rendering for multi-view reconstruction.
\newblock {\em NeurIPS}, pages 27171--27183, 2021.

\bibitem{yang2022ps}
Wenqi Yang, Guanying Chen, Chaofeng Chen, Zhenfang Chen, and Kwan-Yee~K Wong.
\newblock Ps-nerf: Neural inverse rendering for multi-view photometric stereo.
\newblock In {\em Computer Vision--ECCV 2022: 17th European Conference, Tel Aviv, Israel, October 23--27, 2022, Proceedings, Part I}, pages 266--284. Springer, 2022.

\bibitem{yao2022neilf}
Yao Yao, Jingyang Zhang, Jingbo Liu, Yihang Qu, Tian Fang, David McKinnon, Yanghai Tsin, and Long Quan.
\newblock Neilf: Neural incident light field for physically-based material estimation.
\newblock In {\em Computer Vision--ECCV 2022: 17th European Conference, Tel Aviv, Israel, October 23--27, 2022, Proceedings, Part XXXI}, pages 700--716. Springer, 2022.

\bibitem{yariv2020multiview}
Lior Yariv, Yoni Kasten, Dror Moran, Meirav Galun, Matan Atzmon, Basri Ronen, and Yaron Lipman.
\newblock Multiview neural surface reconstruction by disentangling geometry and appearance.
\newblock {\em Advances in Neural Information Processing Systems}, 33:2492--2502, 2020.

\bibitem{yu2021plenoctrees}
Alex Yu, Ruilong Li, Matthew Tancik, Hao Li, Ren Ng, and Angjoo Kanazawa.
\newblock Plenoctrees for real-time rendering of neural radiance fields.
\newblock In {\em Proceedings of the IEEE/CVF International Conference on Computer Vision}, pages 5752--5761, 2021.

\bibitem{yu2019inverserendernet}
Ye Yu and William~AP Smith.
\newblock Inverserendernet: Learning single image inverse rendering.
\newblock In {\em Proceedings of the IEEE/CVF Conference on Computer Vision and Pattern Recognition}, pages 3155--3164, 2019.

\bibitem{zhang2021ners}
Jason Zhang, Gengshan Yang, Shubham Tulsiani, and Deva Ramanan.
\newblock Ners: neural reflectance surfaces for sparse-view 3d reconstruction in the wild.
\newblock {\em Advances in Neural Information Processing Systems}, 34:29835--29847, 2021.

\bibitem{zhang2022iron}
Kai Zhang, Fujun Luan, Zhengqi Li, and Noah Snavely.
\newblock Iron: Inverse rendering by optimizing neural sdfs and materials from photometric images.
\newblock In {\em Proceedings of the IEEE/CVF Conference on Computer Vision and Pattern Recognition}, pages 5565--5574, 2022.

\bibitem{zhang2021physg}
Kai Zhang, Fujun Luan, Qianqian Wang, Kavita Bala, and Noah Snavely.
\newblock Physg: Inverse rendering with spherical gaussians for physics-based material editing and relighting.
\newblock In {\em Proceedings of the IEEE/CVF Conference on Computer Vision and Pattern Recognition}, pages 5453--5462, 2021.

\bibitem{zhang2021nerfactor}
Xiuming Zhang, Pratul~P Srinivasan, Boyang Deng, Paul Debevec, William~T Freeman, and Jonathan~T Barron.
\newblock Nerfactor: Neural factorization of shape and reflectance under an unknown illumination.
\newblock {\em ACM Transactions on Graphics (TOG)}, 40(6):1--18, 2021.

\bibitem{zhang2022modeling}
Yuanqing Zhang, Jiaming Sun, Xingyi He, Huan Fu, Rongfei Jia, and Xiaowei Zhou.
\newblock Modeling indirect illumination for inverse rendering.
\newblock In {\em Proceedings of the IEEE/CVF Conference on Computer Vision and Pattern Recognition}, pages 18643--18652, 2022.

\end{thebibliography}
